\documentclass{article} 
\usepackage{iclr2022_conference,times}

\newcommand{\Appendix}[1]{the full version for}

\newcommand{\x}{\bm{x}}

\newcommand{\z}{\bm{z}}

\newcommand{\I}{\bm{I}}

\newcommand{\R}{\mathbb{R}}
\renewcommand{\Re}{\mathbb{R}}

\newcommand{\X}{\bm{X}}

\newcommand{\Z}{\bm{Z}}

\renewcommand{\mathbf}{\boldsymbol}

\usepackage{url}
\usepackage{bm}
\usepackage{amsfonts}
\usepackage{amsmath}
\usepackage{amssymb}
\usepackage{amsthm}
\usepackage{graphicx}
\usepackage{subfigure}
\usepackage{caption}
\usepackage[normalem]{ulem}
\usepackage{wrapfig}
\usepackage{times,epsfig,graphics,graphicx,caption,float,booktabs,xcolor,multirow,multicol,array,color,ifthen,tabu,colortbl,url,xparse,mathtools,patchcmd,enumitem,amssymb,xspace,nicefrac,microtype,amsmath,amsfonts}

\usepackage[pdftex,bookmarksnumbered,bookmarksopen,colorlinks,citecolor=blue,linkcolor=blue,urlcolor=blue]{hyperref}

\definecolor{fuchsia(web)}{rgb}{0.65, 0.2, 0.65}

\newcommand{\ours}{LDR}

\title{\centering{ Closed-Loop Data Transcription to an LDR via Minimaxing Rate Reduction} }

\author{
\centerline{
Xili Dai\textsuperscript{\rm 1,2}\quad
Shengbang Tong\textsuperscript{\rm 1} \quad
Mingyang Li\textsuperscript{\rm 3} \quad
Ziyang Wu\textsuperscript{\rm 4}} \quad \\
\centerline{
\textbf{Michael Psenka}\textsuperscript{\rm 1}\quad  
\textbf{Kwan Ho Ryan Chan}\textsuperscript{\rm 5}\quad 
\textbf{Pengyuan Zhai}\textsuperscript{\rm 6}\quad
\textbf{Yaodong Yu}\textsuperscript{\rm 1} }\\
\centerline{
\textbf{Xiaojun Yuan}\textsuperscript{\rm 2}\quad
\textbf{Heung-Yeung Shum}\textsuperscript{\rm 4}\quad
\textbf{Yi Ma}\textsuperscript{\rm 1,3}
}\\\\
\centerline{
\textsuperscript{\rm 1}University of California, Berkeley  \quad
\textsuperscript{\rm 2}University of Electronic Science and Technology of China }\\
\centerline{
\textsuperscript{\rm 3}Tsinghua-Berkeley Shenzhen Institute (TBSI) \quad \textsuperscript{\rm 4}International Digital Economy Academy (IDEA) }
\\
\centerline{
 \textsuperscript{\rm 5}Johns Hopkins University \quad
\textsuperscript{\rm 6}Harvard University}
}

\iclrfinalcopy 
\begin{document}

\maketitle

\begin{abstract}
This work proposes a new computational framework for learning a structured generative model for real-world  datasets. In particular, we propose to learn  {\em a closed-loop transcription} between a multi-class multi-dimensional data distribution and a {linear discriminative representation (LDR)} in the feature space that consists of multiple  independent  multi-dimensional linear  subspaces. In particular, we argue that the optimal encoding and decoding mappings sought can be formulated as the equilibrium point of a {\em two-player minimax game between the encoder and decoder}. A natural utility function for this game is the so-called {\em rate reduction}, a simple information-theoretic measure for distances between mixtures of subspace-like Gaussians in the feature space. Our formulation draws inspiration from closed-loop error feedback from control systems and avoids expensive evaluating and minimizing approximated distances between arbitrary distributions in either the data space or the feature space. To a large extent, this new formulation unifies the concepts and benefits of Auto-Encoding and GAN and naturally extends them to the settings of learning a {\em both discriminative and generative} representation for  multi-class and multi-dimensional real-world data. Our extensive experiments on many benchmark imagery datasets demonstrate tremendous potential of this new closed-loop formulation: under fair comparison, visual quality of the learned decoder and classification performance of the encoder is competitive and often better than existing methods based on GAN, VAE, or a combination of both. Unlike existing generative models, the so learned features of the multiple classes are structured: different classes are  explicitly mapped onto corresponding {\em independent principal subspaces} in the feature space; and diverse visual attributes within each class are modeled by the {\em independent principal components} within each subspace. Source code can be found at \href{https://github.com/Delay-Xili/LDR}{\textsc{}{https://github.com/Delay-Xili/LDR}}.

\end{abstract}

\vspace{1mm}

\vspace{1cm}
\begin{quotation}
\normalsize
$~$ \hfill ``{\em Learners need endless feedback more than they need endless teaching.}''\\
$~$ \hspace{\fill} --- Grant Wiggins
\end{quotation}

\newpage
\tableofcontents
\newpage

\section{Introduction}
One of the most fundamental tasks in modern data science and  machine learning  is to learn and model complex distributions (or structures) of real-world data, such as images or texts, from a set of observed samples. By ``to learn and model'', one typically means that we want to establish a (parameteric) mapping between the distribution of the real data, say $\x \in \Re^D$, and a more compact random variable, say $\z \in \Re^d$:
\begin{equation}
    f(\cdot,\theta): \x \in \Re^D \mapsto \z \in \Re^d \quad \mbox{or the inverse} \quad g(\cdot, \eta): \z \in \Re^d \mapsto \x \in \Re^D,
    \label{eqn:mappings}
\end{equation}
where $\z$ has certain standard structure or distribution (e.g. normal distributions). The so-learned representation or feature $\z$ would be much easier to use for either generative (e.g. decoding or replaying) or discriminative (e.g. classification) purposes, or both. 

\paragraph{Data embedding versus data transcription.} {\em Be aware} that the support of the distribution of $\x$ (and that of $\z$) is typically {\em extremely low-dimensional} compared to that of the ambient space,\footnote{For instance, the well-known CIFAR-10 datasets consist of RGB images with a resolution $32\times 32$. Despite the images are in a space of $\Re^{3072}$, our experiments will show that the intrinsic dimension of each class is less than a dozen, even after they are mapped into a feature space of $\Re^{128}$.} hence the above mapping(s) may not be uniquely defined off the support in the space $\Re^D$ (or $\Re^d)$. Also, the data $\x$ may contain multiple components (e.g. modes, classes), and the intrinsic dimensions of these components are not necessarily the same. Hence, without loss of generality, we may assume the data $\x$ is distributed over a union of low-dimensional nonlinear submanifolds $\cup_{j=1}^k \mathcal{M}_j \subset \Re^D$ where each submanifold $\mathcal{M}_j$ is of dimension $d_j \ll D$. Regardless, we hope the learned mappings $f$ and $g$ are (locally dimension-preserving) {\em embedding} maps  \citep{Lee2002IntroductionTS}, when restricted to each of the components $\mathcal{M}_j$. 

In general, the dimension of the feature space $d$ needs to be significantly higher than all of these intrinsic dimensions of the data: $ d > d_j$. In fact, it is preferably higher than the sum of all the intrinsic dimensions: $d \ge d_1 + \cdots + d_k$ since we normally  expect the features of different  components/classes can be made fully independent or orthogonal in $\Re^d$. Hence, without any explicit control of the mapping process, the actual features associated with images of the data under the embedding could still lie on some arbitrary nonlinear low-dimensional submanifolds inside the feature space $\Re^d$. The distribution of the learned features remains ``latent'' or ``hidden'' in the feature space.

So for features of the learned mappings \eqref{eqn:mappings} to be truly convenient to use for purposes such as data classification and generation, the goals of learning such mappings should not only simply reduce the dimension of the data $\x$ from $D$ to $d$ but also determine explicitly and precisely how the mapped feature $\z = f(\x)$ is distributed within the feature space $\Re^d$, both its support and density! Moreover, we want to establish an explicit map $g(\cdot)$ from this distribution of feature $\z$ back to the data space such that the distribution of its image $\hat{\x} = g(\z)$ (closely) matches that of $\x$. To differentiate from finding arbitrary feature embeddings (as most existing methods do), we call embeddings of data onto an explicit family of models (structures or distributions) in the feature space as {\em data transcription}. 

\paragraph{Paper Outline.} This work is to show how such transcription can be done for real-world visual data with one important family of models: the linear discriminative representation (LDR)  introduced by \cite{ReduNet}. Before we formally introduce our approach in Section \ref{sec:drive-minimax}, for the remainder of this section, we first discuss two existing approaches, namely autoencoding and GAN, that are closely related to ours. As these approaches are rather popular and known to the readers, we will mainly point out some of their main conceptual and practical limitations that have motivated this work. Although our objective and framework will be mathematically formulated, the main purpose of this work is to verify the effectiveness of this new approach empirically through extensive experimentation, organized and presented in Section \ref{sec:experiments} and Appendix \ref{app:appendix}. Our work presents compelling evidences that the closed-loop data transcription problem and our rate reduction based formulation deserve serious attention from the information-theoretical and mathematical community. It has raised many exciting and open theoretical problems or hypotheses about learning, representing, and generating distributions or manifolds of high-dimensional real-world data. We discuss some open problems in Section \ref{sec:discussions} and new directions in Section \ref{sec:conclusion}.

\subsection{Learning Generative Models via Auto-Encoding or GAN}
\noindent\textbf{Auto-Encoding and its variants.} In the machine learning literature, roughly speaking,  there have been two representative approaches to such a distribution-learning task. One is the classic ``Auto Encoding'' (AE) approach \citep{Kramer1991NonlinearPC,auto-encoding} that aims to simultaneously learn an encoding mapping $f$ from $\x$ to $\z$ and an (inverse) decoding mapping $g$ from $\z$ back to $\x$:
\begin{equation}
    \X \xrightarrow{\hspace{2mm} f(\x, \theta)\hspace{2mm}} \Z \xrightarrow{\hspace{2mm} g(\z,\eta) \hspace{2mm}} \hat \X.
    \label{eqn:auto-encoding}
\end{equation}
Here we use bold capital letters to indicate a matrix of finite samples  $\X = [\x^1, \ldots, \x^n] \in \Re^{D \times n}$ of $\x$ and their mapped features $\Z = [\z^1, \ldots, \z^n] \subset \Re^{d \times n}$, respectively. Typically, one wishes for two properties: firstly, the decoded samples $\hat \X$ are ``similar'' or close to the original $\X$, say in terms of maximum likelihood $p(\X)$; and secondly, the (empirical) distribution of the mapped samples $\Z$, denoted as $\hat{p}(\z | \X)$, is close to certain desired prior distribution $p(\z)$, say some much lower-dimensional multivariate Gaussian\footnote{The classical PCA can be viewed as a special case of this task. In fact, the original auto-encoding is precisely cast as {\em nonlinear} PCA \citep{Kramer1991NonlinearPC}, assuming the data lie on only one nonlinear submanifold $\mathcal{M}$.}. 

However it is typically very difficult, often computationally intractable to maximize the likelihood function $p(\X)$ or to minimize certain ``distance'', say the {\em KL-divergence} $\mathcal{D}_{KL}(\hat{p}, p)$, between $\hat p(\z|\X)$ and $p(\z)$. Except for simple distributions such as Gaussian, the KL divergence usually does not have a closed-form, even for a mixture of Gaussians. The likelihood and the KL-divergence become ill-conditioned when the supports of the distributions are low-dimensional (i.e. degenerate) and not overlapping.\footnote{which is almost always the case in practice when dealing with distributions of  high-dimensional data in high-dimensional spaces.} So in practice, one typically chooses to minimize instead certain approximate bounds or  surrogates derived with various simplifying assumptions on the distributions involved, as done in variational auto-encoding (VAE) \citep{kingma2013auto,InfoVAE}. As result, even after learning, the precise posterior distribution of $\hat{p}(\z | \X)$ remains unclear or hidden inside the feature space. 

In this work, we will show that if we impose specific requirements on the (distribution of) learned feature $\z$ to be a mixture of subspace-like Gaussians, a natural closed-form distance can be introduced for such distributions based on rate distortion from the information theory. In addition, the optimal solution to the feature representation within this family can be learned directly from the data {\em without specifying any target $p(\z)$ in advance}, which is particularly difficult to do in practice when the distribution of a mixed dataset is multi-modal and each component may have a different dimension.

\noindent\textbf{GAN and its variants.} Compared to measuring distribution distance in the (often controlled) feature space $\z$, a much more challenging issue with the above auto-encoding approach is how to effectively measure the distance between the decoded samples $\hat \X$ and the original $\X$ in the data space $\x$. For instance, for visual data such as images, their distributions $p(\X)$ or generative models $p(\X | \z)$ are often not known. Despite extensive studies in the computer vision and image processing literature \citep{image-similarity}, it remains elusive to find a good measure for similarity of real images that is both efficient to compute and effective in capturing visual quality and semantic information of the images equally well. Precisely due to such difficulties, it has been suggested early on by \cite{Tu-2007} that one may have to take a discriminative approach to learn the distribution or a generative model for visual data. More recently, {\em Generative Adversarial Nets (GAN)} \citep{goodfellow2014generative} offers an ingenious idea to alleviate this difficulty by utilizing a powerful discriminator $d$, usually modeled and learned by a deep network, to discern differences between the generated samples $\hat \X$ and the real ones $\X$:
\begin{equation}
 \Z \xrightarrow{\hspace{2mm} g(\z,\eta) \hspace{2mm}} \hat \X, \, \X \xrightarrow{\hspace{2mm} d(\x, \theta)\hspace{2mm}} \mathbf 0, \mathbf 1.
 \label{eqn:GAN}
\end{equation}
To a large extent, such a discriminator plays the role of minimizing certain distributional distance, e.g. the {\em Jensen-Shannon divergence}, between the data $\X$ and $\hat \X$. Compared to the KL-divergence, the JS-divergence is well defined even if the supports of the two distributions are non-overlapping.\footnote{However, JS-divergence does not have a closed-form expression even between two Gaussians whereas KL-divergence has.}  But as shown in \citep{arjovsky2017wasserstein}, since the data distributions are low-dimensional, the JS-divergence can be highly ill-conditioned to optimize.\footnote{This may explain why many additional heuristics are typically used in many subsequent variants of GAN.} So instead, one may choose to replace the JS-divergence with the earth mover's distance or the Wasserstein distance. However both JS-divergence and W-distance can only be approximately computed between two general distributions.\footnote{For instance, the W-distance requires to compute the maximal difference between expectations of the two distributions over all 1-Lipschitz functions.} Furthermore, neither the JS-divergence nor the W-distance has closed-form formulae even for the Gaussian distributions.\footnote{The ($\ell^1$-norm) W-distance can be bounded by the ($\ell^2$-norm) W2-distance which has a closed-form \citep{salmona2021gromovwasserstein}. However, as it is well known in high-dimensional geometry, $\ell^1$-norm and $\ell^2$ norm deviate significantly in terms of their geometric and statistical properties as the dimension gets high \citep{Wright-Ma-2021}. The bound can become very loose.} But from a data representation perspective, {\em subspace-like Gaussian  (e.g. PCA) or a mixture of them are the most desirable family of distributions that we wish our features to become!} It would make all subsequent tasks (generative or discriminative) much easier. {In this work we will show how to achieve this with a different fundamental metric, known as the rate reduction}, introduced by  \citep{yu2020learning}.

The original GAN aims to directly learn a mapping $g(\cdot)$, called a generator, from a standard distribution (say a low-dimensional Gaussian random field) to the real (visual) data distribution in a high-dimensional space. However, distributions of real-world data can be rather sophisticated and often contain {\em multiple} classes and {\em multiple} factors in each class \citep{representation_learning}. That makes learning the mapping $g$ rather challenging in practice, suffering difficulties such as {\em mode-collapse} \citep{veegan}. As a result, many variants of GAN have been subsequently developed in order to improve the stability and  performance in learning multiple modes and disentangling different factors in the data distribution, such as {\em Conditional GAN} \citep{cgan,cvae,disentangling_factors,van2016conditional,wang2018high}, {\em InfoGAN} \citep{infogan,disentangled-Tang}, or {\em Implicit Maximum  Likelihood Estimation (IMLE)} \citep{li2018implicit,li2020multimodal-IJCV}. In particular, to learn a generator for multi-class data, prevalent conditional GAN literature requires label information as conditional inputs  \citep{cgan,odena2017conditional,dumoulin2016learned,brock2018large}. Recently \cite{wu2019deep,wu2019logan} has proposed to train a $k$-class GAN by generalizing the two-class cross entropy to a $(k+1)$-class cross entropy. In this work, {\em we will introduce a more refined $2k$-class measure} for the $k$ real and $k$ generated classes. In addition, to avoid features for each class to collapse to a singleton \citep{papyan2020prevalence}, instead of cross entropy, {\em we will use the so-called rate reduction measure that promotes multi-mode and  multi-dimension in the learned features} \citep{yu2020learning}. One may view the rate reduction as a metric distance that has closed-form formulae for a mixture of (subspace-like) Gaussians whereas neither JS-divergence nor W-distance can be computed in closed form (even between two Gaussians).

Another line of research is about how to stablize the training of GAN. SN-GAN \citep{miyato2018spectral} has shown spectral normalization on the discriminator is rather effective, which we will adopt in our work, although our formulation is not so sensitive to such choice designed for GAN (see ablation study in Appendix \ref{app:spectral-normalization}). PacGAN \citep{lin2018pacgan} shows that the training stability can be significantly improved by packing a pair of real and generated images together for the discriminator. Inspired by this work, {\em we show how to generalize such an idea to discriminating an arbitrary number of pairs of real and decoded samples without concatenating the samples.} Our results in this work will even suggest that the larger the batch size discriminated, the merrier (see ablation study in Appendix \ref{app:ablation-batch-size}). Also, \cite{wu2019logan} has shown that optimizing the latent features  leads to state of the art visual quality. Their method is based on the deep compressed sensing GAN \citep{wu2019deep}. Hence there are strong reasons to believe that their method essentially utilizes the {\em compressed sensing} principle \citep{Wright-Ma-2021} to implicitly exploit  low-dimensionality of the feature distribution. Our framework {\em will explicitly expose and exploit such low-dimensional structures on the learned feature distribution.}

\paragraph{Combination of AE and GAN.}
Although AE (VAE) and GAN have originated with somewhat different motivations, they have evolved into popular and effective frameworks for learning and modeling complex distributions of many real-world data such as images.\footnote{In fact, in some idealistic settings, it can be shown that AE and GAN are actually equivalent: for instance, in the LOG settings, \cite{GAN-LOG} have shown that GAN coincides with the classic PCA, which is precisely the solution to Auto-Encoding in the linear case.}  Many recent efforts tend to combine both Auto-Encoding and GAN to generate more powerful generative frameworks for more diverse data sets, such as \cite{VAE-GAN,alpha_gan,veegan,bao2017cvae,huang2018introvae,donahue2016adversarial,dumoulin2016adversarially,Ulyanov2018AGE,vahdat2020nvae,parmar2021dual}. As we will see, in our framework, AE and GAN can be naturally interpreted as two different segments of a closed-loop data transcription process. But unlike GAN or AE (VAE), the ``origin'' or ``target'' distribution of the feature $\z$ will no longer be specified {\em a priori} and is instead learned from the data $\x$. In addition, {\em this intrinsically low-dimensional distribution of $\z$ (with all its low-dimensional supports) is explicitly modeled as a mixture of orthogonal subspaces (or independent Gaussians)  within the feature space $\Re^d$}, sometimes known as the principal subspaces.

\paragraph{Universality of Representations.} Note that GANs (and most VAEs) are typically designed without explicit modeling assumptions on the distribution of the data nor on the feature. Many even believe that it is this ``universal'' distribution learning capability\footnote{Assuming minimizing distances between arbitrary distributions in high-dimensional space can be solved efficiently, which unfortunately has many caveats and often is impractical.} that has attributed to their empirical success in learning distributions of complicated data such as images. In this work, we will provide empirical evidence that such an ``arbitrary distribution learning machine'' might not be necessary. A {\em controlled and deformed} family of low-dimensional linear subspaces (Gaussians) can be more than powerful and expressive enough to model real-world visual data.\footnote{In fact, a Gaussian mixture model is already a universal approximator of almost arbitrary densities \citep{GMMs}. Hence we do not loose any generality at all.} As we will also see, once we can put a proper and precise metric on such models, the associated learning problems can become much better conditioned and more amenable to rigorous analysis and performance guarantees in the future. 

\subsection{Learning Linear Discriminative Representation via Rate Reduction} 
Recently, \cite{ReduNet} proposed a new objective for deep learning that aims to learn a {\em linear discriminative representation} (LDR) for multi-class data. The basic idea is to map distributions of real data, potentially on {\em multiple} nonlinear submanifolds $\cup_{j=1}^k \mathcal{M}_j \subset \Re^D$,\footnote{In classical statistical settings, such nonlinear structures of the data were also referred to as principal curves or surfaces   \citep{HastieT1984,HastieT1987}. There has been a long quest trying to extend PCA to handle potential nonlinear low-dimensional structures in data distribution. See \citep{Vidal:Springer16} for a thorough survey.} to a family of canonical models consisting of multiple independent (or orthogonal) linear subspaces, denoted as $\cup_{j=1}^k \mathcal{S}_j \subset \Re^d$. To some extent, this generalizes the classic nonlinear PCA \citep{Kramer1991NonlinearPC} to more general/realistic settings where we simultaneously apply {\em multiple nonlinear PCAs} to data on multiple nonlinear submanifolds. Or equivalently, the problem can also be viewed as a nonlinear extension to the classic {\em  Generalized PCA} (GPCA)  \citep{Vidal:Springer16}.\footnote{Conventionally, ``generalized PCA'' refers to generalizing the setting of PCA to multiple {\em linear} subspaces. Here we need to further generalize it multiple {\em nonlinear} submanifolds!} Unlike conventional discriminative methods that only aim to predict class labels as one-hot vectors, the LDR aims to learn the likely multi-dimensional distribution of the data, hence is suitable for both discriminative and generative purposes. It has been shown that this can be achieved via maximizing the so-called ``rate reduction''  objective based on the rate distortion of subspace-like Gaussians \citep{ma2007segmentation}. 

\paragraph{LDR via MCR$^2$.} More precisely, consider a set of data samples $\X = [\x^1, \ldots, \x^n] \in \Re^{D \times n}$ from $k$ different classes. That is, we have $\X = \cup_{j=1}^k \X_j$ with each subset of samples $\X_j$ belonging to one of the low-dimensional submanifolds: $\X_j \subset \mathcal{M}_j, j = 1,\ldots, k$. Following the notation in \citep{ReduNet}, we use a matrix $\bm \Pi^j(i,i) = 1$ to denote the membership of sample $i$ belonging to class $j$ (and $\bm \Pi^j = 0$ otherwise). One seeks a continuous mapping $f(\cdot,\theta): \x \mapsto \z$ from $\X$ to an optimal representation $\Z = [\z^1, \ldots, \z^n] \subset \Re^{d \times n}$:
\begin{equation}
\bm X  \xrightarrow{\hspace{2mm} f(\x, \theta)\hspace{2mm}} \bm Z, 
\label{eqn:LDR}
\end{equation}
which maximizes the following coding rate reduction objective, known as {\em the MCR$^2$ principle} \citep{yu2020learning}: 
\begin{equation}\label{eq:mcr2-formulation}
\begin{split}
\max_{\Z} \; \Delta R(\Z\, | \, \bm{\Pi}, \epsilon) 
&\doteq \underbrace{\frac{1}{2}\log\det \Big(\I + {\alpha} \Z \Z^{*} \Big)}_{R(\Z\, | \epsilon)} \;-\; \underbrace{\sum_{j=1}^{k}\frac{\gamma_j}{2} \log\det\Big(\I + {\alpha_j} \Z \bm{\Pi}^{j} \Z^{*} \Big)}_{R_c(\Z \,|\bm \Pi, \epsilon)},
\end{split}
\end{equation}
where $\alpha=\frac{d}{n\epsilon^2}$, $\alpha_j=\frac{d}{\textsf{tr}(\bm{\Pi}^{j})\epsilon^2}$, $\gamma_j=\frac{\textsf{tr}(\bm{\Pi}^{j})}{n}$ for $j = 1,\ldots, k$. In this paper, for simplicity we denote $\Delta R(\Z \, | \bm{\Pi}, \epsilon)$ as $\Delta R(\Z)$ assuming $\bm \Pi, \epsilon$ are known and fixed. The first term $R(\Z\, | \epsilon)$, or $R(\Z)$ for short, is the coding rate of the whole feature set $\Z$ (coded as a Gaussian source) with a prescribed precision $\epsilon$; the second term $R_c(\Z \,|\bm \Pi, \epsilon)$, or simply $R_c(\Z)$, is the average coding rate of the $k$ subsets of features $\Z_j = f(\X_j)$ (each coded as a Gaussian). 

As it has been shown by \cite{yu2020learning}, maximizing the difference between the two terms will expand the whole feature set while compressing and linearizing features of each of the $k$ classes. If the mapping $f$ maximizes the rate reduction, it maps the features  of different classes into independent (orthogonal) subspaces in $\Re^d$. Figure \ref{fig:MCR2} illustrates a simple example of data with $k=2$ classes (on two submanifolds) mapped to two incoherent subspaces (solid black lines). Notice that, compared to AE \eqref{eqn:auto-encoding} and GAN \eqref{eqn:GAN}, the above mapping \eqref{eqn:LDR} is only one-sided: from the data $\X$ to the feature $\Z$. In this work, we will see how to use the rate reduction metric to establish the inverse mapping from the feature $\Z$ back to the data $\X$, while still preserving the subspace structures in the feature space.

\section{Data Transcription via Rate Reduction}\label{sec:drive-minimax}
\subsection{Closed-Loop LDR Transcription}
One issue with this one-sided LDR learning \eqref{eqn:LDR} is that maximizing the above objective \eqref{eq:mcr2-formulation} tends to expand the dimension of the learned subspace for features in each class.\footnote{If the dimension of the feature space $d$ is too high, maximizing the rate reduction may over-estimate the dimension of each class. Hence to learn a good representation, one needs to pre-select a proper dimension for the feature space, as done in the experiments in \citep{yu2020learning}. In fact the same ``model selection'' problem persists even in the simplest single subspace case, a.k.a. the classic PCA \citep{Jolliffe1986}. Selecting the correct number of principal components in heterogeneous noisy situation remains an active research topic \citep{hong2020selecting}.} To verify whether the learned features are neither over-estimating nor under-estimating the data structure, we may consider learning a decoder $g(\cdot,\eta): \z \mapsto  \x$ from the representation $\Z = f(\X,\theta)$ back to the data space $\x$: $\hat \X = g(\Z, \eta)$, and check how close $\X$ and $\hat \X$ are or how close their features $\Z$ and $\hat \Z = f(\hat \X, \theta)$ are. In principle, the decoder $g$ should examine if all the learned features by the encoder $f$ are both  necessary and sufficient for achieving this task. The overall pipeline can be illustrated by the following ``closed-loop''  diagram: 
\begin{equation}
    \X \xrightarrow{\hspace{2mm} f(\x, \theta)\hspace{2mm}} \Z \xrightarrow{\hspace{2mm} g(\z,\eta) \hspace{2mm}} \hat \X \xrightarrow{\hspace{2mm} f(\x, \theta)\hspace{2mm}} \ \hat \Z, 
\end{equation}
where the overall model has parameters: $\Theta = \{\theta, \eta\}$. 

Notice that in the above process, the segment from $\X$ to $\hat \X$ resembles a typical {\em Auto-Encoding} process. Although,  as we will soon see, our MCR$^2$-based encoder $f$ plays an additional role as a discriminator. The segment from $\Z$ to $\hat \Z$ draws resemblance to the typical GAN process; although, in our context,  the distribution of the latent variable $\z$ will be learned from the data $\x$. Despite these connections, as we will soon see, this new closed-loop formulation will allow us to utilize the {\em error feedback} mechanism (widely practiced in control systems) and directly enforce loop-consistency between  encoding and decoding (networks) {\em without} using any additional discriminator(s) that are typically needed in existing VAE/GAN architectures.

\begin{figure}[t]
\centerline{\includegraphics[width=5.3in]{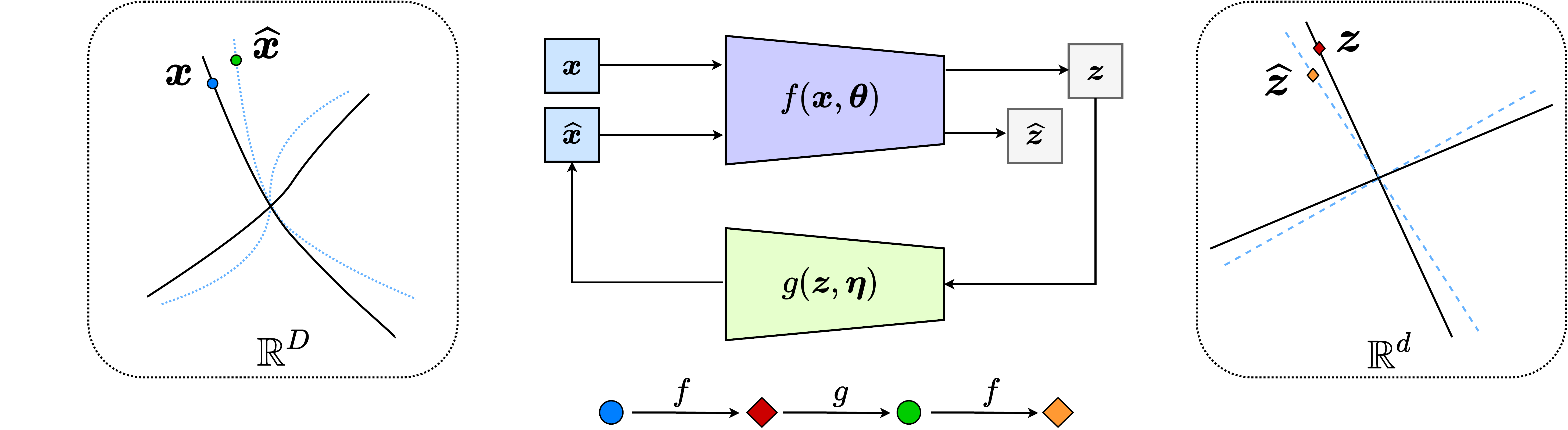}}
\caption{{\bf A Closed-loop LDR Transcription Game.} The encoder $f$ has dual roles: it learns an LDR $\z$ for the data $\x$ via maximizing the rate reduction of $\z$ and it is also a ``feedback sensor'' for any discrepancy between the data $\x$ and the decoded $\hat \x$. The decoder $g$ also has dual roles: it is a ``controller'' that corrects  the  discrepancy between $\x$ and $\hat \x$ and it also aims to minimize the overall coding rate for the learned LDR.} \label{fig:auto-encoding} \label{fig:MCR2}
\end{figure}

Here, in the specific context of rate reduction, we name this special auto-encoding process ``{\em LDR Transcription}'' since the maximal rate reduction principle explicitly transcribes the data $\X$, via $f$, to features $\Z$ on a linear discriminative representation (LDR),\footnote{Through our extensive experiments on diverse real world visual datasets, one does not lose any generality and expressiveness by restricting to this special but rich  class of models. On the contrary, the restriction significantly simplifies and improves the learning process.} which can be subsequently decoded back to the data space $\hat \X$,  via $g$. Hence, the encoding and decoding maps $f$ and $g$ together form a ``closed-loop'' process, as illustrated in Figure \ref{fig:auto-encoding}. We wish this closed-loop transcription process to have the following good properties:
\begin{itemize}
\item {\bf Injectivity:} the generated $\hat \x = g(f(\x, \theta), \eta) \in \hat \X$ to be as close to (ideally the same as) the original data $\x \in \X$, in terms of certain measure of similarity or distance.
\item {\bf Surjectivity:} for all mapped images $\z = f(\x) \in \Z$ of the training  data $\x \in \X$, there are decoded samples  $\hat \z = f(g(\z,\eta), \theta) \in \hat \Z$ close to (ideally the same as) $\z$. 
\end{itemize}
Mathematically, we seek an {\em embedding} of the data $\x$ supported on certain nonlinear submanifolds $\cup_{j=1}^k \mathcal{M}_j$ in the space $\Re^D$ to feature $\z$ on a set of (discriminative) linear subspaces $\cup_{j=1}^k \mathcal{S}_j$ in the feature space $\Re^d$. Ideally, both $f$ and $g$ should be embeddings \citep{Lee2002IntroductionTS},  when restricted on the support of the data distribution or that of the features.\footnote{That is, we hope $f\mid_{\mathcal{M}_j}$ and $g\mid_{\mathcal{S}_j} $ are all embeddings for all $j=1, \ldots, k$.} Also, more ideally, we would hope $f$ and $g$ are mutually inverse embeddings: $g\circ f = \mbox{Id}$ (when restricted on the submanifolds). Nevertheless, if we are only interested in learning the distribution, embeddings of the support would often suffice the purposes (e.g. classification or generative purposes). Notice that the above goals are similar to many VAE+GAN related methods in the machine learning literature, such as BiGAN \citep{donahue2016adversarial} and ALI \citep{dumoulin2016adversarially}. We will discuss the differences of our approach from these existing methods in Section \ref{sec:game-formulation} (as well as some experimental comparison in the Appendix). 

At first sight, this is a rather daunting task  since we are trying to learn over a (seemingly infinite-dimensional) functional space of all embeddings and distributions from finite samples! In this work, we will take a more  pragmatic approach and show how one can learn a  good encoding, decoding, and representation tuple:  $(f, g, \z)$ from $\X$ via tractable computational means. In particular, we will  convert the above goals to certain feasible programs that optimize a  sensible measure of goodness for the learned representations $\Z$. 

\subsection{Measuring Distances in the Feature Space and Data Space}\label{sec:measures}
\paragraph{Contractive measure for the decoder.} For the {\em second} item in the above wishlist, as the representations in the feature space $\z$ are by design   linear subspaces or (degenerate) Gaussians, we have geometrically or statistically meaningful metrics for both samples and distributions in the feature space $\z$. For example, we care about distance between distributions between the features of the original data $\Z$ and the transcribed $\hat \Z$. Since the features of each class, $\Z_j$ and $\hat{\Z}_j$, are like subspaces/Gaussians, their ``distance'' can be measured by the rate reduction, {with \eqref{eq:mcr2-formulation} restricted to two sets of equal size)}:
\begin{equation}
\Delta R\big(\Z_j, \hat{\Z}_j\big) \doteq R\big(\Z_j \cup \hat{\Z}_j\big) - \frac{1}{2} \big( R\big(\Z_j) + R\big(\hat \Z_j)\big).
\end{equation}
According to the interpretation of the rate reduction given in \citep{yu2020learning}, the above quantity precisely measures the volume of the space between $\Z_j$ and $\hat \Z_j$, illustrated as a pair of black and blue lines in Fig. \ref{fig:MCR2}. Then for the ``distance'' of all, say $k$, classes, we simply sum the rate reduction for all pairs: 
\begin{equation}
d(\Z, \hat \Z) \doteq \min_\eta  \sum_{j=1}^k \Delta R\big(\Z_j, \hat{\Z}_j\big) =  \min_\eta  \sum_{j=1}^k \Delta R\big(\Z_j, f(g(\Z_j, \eta),\theta)\big),
\label{eqn:min-distance}
\end{equation}
where $\Z_j = f(\X_j,\theta)$ and $\hat \Z_j = f(\hat{\X}_j,\theta)$. Obviously, a main goal of the learned decoder $g(\cdot, \eta)$ is to {\em minimize} the distance between these distributions. Notice that if the encoder $f$ preserves (i.e. injective for) the intrinsic structures of the original data $\X$,\footnote{This is typically the case for MCR$^2$-based feature representation \citep{yu2020learning}.} this criterion essentially aims to ensure there will be some decoded sample $\hat \x$ close to every data sample $\x$ -- hence the decoder $g$ should be ``surjective.'' According to the ideas of IMLE \citep{li2018implicit}, such a requirement could effectively help avoid mode-collapsing or mode-dropping.

\paragraph{Contrastive measure for the encoder.} For the {\em first} item in our wish-list, however, we normally do not have a natural metric or ``distance'' for similarity of samples or distributions in the original data space $\x$ for data such as images.\footnote{As mentioned before, finding proper metrics or distance functions on natural images has always been an elusive and challenging task \citep{image-similarity}.} To alleviate this difficulty, we can measure the similarity or difference between $\hat \X$ and $\X$ through their mapped features $\hat \Z$ and $ \Z$ in the feature space (again assuming $f$ is structure-preserving). If we are interested in discerning {\em any} difference in the distributions of the original and transcribed  samples, we may view the MCR$^2$ feature encoder $f(\cdot, \theta)$ as a ``discriminator'' to {\em magnify} any difference between all pairs of $\X_j$ and $\hat \X_j$, by simply maximizing, instead of minimizing, the {\em same quantity} in \eqref{eqn:min-distance}:
\begin{equation}
    d(\X, \hat \X) \doteq \max_{\theta} \sum_{j=1}^k \Delta R\big(\Z_j, \hat{\Z}_j\big) = \max_{\theta} \sum_{j=1}^k \Delta R\big(f(\X_j,\theta), f(\hat{\X}_j,\theta)\big).
    \label{eqn:max-distance}
\end{equation}
That is, a ``distance'' between $\X$ and $\hat \X$ can be measured as the maximally achievable rate reduction between all pairs of classes in these two sets. In a way, this measures how well or bad the decoded $\hat \X$ aligns with the original data $\X$ -- hence measuring the goodness of ``injectivity'' of the encoder $f$. Notice that such a discriminative measure is consistent with the idea of GAN \citep{goodfellow2014generative} that tries to separate $\X$ and $\hat \X$ into two classes, measured by the cross-entropy. Nevertheless, here the MCR$^2$-based discriminator $f$ naturally generalizes to cases when the data distributions are multi-class and multi-modal, and the discriminativeness is measured with a more refined measure -- the rate reduction, instead of the typical two-class loss (e.g. cross entropy) used in GANs. See Appendix \ref{app:ablation-objective} for the comparison with some ablation study.

\begin{figure}[t]
\centerline{\includegraphics[width=3.0in]{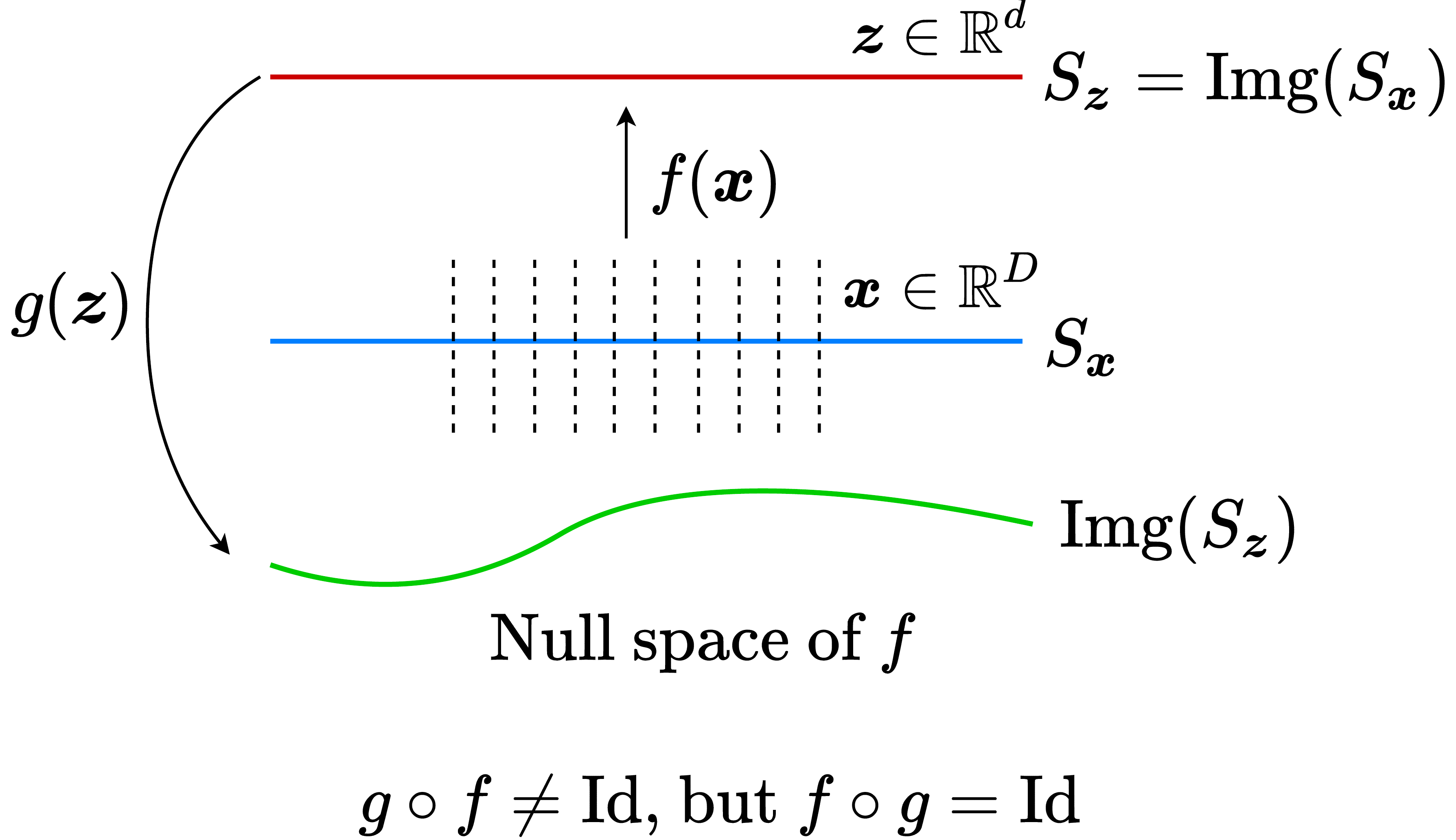}}
\caption{\textbf{Embeddings of Low-Dimensional Submanifolds in High-Dimensional Spaces.} $S_{\x}$ (blue) is the submanifold for the original data $\x$; $S_{\z}$ (red) is the image of $S_{\x}$ under the mapping $f$, representing the learned feature $\z$; and the green curve  is the image of the feature $\z$ under the decoding mapping $g$. } \label{fig:decoder}
\end{figure}

One may wonder the reason why we need the mapping $f(\cdot, \theta)$ to function as a discriminator between $\X$ and $\hat \X$ by maximizing $\max_\theta \Delta R\big(f(\X,\theta), f(\hat \X,\theta)\big)$? Figure \ref{fig:decoder} gives a simple illustration: there might be many decoders $g$ such that $f\circ g$ is an identity (Id) mapping\footnote{Here we use the notion of ``identity mapping'' in a loose sense: depending on the context, it could simply mean an embedding from $S_{\z}$ to $S_{\z}$.} $f\circ g(\z) = \z $ for all $\z$ in the subspace $S_{\z}$ in the feature space. However, $g\circ f$ is not necessarily an auto-encoding map for $\x$ in the original distribution $S_{\x}$ (here for simplicity drawn as a subspace). That is, $g\circ f(S_{\x}) \not\subset S_{\x}$, let alone $g\circ f(S_{\x}) = S_{\x}$ or $g\circ f(\x) = \x$. One should expect, without careful control of the image of $g$, with high probability this would be the case especially when the support of the distribution of $\x$ is extremely  low-dimensional in the original high-dimensional data space. For example, as we will see in the experiments, the intrinsic dimension of submanifold associated with each image category is about a dozen whereas images are embedded in a (pixel) space of thousands or tens of thousands of dimension.

\paragraph{Remark: representing the encoding and decoding mappings.} Some practical questions arise immediately: how rich the families of functions we should consider to use for the encoder $f$ and decoder $g$ that can optimize the above rate reduction type objectives?\footnote{Actually, similar questions exist for the formulation of GAN, regarding the realizability of the data distribution by the generator, see \citep{Nash-Gan}. Conceptually, here we know the encoder $f$ needs to be rich enough to discriminate (small) deviation from the true data support $\mathcal{M}_j$, while the decoder $g$ needs to be expressive enough to generate the data distribution from the learned mixture of subspace-Gaussians.} How should we represent or parameterize them, hence make our objectives computable and optimizable? For the most general cases, these remain widely open and challenging mathematical and computational problems. As we mentioned earlier, in this work, we will take a more pragmatic approach by simply representing these mappings with popular neural networks that have  empirically proven to be good at approximating distributions of practical (visual) datasets or for achieving the maximum of the rate reduction type objectives \citep{yu2020learning}. Nevertheless, our experiments indicate that our formulation and objectives are {\em not so sensitive} to particular choices in network structures or many of the tricks used to train them. Also, in the special cases when the real data distribution is benignly deformed from an LDR, the work of \cite{ReduNet} has shown that one can explicitly construct these mappings from the rate reduction objectives in the form of a deep network known as the ReduNet. However, it remains unclear how such constructions could be generalized to the closed-loop settings. Regardless, answers to these questions are beyond the scope of this work as our purposes here are mainly to verify empirically the validity of the proposed closed-loop data transcription framework.

\subsection{Encoding and Decoding as a Two-Player MiniMax Game}\label{sec:game-formulation}
Comparing the contractive and contrastive nature of \eqref{eqn:min-distance} and \eqref{eqn:max-distance} on the same utility, we see the roles of the encoder $f(\cdot, \theta)$ and the decoder $g(\cdot, \eta)$ naturally as ``{\bf a  two-player game}'': {\em while the encoder $f$ tries to magnify the difference between the original data and their transcribed; the decoder $g$ aims to minimize the difference.} Now for convenience, let us define the ``closed-loop encoding'' function:
\begin{equation}
    h(\x, \theta, \eta) \doteq f\big(g\big(f(\x, \theta), \eta\big), \theta\big): \; \x \mapsto \z.
\end{equation}
Ideally, we want this function to be very close to $f(\x, \theta)$ or at least the distributions of their images should be close. With this notation, combining \eqref{eqn:min-distance} and \eqref{eqn:max-distance}, a closed-loop notion of ``distance'' between $\X$ and $\hat \X$ can be computed as {\em an equilibrium point} to the following Min-Max program for the same utility in terms of rate reduction:
\begin{equation}
\mathcal{D}(\X, \hat \X) \doteq  \min_\eta \max_\theta \sum_{j=1}^k \Delta R\big(f(\X_j,\theta), h(\X_j,\theta,\eta)\big).
    \label{eq:MCR2-GAN-pair}
\end{equation}

Notice that this only measures the difference between (features of) the original data and its transcribed version. It does not measure how good the representation $\Z$ (or $\hat \Z$) is for the multiple classes within $\X$ (or $\hat \X$). To this end, we may combine the above distance with the original MCR$^2$-type objectives  \eqref{eq:mcr2-formulation}: namely, the rate reduction $\Delta R(\Z)$ and $\Delta R(\hat \Z)$ for the learned LDR $\Z$ for $\X$ and $\hat \Z$ for the decoded $\hat \X$. Notice that although the encoder $f$ tries to {\em maximize} the multi-class rate reduction of the features $\Z$ of the data $\X$,  the decoder $g$ should {\em minimize} the rate reduction of the multi-class  features $\hat \Z$ of the decoded $\hat \X$. That is, the decoder $g$ tries to use a minimal  coding rate needed to achieve a good decoding quality. 

Hence, the overall ``multi-class''  Min-Max program for learning the LDR transcription is:\footnote{In this work we only consider the simple case by adding these rate reduction quantities together. Of course, in the future one may consider other more delicate formulations. For instance, we may consider a  Min-Max game on the third term \eqref{eq:MCR2-GAN-pair}, subject to certain constraints (upper or lower bounds) on the first term and the second term. Such constrained minimax games have also started to draw attention lately \citep{dai2020optimality}.}
\begin{eqnarray}
\min_\eta \max_\theta  \mathcal{T}_{\X}(\theta, \eta) &\doteq&  \underbrace{\Delta R\big(f(\X,\theta)\big)}_{\text{Expansive encode}} + \underbrace{\Delta R\big(h(\X,\theta, \eta)\big)}_{\text{Compressive decode}} + \sum_{j=1}^k \underbrace{\Delta R\big(f(\X_j,\theta), h(\X_j,\theta,\eta)\big)}_{\text{Contrastive encode \& Contractive decode}} \nonumber \\
&=& \Delta R\big(\Z(\theta) \big) + \Delta R\big(\hat \Z(\theta, \eta)\big) + \sum_{j=1}^k \Delta R\big(\Z_j(\theta), \hat \Z_j(\theta, \eta) \big).
    \label{eq:MCR2-GAN-objective}
\end{eqnarray}
Empirically we have evaluated the necessity of these terms in an ablation study (see Appendix \ref{app:rate-reduction-terms}). Notice that, without the terms associated with the generative part $h$ or with all such terms fixed as constant, the above objective is precisely the original MCR$^2$ objective proposed by \cite{yu2020learning}.\footnote{In an unsupervised setting, if we view each sample (and its augmentations) as its own class, the above formulation remains exactly the same! The number of classes $k$ is simply the number of independent samples.} Also, notice that the minimax objective function depends only on (features of) the data $\X$ hence one can learn the encoder and decoder (parameters) without the need of sampling or matching any additional  distribution (as typically needed in GANs or VAEs)!

As a special case, if $\X$ only has one class, the above Min-Max program reduces\footnote{as the first two rate reduction terms automatically become zero.} to a special ``two-class'' or ``binary''  form between $\X$ and the decoded $\hat\X$:\footnote{Notice that this binary case resembles formulation of the original GAN  \eqref{eqn:GAN} by viewing $\X$ and $\hat \X$ as two classes $\{\bm 0, \bm 1\}$. Nevertheless, instead of using cross entropy, our formulation adopts a more refined rate reduction measure, which has been shown to promote diversity in the learned representation \citep{yu2020learning}.}
\begin{equation}
\mbox{Binary:} \quad
 \min_\eta \max_\theta \mathcal{T}^b_{\X}(\theta, \eta) \doteq \Delta R\big(f(\X,\theta), h(\X,\theta,\eta)\big) = \Delta R\big(\Z(\theta), \hat \Z(\theta, \eta)\big).
    \label{eq:MCR2-GAN-objective-binary}
\end{equation}
Sometimes, even when $\X$ contains multiple classes/modes, one could still view all classes together as one class. Then the above binary objective is to align the union distribution of all classes with their decoded $\hat \X$. This is typically a simpler task to achieve than the multi-class one \eqref{eq:MCR2-GAN-objective} since it does not require to learn a more refined multi-class LDR for the data, as we will later see in experiments. Notice that one good characteristic of the above formulation is that {\em all quantities in the objectives are measured in terms of rate reduction for the learned features} (assuming features eventually become subspace Gaussians). 

In all of our subsequent experiments, we solve the above minimax programs using the most basic gradient descent-ascent (GDA)  algorithm \citep{korpelevich1976extragradient} that alternates between the minimization and maximization, with the same learning rate and without any timescale separation (as typically needed for training GANs \cite{Fiez2020GradientDP}). Although more refined optimization schemes likely can further improve the efficiency and performance, we leave these for future investigation.

\paragraph{Remark: closed-loop error correction.} One may notice that our framework (see Figure \ref{fig:MCR2})  draws inspiration from closed-loop error correction widely practiced in feedback control systems. In the machine learning and deep learning literature, the idea of closed-loop error correction and closed-loop fixed point has been explored before to interpret the recursive error correcting mechanism and explain stability in a forward (predictive) deep neural network, for example the {\em deep equilibrium networks} \citep{equilibrium} and the {\em deep implicit networks} \citep{implicit}, again drawing inspiration from feedback control. Here in our framework, the closed-loop mechanism is not used to interpret the  encoding or decoding (forward) networks $f$ and $g$. Instead, it is used to form an overall feedback system between the two encoding and decoding networks for correcting the ``error'' in the distributions between the data $\x$ and the decoded $\hat \x$. Using terminology from control theory, one may view the encoding network $f$ as a ``sensor'' for error feedback while the decoding network $g$ as a ``controller'' for error correction. However, notice that here the ``target'' for control is not a scalar nor a finite dimensional vector, but a continuous mapping -- for the distribution of $\hat \x$ to match that of the data $\x$. This is in general a control problem in an infinite dimensional space.\footnote{The space of diffeomorphisms of submanifolds is infinite dimensional \citep{Lee2002IntroductionTS}.} Ideally, we hope when the sensor $f$ and the controller $g$ are optimal, the distribution of $\x$ becomes a ``fixed point'' for the closed loop while the distribution of $\z$ reaches a compact LDR. Hence the minimax programs \eqref{eq:MCR2-GAN-objective} and \eqref{eq:MCR2-GAN-objective-binary} can also be interpreted as games between an error-feedback sensor and an error-reducing controller.

\paragraph{Remark: relation to bi-directional or cycle  consistency.} The notion of ``bi-directional'' and ``cycle'' consistency between encoding and decoding has been exploited in the work of BiGAN \citep{donahue2016adversarial} and ALI \citep{dumoulin2016adversarially} for mappings between the data and feature and in the work of CycleGAN \citep{zhu2017unpaired} for mappings between two different data distributions. In our context, that is similar to promote $g\circ f$ and $f\circ g$ to be close to identity mappings (either for the distributions or for the samples). Interestingly, our new closed-loop formulation actually ``decouples'' the data $\X$, say observed from the external world, from their internally represented features $\Z$.  The objectives \eqref{eq:MCR2-GAN-objective} and \eqref{eq:MCR2-GAN-objective-binary} are functions of {\em only} the internal features $\Z(\theta)$ and $\hat \Z(\theta, \eta)$ which can be  learned and optimized by adjusting the neural networks $f(\cdot, \theta)$ and $g(\cdot, \eta)$ alone. There is no need of any additional external metrics or heuristics to promote how ``close'' the decoded images $\hat \X$ are to $\X$. This is very different from most VAE/GAN type methods such as BiGAN and ALI that require additional discriminators (networks) for the images and the features. Some experimental comparison are given in the Appendix \ref{app:MNIST}. In addition, in Appendix \ref{app:closed-loop} we provide some ablation study to illustrate the importance and benefit of a closed loop for enforcing the consistency between the encoder and decoder.

\paragraph{Remark: transparent versus hidden distribution of the learned features.} Notice that in our framework, there is no need to explicitly specify a prior distribution either as a target distribution to map to for AE \eqref{eqn:auto-encoding} or as an initial distribution to sample from for GAN \eqref{eqn:GAN}. The common practice in AEs or GANs is to specify the prior distribution as a generic Gaussian. This is however particularly problematic when the data distribution is multi-modal and has multiple low-dimensional structures which is commonplace for multi-class data. In this case, the common practice in AEs or GANs is to train a conditional GAN for different classes or different attributes. However, here we only need to assume the desired target distribution belonging to the family of LDRs. The specific  optimal distribution of the feature within this family is then learned from the data directly and then can be represented {\em explicitly}, as a mixture of independent subspace Gaussians (or equivalently, a mixture of PCAs on independent subspaces). We will give more details in the experimental Section \ref{sec:experiments} as well as more examples in Appendix \ref{app:MNIST} to  \ref{app:CIFAR}. Although many GAN+VAE type methods can learn a bidirectional encoding and decoding mappings, the distribution of the learned features inside the feature space remains {\em hidden} or even {\em entangled}. This makes it difficult to sample the feature space for generative purposes or to use the features for discriminative tasks.\footnote{For instance, typically one can only use so-learned features for nearest neighbor type classifiers \citep{donahue2016adversarial}, instead of nearest subspace as we do in this work, see Section \ref{sec:use-LDR}.} 

\section{Empirical Verification on Real-World Imagery Datasets}
\label{sec:experiments}
This experiment section serves three purposes: First, we empirically justify the proposed formulation for data transcription by demonstrating good properties of the learned encoder, decoder, and representation tuple $(f, g, \z)$ from $\X$. Second, we compare our method with several representative methods from the GAN family and VAE family. The purpose of the comparison is {\em not} to compete for any state of the art performance. Instead, we want to convincingly verify the validity of the proposed framework and its potential in going beyond. Finally, we evaluate the so learned LDR through both generative tasks (controlled visualization) and discriminative (classification) tasks. More extensive experimental results, evaluations, and ablation studies can be found in the Appendix.

\paragraph{Datasets.} We provide extensive qualitative and quantitative experimental results on the following datasets: MNIST \citep{lecun1998gradient}, CIFAR-10 \citep{krizhevsky2009learning}, STL-10 \citep{coates2011analysis}, CelebA \citep{liu2015faceattributes}, LSUN bedroom \citep{yu2015lsun}, and ImageNet ILSVRC 2012 \citep{russakovsky2015imagenet}. The network architectures and implementation details can be found in Appendix~\ref{app:settings} and corresponding appendix section for each dataset.

\subsection{Empirical Justification of \ours{} Transcription}
To empirically validate our new framework, we conduct experiments from a small low-variety dataset (MNIST), to a small dataset of diverse real-world objects (CIFAR-10), to higher resolution images (STL-10, CelebA, LSUN-bedroom), to a large-scale diverse image set (ImageNet). The results are evaluated both quantitatively and qualitatively. Implementation details, more experimental results, and ablation studies  are given in the Appendix.

\begin{figure}[t]
\centering
 \subfigure[MNIST]{
     \includegraphics[width=0.30\textwidth]{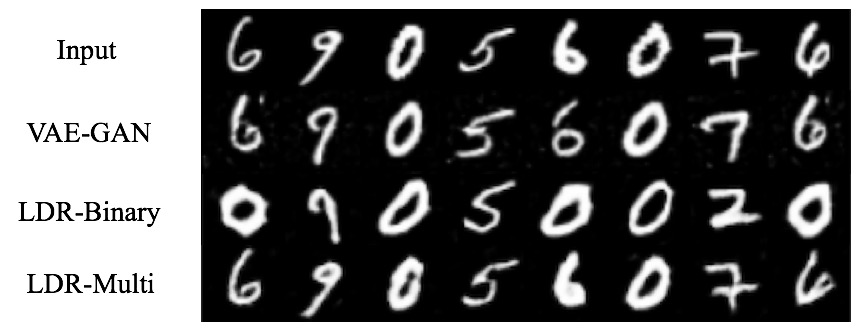}
 }
 \subfigure[CIFAR-10]{
     \includegraphics[width=0.29\textwidth]{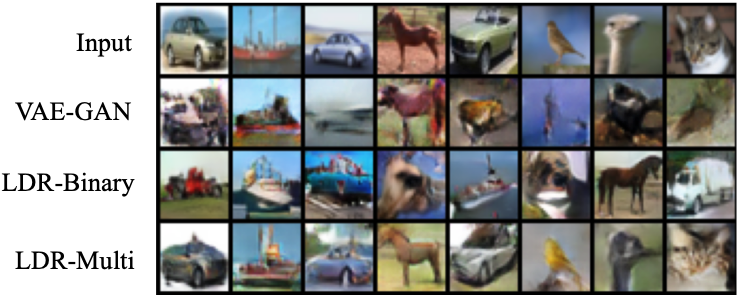}
 }
 \subfigure[ImageNet]{
     \includegraphics[width=0.36\textwidth]{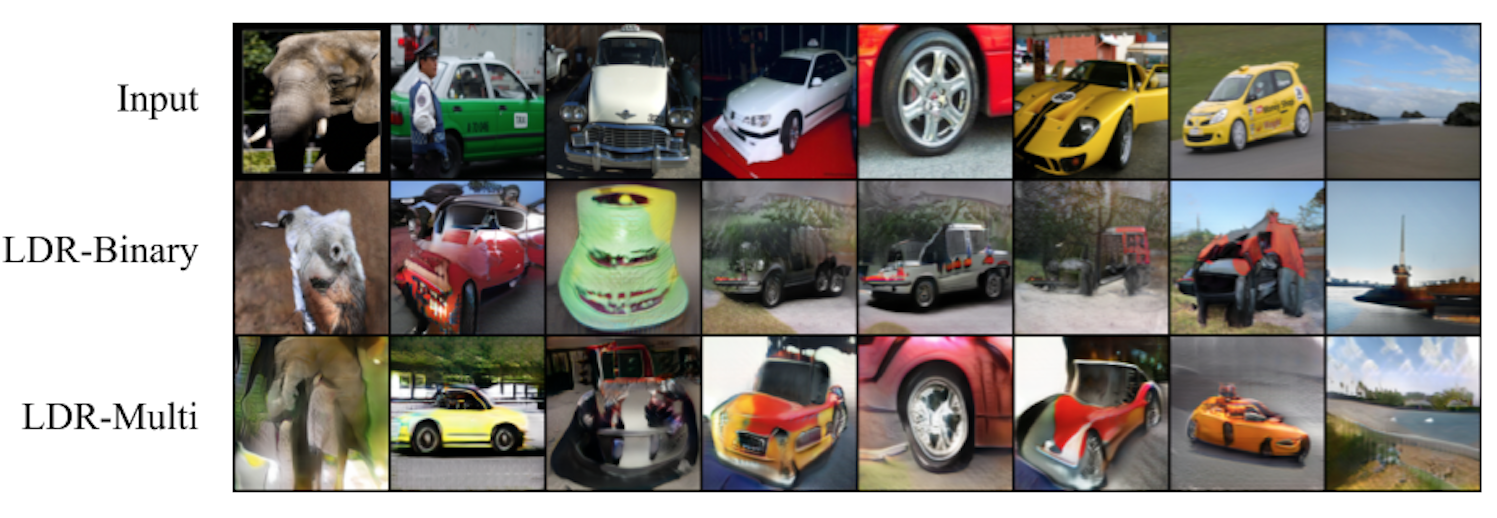}
} 
\caption{Qualitative comparison on MNIST, CIFAR-10 and ImageNet. First row: original $\X$; Other rows: reconstructed $\hat \X$ for different methods.} \label{fig:ablation_qualitative}
\end{figure}

\begin{table}[t]
    \centering
    \small
    \setlength{\tabcolsep}{3.5pt}
    \renewcommand{\arraystretch}{1.25}
    \begin{tabular}{cc|ccccc}
    \multicolumn{2}{c|}{Method}               & GAN   & GAN (\ours{}-Binary) & VAE-GAN & \ours{}-Binary & \ours{}-Multi \\
    \hline
    \hline
    \multirow{2}{*}{MNIST} & IS $\uparrow$    &  2.08 & 1.95  & \textbf{2.21} & 2.02  & 2.07  \\
    ~                      &FID $\downarrow$  & 24.78 & 20.15 & 33.65& \textbf{16.43} & 16.47 \\
    \hline
    \multirow{2}{*}{CIFAR-10} & IS $\uparrow$ & 7.32  & 7.23  & 7.11 & \textbf{8.11}  & 7.13  \\
    ~                      &FID $\downarrow$  & 26.06 & 22.16 & 43.25& \textbf{19.63} & 23.91 \\
    \end{tabular}
    \caption{Quantitative comparison on MNIST and CIFAR-10. Average Inception scores (IS) \citep{salimans2016improved} and FID scores \citep{heusel2017gans}. $\uparrow$ means higher is better. $\downarrow$ means lower is better.}
    \label{tab:ablation_quantitative}
\end{table}

\textbf{Comparison (IS and FID) with other formulations.} First, we conduct five  experiments to fairly compare our formulation with GAN \citep{radford2015unsupervised} and VAE(-GAN) \citep{larsen2016autoencoding} on MNIST and CIFAR-10. Except for the objective function, everything else is exactly the same for all methods (e.g. networks, training data, optimization method). These experiments are: 1). GAN; 2). GAN with its objective replaced by that of the LDR-Binary (\ref{eq:MCR2-GAN-objective-binary}); 3). VAE-GAN ; 4). Binary \ours{} \eqref{eq:MCR2-GAN-objective-binary}; and 5). Multi-class \ours{} \eqref{eq:MCR2-GAN-objective}. Some visual comparison is given in Figure. \ref{fig:ablation_qualitative}. IS \citep{salimans2016improved} and FID  \citep{heusel2017gans} scores are summarized in Table \ref{tab:ablation_quantitative}.\footnote{Here for simplicity, we have chosen a uniform feature dimension $d=128$ for all datasets. If we choose a higher feature dimension, say $d=512$, for the more complex CIFAR-10 dataset, the visual quality can be further improved, see Table \ref{tab:ablation_zdim} in Appendix \ref{app:feature-dim}.} 

As we see from the above Table~\ref{tab:ablation_quantitative}, replacing cross-entropy with the Equation~\eqref{eq:MCR2-GAN-objective-binary} can improve the generative quality. The two LDR formulations are clearly on par with the others in terms of IS and significantly better in FID. Finally, with the same training datasets, quality of LDR-Multi is lower than LDR-Binary. This is expected as the multi-class task is more challenging. Nevertheless, as we will see soon, images decoded by LDR-Multi align much better with their classes than Binary.

\begin{figure}[t]
\subfigure[MNIST]{
    \includegraphics[height=3.5cm]{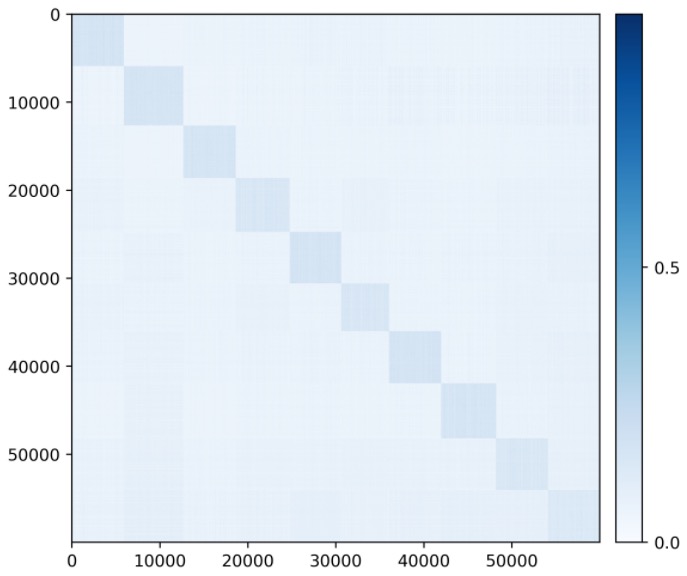}
}
\subfigure[CIFAR10]{
    \includegraphics[height=3.5cm]{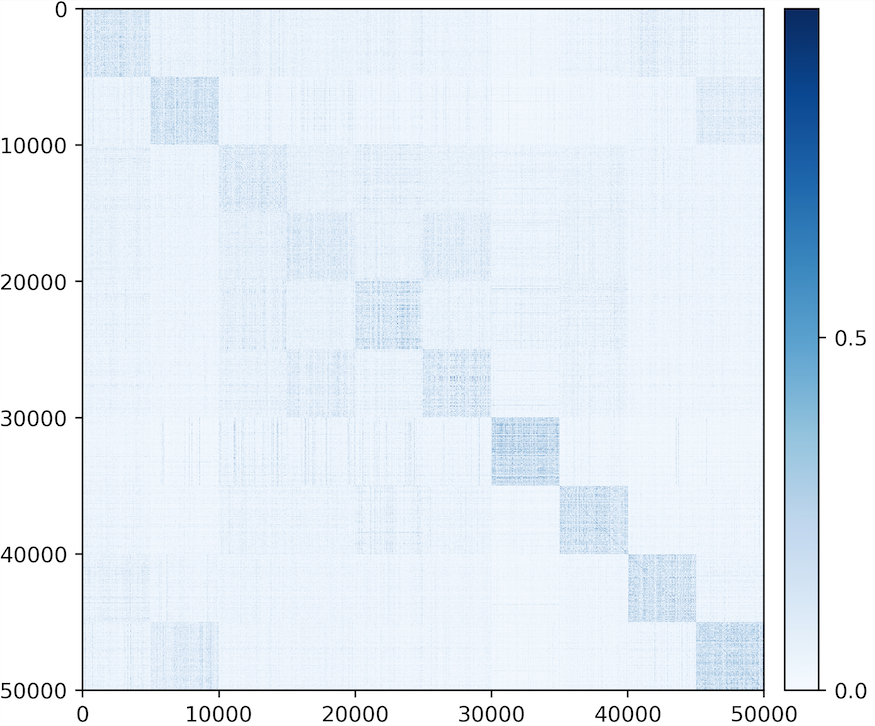}
}
\subfigure[ImageNet]{
    \includegraphics[height=3.5cm]{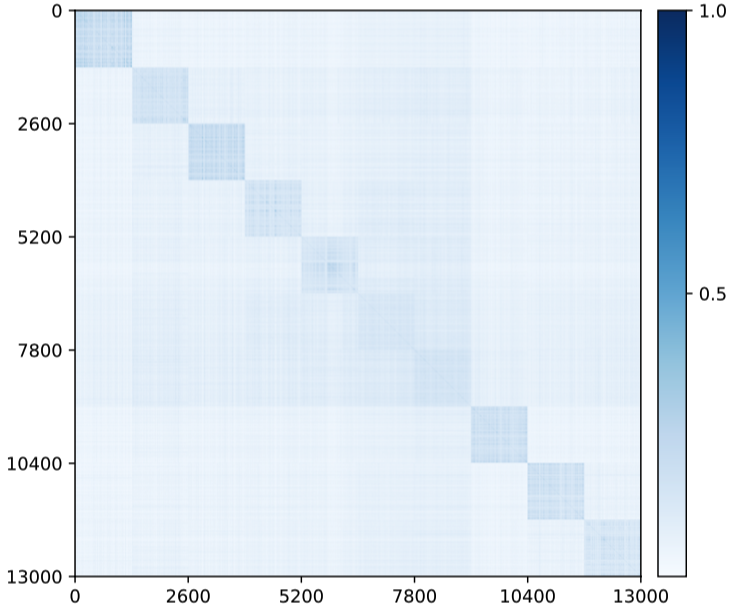}
}
\caption{Visualizing the alignment between $\Z$ and $\hat{\Z}$: $|\Z^\top \hat{\Z}|$ and  in the feature space for (a) MNIST, (b) CIFAR-10, and (c) ImageNet-10-Class.}
\label{fig:justifyz=z}
\end{figure}

\paragraph{Visualizing correlation of features $\Z$ and decoded features $\hat \Z$.} We visualize the cosine similarity between $\Z$ and $\hat{\Z}$ learned from the multi-class objective \eqref{eq:MCR2-GAN-objective} on MNIST, CIFAR-10 and ImageNet (10 classes), which indicates how close $\hat{\z} = f\circ g(\z)$ is from $\z$. Results in Figure \ref{fig:justifyz=z} show that $\Z$ and $\hat{\Z}$ are aligned very well within each class. The block-diagonal patterns for MNIST are sharper than those for CIFAR-10 and ImageNet, as images in CIFAR-10 and ImageNet have more diverse visual appearances.

\begin{figure}[t]
     \footnotesize
     \centering
     \subfigure[MNIST $\X$]{
         \includegraphics[trim=2.5cm 2cm 2cm 2.5cm ,clip,width=0.15\textwidth]{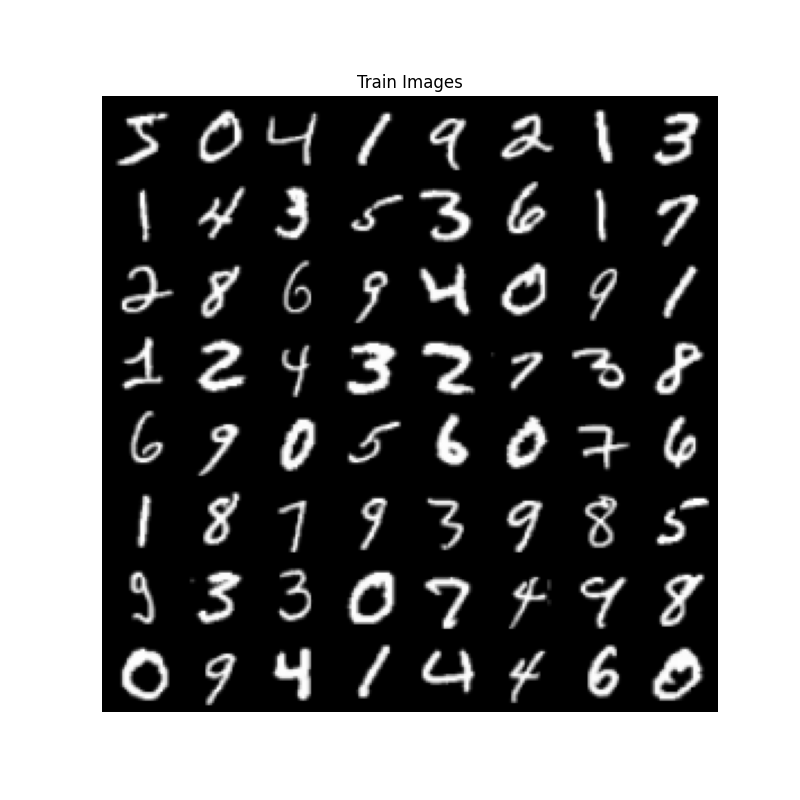}
     }
     \subfigure[MNIST $\hat{\X}$]{
         \includegraphics[trim=2.5cm 2cm 2cm 2.5cm ,clip,width=0.15\textwidth]{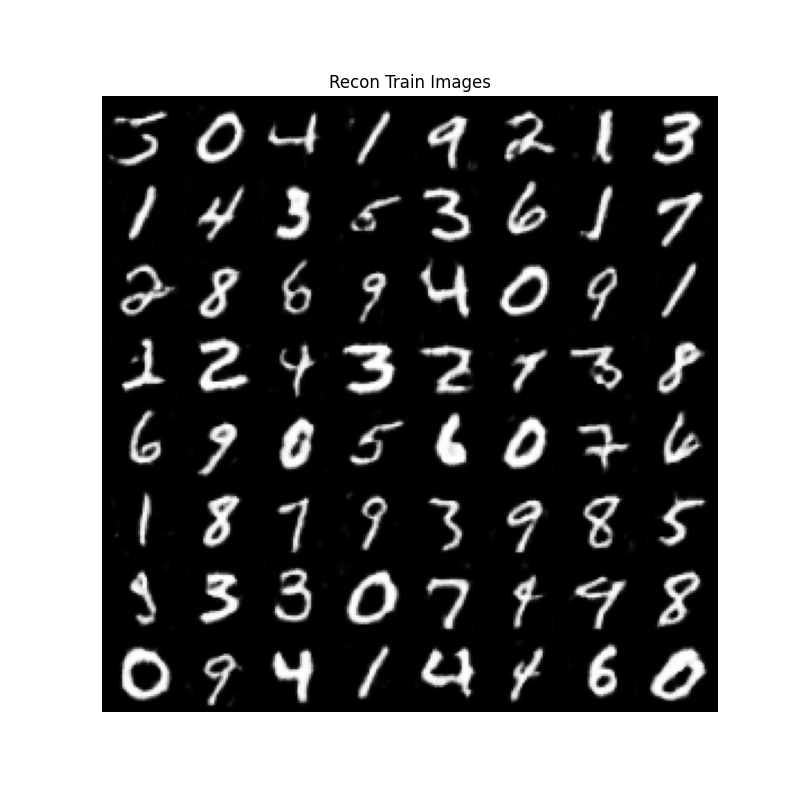}
     }
     \subfigure[CIFAR-10 $\X$]{
         \includegraphics[width=0.15\textwidth]{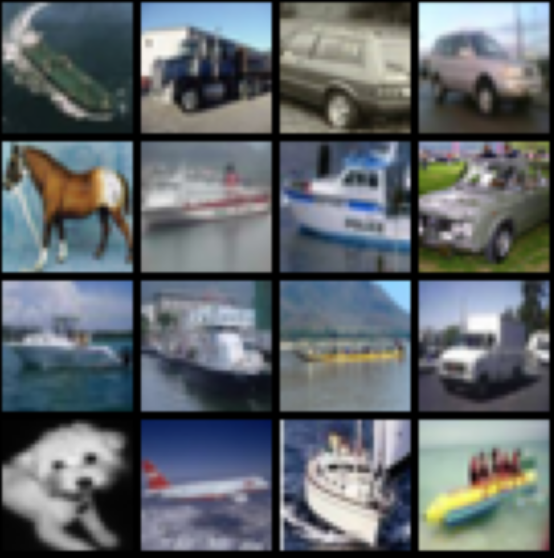}
     }
     \subfigure[CIFAR-10 $\hat{\X}$]{
         \includegraphics[width=0.15\textwidth]{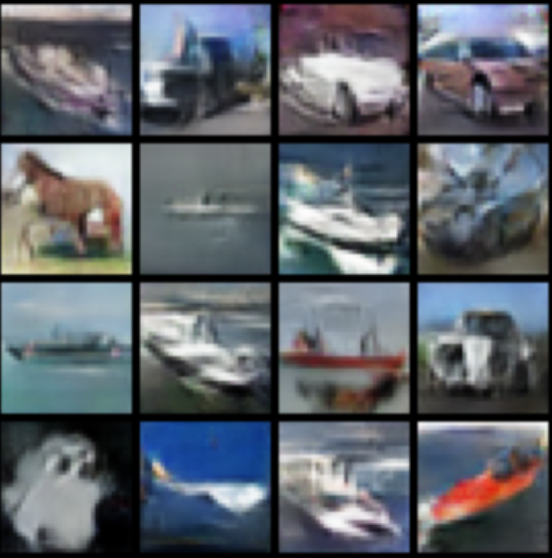}
     }
     \subfigure[ImageNet $\X$]{
         \includegraphics[width=0.15\textwidth]{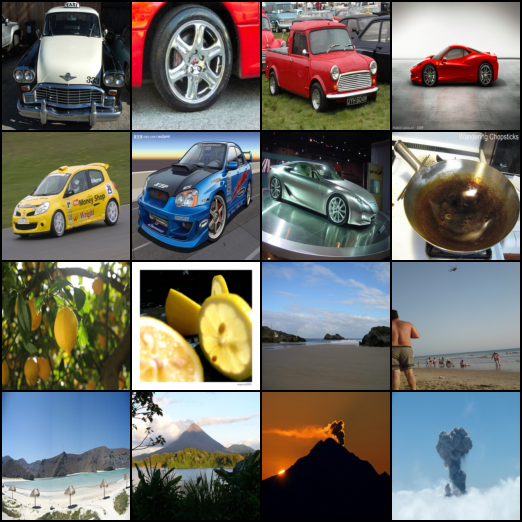}
     }
     \subfigure[ImageNet $\hat{\X}$]{
         \includegraphics[width=0.15\textwidth]{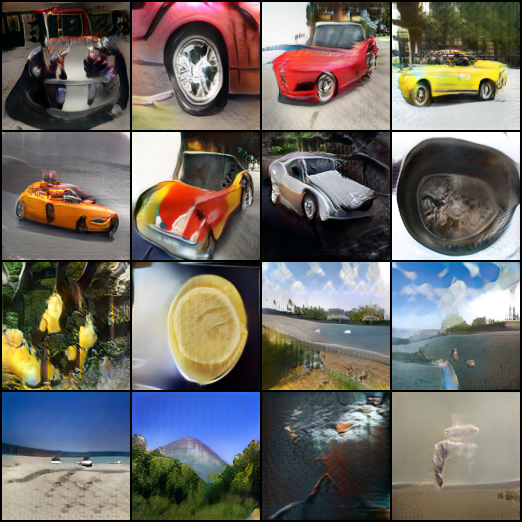}
    }
    \caption{Visualizing the auto-encoding property of the learned \ours{} ($\x \approx \hat{\x} = g\circ f(\x)$) on MNIST, CIFAR-10, and ImageNet (zoom in for better visualization).}
        \label{fig:justifyx=x}
\end{figure}

\paragraph{Visualizing auto-encoding of the data $\X$ and the decoded $\hat \X$.} We compare some representative $\X$ and $\hat{\X}$ on MNIST, CIFAR-10 and ImageNet (10 classes) to verify how close $\hat \x = g\circ f(\x)$ is to $\x$. The results are shown in Figure \ref{fig:justifyx=x}, and visualizations are created from training samples. Visually, the auto-encoded $\hat \x$ faithfully captures major visual features from its respective training sample $\x$, especially the pose, shape, and layout. For the simpler dataset such as MNIST, auto-encoded images are almost identical to the original! The visual quality is clearly better than other GAN+VAE type methods, such as  VAE-GAN \citep{VAE-GAN} and BiGAN \citep{donahue2016adversarial}. We refer the reader to Appendix \ref{app:MNIST}, \ref{app:CIFAR}, and \ref{app:imagenet} for more visualization of results on these datasets, including similar results on transformed MNIST digits. More visualization results for learned models on real-life image datasets such as STL-10, CeleB, and LSUN can be found in the Appendix \ref{app:stl-10}-\ref{app:celeba_lsun}.

\subsection{Comparison to Existing Generative Methods}
Table~\ref{tab:comparsion_partial} gives a quantitative comparison of visual quality of our method with others on CIFAR-10, STL-10, and ImageNet. In general, there is a large difference in terms of FID and IS scores between the GAN family and the VAE family of models. SNGAN \citep{miyato2018spectral} are commonly used methods in most generative applications while LOGAN \citep{wu2019logan} is the state-of-the-art method on ImageNet in terms of FID and IS.  More comparisons with existing methods,  including results on on the higher-resolution ImageNet dataset,  can be found in Table \ref{tab:comparsion_full} of the Appendix \ref{app:imagenet}. 

As we see, even if the rate reduction objectives \eqref{eq:MCR2-GAN-objective} and \eqref{eq:MCR2-GAN-objective-binary} are not specifically designed nor engineered for visual quality\footnote{In our current implementation, the original objectives are used without any other heuristics or regularization.} and the networks and hyper-parameters adopted in our experiments are rather basic compared to many of the state of the art generative methods, our method is still rather competitive in terms of these metrics. The simplicity of our framework and formulation suggests that there is significant room for further improvement. For instance, in all experiments on all datasets, we have chosen a feature dimension of $d=128$ for simplicity and uniformity. In the last Appendix \ref{app:feature-dim}, we have conducted an ablation study on using a higher feature dimension $d =512$. The visual quality of the learned model can be significantly improved (as shown in Figure \ref{ablation::zdim_viz} and Table \ref{tab:ablation_zdim} of Appendix \ref{app:feature-dim}).

In fact, compared to these methods, our method has learned not just any generative model. It has learned a {\em structured} generative model that has many additional beneficial properties that we now present.

\begin{table}[t]
    \centering
    \small
    \setlength{\tabcolsep}{3.0pt}
    \renewcommand{\arraystretch}{1.25}
    \begin{tabular}{ll|ccc|ccccc}
    \multicolumn{2}{c|}{Method} & \multicolumn{3}{c|}{GAN based methods} & \multicolumn{5}{c}{VAE/GAN based methods} \\
    ~ & ~                           & SNGAN & CSGAN & LOGAN       & VAE-GAN & NVAE & DC-VAE & \ours{}-Binary  & \ours{}-Multi  \\
    \hline
    \multirow{2}{*}{CIFAR-10} & IS $\uparrow$  &  7.4  & 8.1   &\textbf{8.7} & 7.4     & -    & \textbf{8.2}&  \textbf{8.1} & 7.1 \\
    ~                         &FID $\downarrow$  &  29.3 & 19.6  &\textbf{17.7}& 39.8    & 50.8 & \textbf{17.9}& \textbf{19.6} & 23.9 \\
    \hline
    \multirow{2}{*}{STL-10}   & IS $\uparrow$  &  \textbf{9.1}  & -     &-            & -       & -    & 8.1 & 8.4 & 7.7 \\
    ~                         &FID $\downarrow$  & 40.1 & -     &-            & -       & -    & 41.9& \textbf{38.6} & 45.7 \\

    \end{tabular}     
    \caption{Comparison on CIFAR-10 and STL-10. Comparison with more existing methods and on ImageNet can be found in Table ~\ref{tab:comparsion_full} in the Appendix.}
    \label{tab:comparsion_partial}
\end{table}

\subsection{Benefits of the Learned LDR Transcription Model}
\label{sec:use-LDR}
As we have argued before, the learned LDR transcription model (including the feature $\z$, the encoder $f$, and the decoder $g$) can be used for both generative and discriminative purposes. In particular, unlike almost all existing generative methods, the internal structures or distribution of the learned $z$ are no longer ``hidden'' as they have clear subspace structures. Hence we can easily derive an explicit (parametrizable) model for the distribution of the learned features as a mixture of independent subspace-like Gaussians. This gives us full control in sampling the learned distribution for generative purposes.

\paragraph{Principal subspaces and principal components for the feature.} To be more specific, given the learned $k$ class features $\cup_{j=1}^k\Z_j$ for the training data,  we have observed that the leading singular subspaces for different classes are all approximately orthogonal to each other: $\Z_i \perp \Z_j$ (see Figure \ref{fig:justifyz=z}). This corroborates with our above discussion about the theoretical properties of the rate reduction objective. They essentially span $k$ independent principal subspaces. We can further calculate the mean  $\Bar{\z}_j$\footnote{Although we conceptually view the support of each class is a subspace, the actual support of the features is close to be on the sphere due to feature (scale) normalization. Hence, it is more precise to find its mean and its support centered around the mean.} and the singular vectors $\{\bm v_{j}^{i}\}_{i=1}^{r_j}$ (or principal components) of the learned features $\Z_j$ for each class. Here $r_j$ is a rank we may choose to model the dimension of each principal subspace (say based on a common threshold on the singular values). Hence, we obtain an explicit model for how the feature $\z$ is distributed in each of the $k$ principal subspaces in the feature space $\Re^d$: 
\begin{equation}
    \z_j \sim \bar{\z}_j +   \sum_{l=1}^{r_j} n_l^j \sigma_{j}^l \bm v_j^l, \quad \mbox{where} \quad n_l^j \sim \mathcal{N}(0, 1), \;\; j = 1, \ldots, k.
    \label{eqn:Gaussian-models}
\end{equation}
Hence this essentially gives an explicit mixture of subspace-like Gaussians model for the learned features: statistical differences between different classes are modeled as $k$ independent principal subspaces; statistical differences within each class $j$ are modeled as $r_j$ independent principal components in the $j$th subspace. 

\begin{figure}[t]
\centering
 \subfigure[Horse]{
     \includegraphics[width=0.30\textwidth]{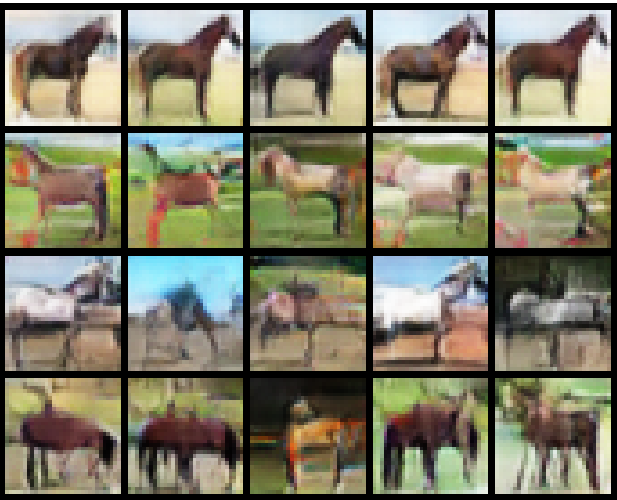}
 }
 \hspace{5mm}
  \subfigure[Ship]{
     \includegraphics[width=0.30\textwidth]{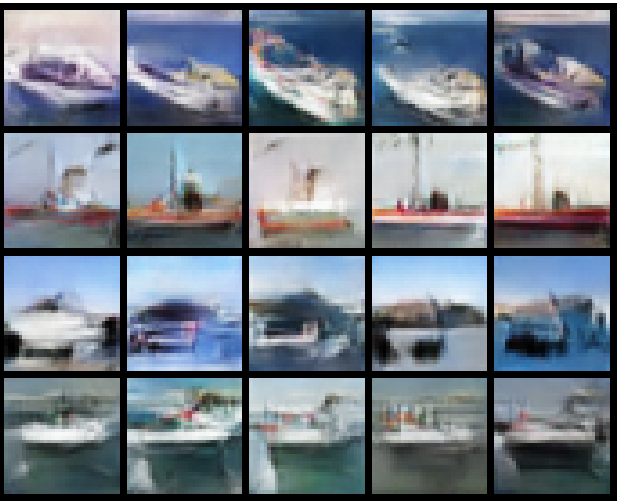}
 }
 \caption{\textbf{CIFAR-10 dataset.} Visualization of top-5 reconstructed $\hat \x=g(\z)$ based on the closest distance of $\z$ to each row (top-4) principal components of data representations for class 7-‘Horse’ and class 8-‘Ship’.}
 \label{fig:cifar_10_pca_sampling_main}
\end{figure}

\begin{figure}[t]
\centering
 \subfigure[Disentangled attributes as principal components]{
     \includegraphics[width=0.5\textwidth]{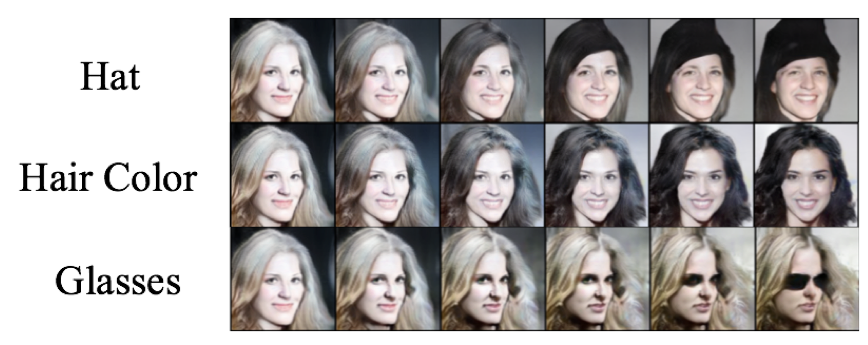}
     \label{fig:sampling_along_pca_img_a}
 }\hspace{1mm}
  \subfigure[Interpolation between distinct samples]{
     \includegraphics[width=0.38\textwidth]{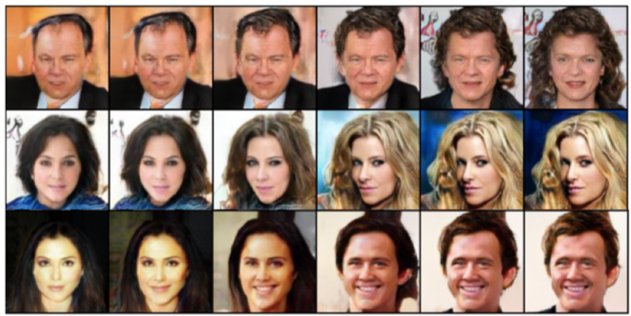}
     \label{fig:sampling_along_pca_img_b}
 }
 \caption{\textbf{CelebA dataset.} (a): Sampling along three principal components that seem to correspond to different visual attributes; (b): Samples decoded by interpolating along the line between features of two distinct samples.}
 \label{fig:sampling_along_pca_img}
\end{figure}

\paragraph{Decoding samples from the feature distribution.} Using the CIFAR-10 and CelebA datatsets, we visualize images decoded from samples of learned feature subspace. For the CIFAR-10 dataset, for each class $j$, we first compute the top-4 principal components of the learned features $\Z_j$ (via SVD). For each class $j$, we then compute $|\langle\z^i_j, \bm{v}^l_j\rangle|$, the cosine similarity between the $l$-th principal direction $\bm v^l_j$ and feature sample $\z_j^i$. After finding the top-5 $\z^i_j$ according to $|\langle\z^i_j, \bm  v^l_j\rangle|$ for each class $j$, we reconstruct images $\hat\x_j^i=g(\z^i_j)$. Each row of Figure~\ref{fig:cifar_10_pca_sampling_main} is for one principal component. We observe that images in the same row share the same visual attributes; images in different rows differ significantly in visual characteristics like shape, background, and style. See Figure \ref{fig:CIFAR10_PCA} of Appendix \ref{app:CIFAR} for more visualization of principal components learned for all 10 classes of CIFAR-10. These results clearly demonstrate the principal components in each subspace-Gaussian disentangle different visual attributes. In addition, we do not observe any mode dropping for any of the classes although the dimensions of the classes were not known {\em a priori}.

\paragraph{Disentangled visual attributes as principal components.} For the CelebA dataset, we calculate the principal components of all learned features in the latent space. Figure~\ref{fig:sampling_along_pca_img_a} shows some decoded images along these principal directions. Again, these principal components seem to clearly {\em disentangle} visual attributes/factors such as wearing a hat, changing hair color, and wearing glasses. More examples can be found in Appendix \ref{app:celeba_lsun}. The results are consistent with {\em the property of MCR$^2$ that promotes diversity of the learned features}. 

\paragraph{Linear interpolation between features of two distinct samples.} Figure~\ref{fig:sampling_along_pca_img_b} shows interpolating features between pairs of training image samples of the CeleA dataset, where for two training images $\x_1$ and $\x_2$, we reconstruct based on their linearly interpolated feature representations by $\hat\x = g(\alpha f(\x_1)+(1-\alpha)f(\x_2)), \alpha \in[0, 1]$. The decoded images show continuous morphing from one sample to another in terms of visual characteristics, as opposed to merely a superposition of the two images. Similar interpolation results between two digits in the MNIST dataset can be found in Figure \ref{fig:mnist_interpolation} of the Appendix \ref{app:MNIST}.

\paragraph{Encoded features for classification.} Notice that not only is the learned decoder good for generative purpose, but also the encoder is good for discriminative tasks. In this experiment, we evaluate the discriminativeness of the learned LDR model by testing how well the encoded features can help classify the images. We use features of the training images to compute the learned subspaces for all classes, then classify features of the test images based on a simple  nearest subspace classifier.\footnote{While many other encoding methods  train a classifier (say with an additional layer)  after the learned features.}  Results in Table~\ref{tab:represent_cls} show that our model gives competitive classification accuracy on MNIST, compared to some of best VAE-based methods. We also tested the classification on CIFAR-10, the accuracy is currently  about $80.7\%$. As expected, the representation learned with the multi-class objective is very discriminative and good for classification tasks.\footnote{Be aware that all generative models, GANs, VAEs, and ours, are not specifically engineered for classification tasks. Hence one should not expect the classification accuracy to compete with supervised-trained classifiers yet.} This demonstrates the learned LDR model is not only generative but also discriminative.

\begin{table}[t]
    \centering
    \setlength{\tabcolsep}{6.5pt}
    \renewcommand{\arraystretch}{1.25}
    \begin{tabular}{l|c|c|c|c|c|c}
    Method   & VAE   & Factor VAE & Guide-VAE & DC-VAE & LDR-Binary & LDR-Multi     \\
    \hline
    \hline
    MNIST    & 97.12\% & 93.65\%      & 98.51\%     & 98.71\%  & 89.12\%      & 98.30\%         \\  
    \end{tabular}      
    \caption{Classification accuracy on MNIST, comparing to classifier based VAE methods \citep{parmar2021dual}. Most of those VAE-based methods require auxiliary classifiers to boost  classification performance.}
     \label{tab:represent_cls}
\end{table}

\section{Open Theoretical Problems}\label{sec:discussions} 
So far, we have given theoretical intuition and derivation for the formulation of LDR transcription, as well as empirical evidence to showcase both the performance and potential of LDR transcription. In this section, we take a step back to explore the theoretical underpinnings of LDR transcription. We organize this section by discussing three primary objectives associated with LDR transcription:

\begin{enumerate}
    \item \emph{Learn a simple linear discriminative representation $f(\X)$ of the data $\X$}, which we can reliably use to classify the data.
    \item \emph{Learn a reconstruction $g\circ f(\X) \sim \X$ of the so learned representation $f(\X)$}, to ensure consistency in the representation. 
    \item \emph{Learn both representation and reconstruction in a closed-loop manner}, using feedback from the encoder $f$ and decoder $g$ to jointly solve the above two tasks.
\end{enumerate}

These three objectives encompass the overarching principle of LDR transcription, and indeed each of these objectives ties to a wide array of mathematical and theoretical problems. We now outline some of the most important theoretical questions or hypotheses implicated by our results, which we leave for future work to study and to answer, likely by a broader range of research communities.

\subsection{Distributions of the LDR representation}

Our primary mode of optimizing for a ``simple representation'' is through the \ours{} framework proposed in \citep{ReduNet}. One important open theoretical problem is finding the right energy function to optimize in order to promote \ours{}. It was shown in \citep{ReduNet} that an LDR can be learned for the multi-class data by maximizing the MCR$^2$ objective $\Delta R(\Z)$ given in \eqref{eq:mcr2-formulation}. This motivates the first two terms in our objective function  \eqref{eq:MCR2-GAN-objective}: maximizing $\Delta R(\Z), \Delta R(\hat{\Z})$ promotes their representations to be \ours{}.

Although \citep{ReduNet} has shown the MCR$^2$ objective can promote the features learned to be in orthogonal subspaces and has characterized the optimal second moments of the distributions, there remain open questions regarding what are the optimal distributions within the subspaces. A standing hypothesis is that the optimal distributions should be Gaussians. There is indeed already theoretical work on similar energy functions: the Brascamp-Lieb inequalities \citep{Brascamp-Lieb-1}, where the authors study a functional which, in certain contexts, is maximized uniquely by Gaussians. Hence an important future theoretical direction for LDR transcription is to exactly characterize distributional properties of the extremals (both minima and maxima) of the MCR$^2$ objective or its variants. Such results can further justify the use of Gaussian models \eqref{eqn:Gaussian-models} to characterize the learned features within the subspaces. 

We also notice that the so learned LDR features have additional striking properties: as shown by examples in Figure \ref{fig:sampling_along_pca_img}, distinctive visual attributes of the imagery data seem to be clearly disentangled by different principal components of the distribution; and along each principal direction one can linearly interpolate the features whereas the original data are nonlinear and cannot be directly interpolated. These results go beyond the guarantees given by \citep{ReduNet}, and an open theoretical problem is studying just how \ours{} transcription learns to disentangle and linearize such visual attributes. This understanding is crucial to extend the \ours{} transcription framework beyond the 2D vision domain.

\subsection{Consistency in the Learned Reconstruction}

If the learned encoder $\Z = f(\X)$ is an embedding of the data submanifolds to the subspaces, it should admit an inverse (decoding) mapping $\hat{\X} = g(\Z)$. As distributional distance in the data space is hard to come by, the rate reduction $\Delta R\big(\Z, \hat \Z\big)$ gives a well-defined distribution distance between $\Z$ and $\hat{\Z}$ which is used to enforce similarity between $\X$ and $\hat{\X}$ in our formulation. Notice that, unlike the KL-divergence or the JS-divergence, the rate reduction is well-defined and easily computable in closed-form between  mixtures of (degenerate) Gaussians. The third term of eq. \eqref{eq:MCR2-GAN-objective}, $\sum_{j=1}^k \Delta R\big(\Z_j(\theta), \hat \Z_j(\theta, \eta) \big)$, is exactly this distributional distance, which is minimized only when the estimated second moments of $\Z_j$ and $\hat{\Z}_j$ are the same. While this distributional distance seems weaker than sample-wise $\ell^2$-distance, we observe strong reconstruction performance nevertheless. 

Notice that the current objectives \eqref{eq:MCR2-GAN-objective} or \eqref{eq:MCR2-GAN-objective-binary} do not impose any constraints on the mappings of individual samples. That is, they do not explicitly specify how an individual sample $\x$ should be related to its decoded version $\hat \x = g(f(\x))$, or how their corresponding features $\z$ and $\hat \z$ are related. Hence, theoretically, nothing is known about relationships between individual samples and their features. But somewhat surprisingly, experimental results with the multi-class objective \eqref{eq:MCR2-GAN-objective} in next section suggest that they actually can be rather close, at least for the given  training samples $\X$! For example, see Figure \ref{fig:justifyx=x}. Of course, one could consider explicitly imposing certain sample-wise requirements in the objectives, such as enforcing $\x^i$ to be close to $\hat \x^i = g(f(\x^i))$. It has been observed that empirically in GANs or VAEs that imposing such sample-wise similarity or dissimilarity would improve visual quality around samples of interest, such as the DC-VAE \citep{parmar2021dual} and the OpenGAN \citep{Open-GAN}. But theoretically, it remains as an open problem how such sample-wise distances or constraints may affect the difficulty or accuracy in learning the correct support and density of the distributions.

\subsection{Properties of the Closed-loop Minimax Game}

Above are the two primary objectives for \ours{} transcription: while the encoder $f$ tries to maximize the expressiveness and discriminativeness of the learned LDR representation, the decoder $g$ tries to minimize the reconstruction error and coding rates. The competing objectives of the encoder $f$ and the decoder $g$ naturally lead to a two-player game. In this paper,  we have formulated this game as a zero-sum game, namely eq. \eqref{eq:MCR2-GAN-objective}. Likewise, we have also implemented the most straightforward algorithm for solving this zero-sum game: gradient descent-ascent (GDA) \citep{korpelevich1976extragradient}, where the minimizer and maximizer take alternating gradient steps.\footnote{These simplifications into a GDA-optimized zero-sum game were made in order to create a concrete algorithm for our experimentation. But simplifying to a zero-sum game and GDA is certainly not the only way to solve the more general game described above.} This game-theoretic formulation puts \ours{} transcription outside of the theoretical realm of \citep{ReduNet}, since we are no longer finding pure maximizers of $\Delta R(\Z)$, but rather stable minimax equilibria.

As is the case with GANs, these equilibria may not necessarily be Nash equilibria \citep{Nash-Gan}, but rather in the more general sense of Stackelberg \citep{fiez2021local}. So the problem of studying minimax equilibria of \eqref{eq:MCR2-GAN-objective} is likely, in its most general form, quite challenging. Nevertheless, our experiments suggest such equilibria tend to be well-behaved, e.g. having large range of attraction.\footnote{Our extensive empirical experiments and ablation studies indicate that in general the minimax objective converges rather stably to good equilibria for all the real datasets without any special optimization tricks or particular requirements on the networks. The only important factor for the stability of the optimization seems to be a large enough batch size (see Appendix \ref{app:ablation-batch-size}).} These observations can be further corroborated with analysis on simpler models: our ongoing work suggests if we restrict our attention to simplified data structures (e.g. $\X$ distributed on a linear subspace), then one can provide theoretical guarantees that the equilibria become efficiently and correctly solvable by the minimax formulation. Extending such analysis to more sophisticated data structures (multiple subspaces, nonlinear submanifolds) remain as exciting new directions for future research.

Despite many possible pathological solutions to the minimax game, empirically, as we have presented in the previous section (alongside many examples in the Appendix), the solution found by the simple GDA algorithm generally strikes a good trade-off between expressiveness and parsimony of the learned model. The solution automatically determines the proper  dimensions for different classes. Ablation studies in Appendix \ref{app:ablation-batch-size} on the large ImageNet dataset further suggest that this formulation is insensitive to over-parameterization by increasing network width as long as the batch size grows accordingly. However, a rigorous justification for such good model-selection properties remains widely open.  

\section{Conclusion and Future Work}\label{sec:conclusion}
This work provides a novel formulation for learning a {\em both generative and discriminative} representation for a multi-class multi-dimensional, possibly nonlinear,  distribution of real-world data. We have provided compelling empirical evidence that the distribution of most datasets can be effectively mapped to an LDR, a union of independent principal subspaces and principal components. This can be achieved with a closed-loop minimax game between the two encoder and the decoder networks without any additional network(s). The objective function is entirely based on an intrinsic information-theoretic measure, the rate reduction, without any other heuristics or regularizing terms.

The main purpose of this paper is to demonstrate the conceptual simplicity and practical potential of this new  framework for distribution/representation learning, instead of striving  for state of the art performance with heavy engineering. Nevertheless, with our preliminary implementation, a more informative LDR of the data can be effectively learned with a simple closed-loop architecture for a variety of real-world multi-class multi-modal visual datasets, from small to large, from low-resolution to higher-resolution, from domain-specific to diverse categories. The so-learned  encoder $f$ already enjoys the benefits of AE/VAEs for their discriminative property and the decoder $g$ with the benefits of GANs for their good generative visual quality. But probably more importantly, the internal structures of the learned feature representation has now become transparent hence {\em fully interpretable and controllable} (for generative purposes): visual differences between classes are naturally ``disentangled'' as independent subspaces, while diverse visual attributes within each class are ``disentangled'' as principal components within each subspace. From extensive ablation studies given in the Appendix, we see that the rate reduction based objective can be stably optimized across a wide range of datasets and network architectures without any additional regularizations or engineering tricks. Both the {\em feedback closed-loop}  and the {\em rate-reduction measure} play indispensable roles in fostering the ease and success of finding the LDR transcription.

One may notice that there are many ways this simple formulation can be significantly improved or extended. Firstly, in this work, we have simply adopted networks that were designed for GANs, but they may not be optimal for the rate reduction type objectives. For example, our ablation study already suggests some of the components of such networks such as spectral normalization is not quite essential. Characteristics from the white-box ReduNet \citep{ReduNet} derived from optimizing rate reduction can be explored in the future. Secondly, notice that our rate reduction objectives do not impose any requirements on how individual samples should be encoded or decoded although the results from the multi-class objective indicate certain level of alignment on the individual samples. Recent studies such as DC-VAE \citep{parmar2021dual} or OpenGAN \citep{Open-GAN} suggest that imposing additional regularization on individual samples may further improve decoded visual quality. Such regularization can certainly be incorporated into this new framework. Last but not the least, compared to GANs and VAEs, our method leads to an {\em explicit} strutured model for the feature distribution: a mixture of incoherent subspace Gaussians. Such an explicit model has the potential of making many subsequent tasks easier and better: better control of  feature sampling for decoding and synthesis \citep{harkonen2020ganspace}, designing more robust generators and classifiers to noise and corruptions based on the low-dimensional structures identified, or even extending to the settings of incremental and online learning \citep{Wu-CVPR2021}. We leave all these new  directions, together with all the open theoretical problems posed in Section \ref{sec:discussions}, for future investigation.

\addcontentsline{toc}{section}{Acknowledgments}
\section*{Acknowledgments}
We would like to acknowledge that this work has been the result of a successful team effort. In particular, {\em the first four authors have contributed almost equally to this work.}

Earliest ideas of this work were germinated during a hiking event of Professor Ma's group on Berkeley hills during the summer of 2020. Former group members Dr. Chong You (now at Google) and Dr. Yichao Zhou (now at Apple) were part of a stimulating discussion on possible extensions or applications of a new rate reduction framework being developed then. During the preparation of this work, we have consulted several experts on some of the related topics. The authors would like to thank Professor Jiantao Jiao of UC Berkeley for discussions about the theoretical conditions for learning distributions via GANs. We thank Dr. Benjamin Haeffele of Johns Hopkins University for sharing thoughts on how to learn subspaces correctly and on how to optimize the rate reduction objectives efficiently. We also like to thank Professor Shankar Sastry and Dr. Manxi Wu of UC Berkeley and Dr. Chaobing Song of Univ. of Wisconsin-Madison for informative discussions on how to solve  minimax games correctly and efficiently, as well as Chih-Yuan Chiu and Druv Pai of UC Berkeley for engaging discussions on theoretical directions for LDR transcription. Last but not the least, we would like to thank professor Stefano Soatto of UCLA for stimulating discussions and sometimes heated debates on how information can be efficiently and effectively encoded in deep networks.

Yi acknowledges support from ONR grants N00014-20-1-2002 and N00014-22-1-2102, the joint Simons Foundation-NSF DMS grant \#2031899, as well as partialsupport from Berkeley FHL Vive Center for Enhanced Reality and Berkeley Center for Augmented Cognition, Tsinghua-Berkeley Shenzhen Institute (TBSI) Research Fund, and Berkeley AI Research (BAIR).

\bibliographystyle{iclr2022_conference}

\newpage
\appendix
\section{Appendix}
\label{sec:appendix}\label{app:appendix}

\subsection{Experiment Settings and Implementation Details}
\label{app:settings}

\textbf{Network backbones.} For MNIST, we use the standard CNN models in Table~\ref{arch:mnist_g} and Table~\ref{arch:mnist_d}, following the DCGAN architecture \citep{radford2015unsupervised}. We resize the MNIST image resolution from 28 $\times$ 28 to 32 $\times$ 32 to fit DCGAN architecture. All $\alpha$ in lReLU (lReLU is short for Leaky-ReLU) of the encoder are set to 0.2.

We adopted conv ResNet architectures for CIFAR-10 in Tables~\ref{arch:cifar_g} and \ref{arch:cifar_d}, and STL-10 shown in Tables~\ref{arch:stl_g} and \ref{arch:stl_d}. Each ResBlock up is same as Resnet, but add an up-sampler after the first conv layer. All batch normalization layers of ResBlock in encoder are replaced with spectral normalization layer. 

Finally, we use the same architecture for CelebA, LSUN-bedroom, ImageNet-128 (see Tables~\ref{arch:celeb_g} and \ref{arch:celeb_d}) as all three datasets have the same 128$\times$128 resolution. Again, each ResBlock up is same as Resnet, but add an up-sampler after the first conv layer. And all batch normalization layers in encoder are replaced with spectral normalization layer. All experiments utilize this lightweight PyTorch library that provides implementations of popular state-of-the-art GANs and evaluation metrics.

\begin{table}[t]
\setlength{\tabcolsep}{0.1cm}
\begin{minipage}{0.5\textwidth}
    \begin{tabular}{cc}
     \hline
     \hline
     $\z \in \R^{1 \times 1 \times 128}$  \\
     \hline
     4 $\times$ 4, stride=1, pad=0 deconv. BN 256 ReLU  \\
     \hline
     4 $\times$ 4, stride=2, pad=1 deconv. BN 128 ReLU  \\
     \hline
     4 $\times$ 4, stride=2, pad=1 deconv. BN 64 ReLU   \\
     \hline
     4 $\times$ 4, stride=2, pad=1 deconv. 1 Tanh  \\
     \hline
     \hline
    \end{tabular}
    \caption{Decoder for MNIST.}
    \label{arch:mnist_g}
\end{minipage}
\hspace{0.03\textwidth}
\begin{minipage}{0.5\textwidth}
    \begin{tabular}{cc}
     \hline
     \hline
     Gray image $\x \in \R^{32 \times 32 \times 1}$  \\
     \hline
     4 $\times$ 4, stride=2, pad=1 conv 64 lReLU    \\
     \hline
     4 $\times$ 4, stride=2, pad=1 conv. BN 128 lReLU   \\
     \hline
     4 $\times$ 4, stride=2, pad=1 conv. BN 256 lReLU   \\
     \hline
     4 $\times$ 4, stride=1, pad=0 conv 128   \\
     \hline
     \hline
    \end{tabular}
    \caption{Encoder for MNIST.}
    \label{arch:mnist_d}
\end{minipage}
\end{table}

\begin{table}[t]
\setlength{\tabcolsep}{0.5cm}
\begin{minipage}{0.5\textwidth}
\begin{tabular}{cc}
 \hline
 \hline
 $\z \in \R^{128}$  \\
 \hline
  dense $\xrightarrow{}$ 4 $\times$ 4 $\times$ 256  \\
 \hline
 ResBlock up 256  \\
 \hline
 ResBlock up 256  \\
 \hline
 ResBlock up 256  \\
 \hline
 BN, ReLU, 3 $\times$ 3 conv, 3 Tanh  \\
 \hline
 \hline
 \end{tabular}
\caption{Decoder for CIFAR-10. }
\label{arch:cifar_g}
\end{minipage}
\hspace{0.03\textwidth}
\begin{minipage}{0.5\textwidth}
\begin{tabular}{cc}
 \hline
 \hline
 RGB image $\x \in \R^{32 \times 32 \times 3}$  \\
 \hline
 ResBlock down 128  \\
 \hline
 ResBlock down 128  \\
 \hline
 ResBlock 128 \\
 \hline
 ResBlock 128 \\
 \hline
 ReLU  \\
 \hline
 Global sum pooling \\
 \hline
 dense $\xrightarrow{}$ 128  \\
 \hline
 \hline
 \end{tabular}
\caption{Encoder for CIFAR-10. }
\label{arch:cifar_d}
\end{minipage}
\end{table}

\begin{table}[t]
\setlength{\tabcolsep}{0.5cm}
\begin{minipage}{0.5\textwidth}
    \begin{tabular}{cc}
 \hline
 \hline
 $\z \in \R^{128}$  \\
 \hline
 dense $\xrightarrow{}$ 6 $\times$ 6 $\times$ 512  \\
 \hline
 ResBlock up 256  \\
 \hline
 ResBlock up 128  \\
 \hline
 ResBlock up 64  \\
 \hline
 BN, ReLU, 3 $\times$ 3 conv, 3 Tanh  \\
 \hline
 \hline
 \end{tabular}
 \caption{Decoder for STL-10. }
 \label{arch:stl_g}
\end{minipage}
\hspace{0.03\textwidth}
\begin{minipage}{0.5\textwidth}
    \begin{tabular}{cc}
 \hline
 \hline
 RGB image $\x \in \R^{48 \times 48 \times 3}$  \\
 \hline
 ResBlock down 64  \\
 \hline
 ResBlock down 128  \\
 \hline
 ResBlock down 256 \\
 \hline
 ResBlock down 512 \\
 \hline
 ResBlock 1024 \\
 \hline
 ReLU  \\
 \hline
 Global sum pooling \\
 \hline
 dense $\xrightarrow{}$ 128  \\
 \hline
 \hline
 \end{tabular}
 \caption{Encoder for STL-10. }
 \label{arch:stl_d}
\end{minipage}
\end{table}

\begin{table}[t]
\setlength{\tabcolsep}{0.5cm}
\begin{minipage}{0.5\textwidth}
    \begin{tabular}{cc}
 \hline
 \hline
 $\z \in \R^{128}$  \\
 \hline
  dense $\xrightarrow{}$ 4 $\times$ 4 $\times$ 1024  \\
 \hline
 ResBlock up 1024  \\
 \hline
 ResBlock up 512  \\
 \hline
 ResBlock up 256  \\
 \hline
 ResBlock up 128  \\
 \hline
 ResBlock up 64  \\
 \hline
 BN, ReLU, 3 $\times$ 3 conv, 3 Tanh  \\
 \hline
 \hline
 \end{tabular}
 \caption{Decoder for CelebA-128, LSUN-bedroom-128, and ImageNet-128. }
 \label{arch:celeb_g}
\end{minipage}
\hspace{0.03\textwidth}
\begin{minipage}{0.5\textwidth}
    \begin{tabular}{cc}
 \hline
 \hline
 RGB image $\x \in \R^{128 \times 128 \times 3}$  \\
 \hline
 ResBlock down 64  \\
 \hline
 ResBlock down 128  \\
 \hline
 ResBlock down 256 \\
 \hline
 ResBlock down 512 \\
 \hline
 ResBlock down 1024 \\
 \hline
 ResBlock 1024 \\
 \hline
 ReLU  \\
 \hline
 Global sum pooling \\
 \hline
 dense $\xrightarrow{}$ 128  \\
 \hline
 \hline
 \end{tabular}
 \caption{Encoder for CelebA-128, LSUN-bedroom-128, and ImageNet-128.}
 \label{arch:celeb_d}
\end{minipage}
\end{table}

\textbf{Optimization and training details.} Across all of our experiments, we use Adam \citep{kingma2014adam} as our optimizer, with hyperparameters $\beta_1 = 0.5, \beta_2=0.999$. We adopt the simple gradient descent-ascent algorithm for alternating minimizing and maximizing the objectives. The initial value of learning rate is set to be 0.00015 and is scheduled with linear decay. We choose $\epsilon^2=0.5$ for both equation~\eqref{eq:MCR2-GAN-objective} and \eqref{eq:MCR2-GAN-objective-binary} in all \ours{} experiments. For all \ours{}-Multi experiments on ImageNet, we only choose 10 classes. The details of the 10 classes as shown in Table~\ref{tab:imagenet10}. Most experiments are trained on RTX 3090ti GPUs. 

\begin{table}[t]
    \centering
    \small
    \setlength{\tabcolsep}{6.5pt}
    \renewcommand{\arraystretch}{1.25}
    \begin{tabular}{c|c}
    ID        & Category  \\

    \hline
    \hline
    n02930766 & cab, hack, taxi, taxicab \\
    n04596742 & wok        \\
    n02974003 & car wheel    \\
    n01491361 & tiger shark, Galeocerdo cuvieri   \\
    n01514859 & hen    \\
    n09472597 & volcano   \\
    n07749582 & lemon \\
    n09428293 & seashore, coast, seacoast, sea-coast \\
    n02504458 & African elephant, Loxodonta africana \\
    n04285008 & sports car, sport car  \\
    \hline
    \end{tabular}      
    \caption{\small ID and correspond category for 10 classes of ImageNet}
    \label{tab:imagenet10}
\end{table}

\subsection{{MNIST}}\label{app:MNIST}
\textbf{Settings.}
On the MNIST dataset, we train our model using DCGAN \citep{radford2015unsupervised} architecture with our proposed models LDR-Multi \eqref{eq:MCR2-GAN-objective} and LDR-Binary \eqref{eq:MCR2-GAN-objective-binary}. We set the learning rate to $10^{-4}$, batch size to 2048, and train for 15,000 iterations. Due to the advantages of the \ours{} objective, we can achieve between-class discriminative representations while the within-class diversity of these representations can be preserved, which are shown in the following experimental results. 

\begin{figure}[htbp]
     \centering
     \subfigure[Original $\X$]{
         \includegraphics[trim=2.5cm 2cm 2cm 2.5cm ,clip,width=0.18\textwidth]{MNIST_MNIST_train_images_epoch200.png}
         \label{fig:MNIST_train_imgs}
     }
    \subfigure[VAE-GAN $\hat \X$]{
         \includegraphics[trim=3.5cm 1.2cm 3.2cm 1.5cm ,clip,width=0.18\textwidth]{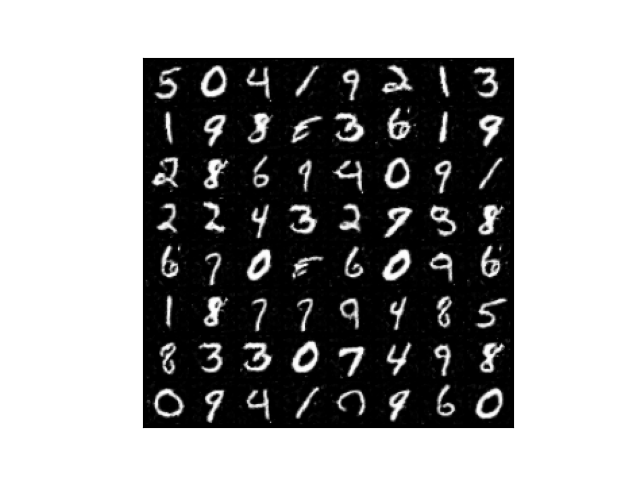}
         \label{fig:MNIST_train_recon_VAEGAN}
     }
    \subfigure[BiGAN $\hat \X$]{
         \includegraphics[width=0.18\textwidth]{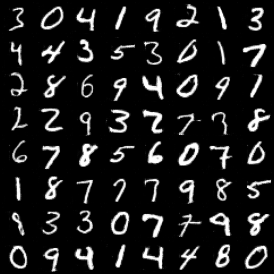}
         \label{fig:MNIST_train_recon_BiGAN}
     }
    \subfigure[LDR-Binary $\hat \X$]{
         \includegraphics[trim=2.5cm 2cm 2cm 2.5cm ,clip,width=0.18\textwidth]{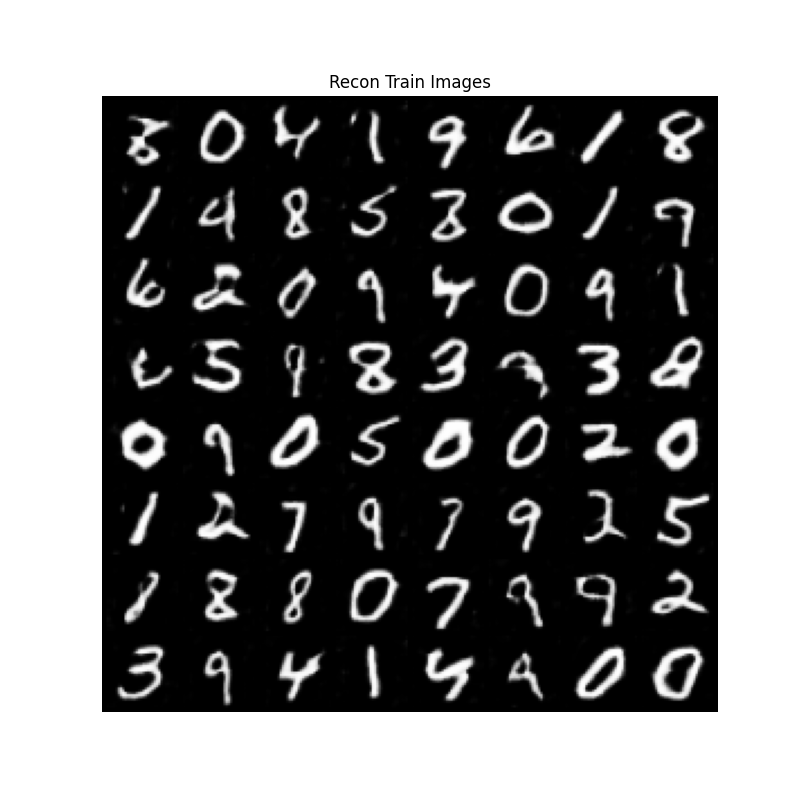}
         \label{fig:MNIST_train_recon_binary}
     }
     \subfigure[LDR-Multi $\hat \X$]{
         \includegraphics[trim=2.5cm 2cm 2cm 2.5cm ,clip,width=0.18\textwidth]{MNIST_MNIST_train_recon_images_epoch200_multi.png}
         \label{fig:MNIST_train_recon_multi}
     }
    \caption{The comparison of the reconstruction results of different methods with the input data.}
    \label{fig:MNIST_recon}
\end{figure}

\textbf{More results illustrating auto-encoding.}
Here we give more reconstruction results, or $\hat \X$, of our LDR-Multi and LDR-Binary models, compared to their corresponding original input $\X$. As shown in the Figure \ref{fig:MNIST_recon}, for the LDR-Binary model, it can generate clean digit-like images but the decoded $\hat\X$ might resemble digits from  similar but different classes to the input data $\X$ since LDR-Binary tends to only align the distribution of all digits. 

In contrast, with the LDR-Multi objective, the decoded $\hat \X$ not only are coherent with the correct class with the input data $\X$, but also show very clear one-to-one mapping between individual sample $\x$ and $\hat \x$ although the objective \eqref{eq:MCR2-GAN-objective} does not enforce that! 
Comparing with the results from the VAE-GAN \citep{VAE-GAN} and BiGAN \citep{donahue2016adversarial}, our decoded images {make less errors in reconstruction and }preserve much better the individual characteristics of the original samples. 

\textbf{Images decoded from random samples on the learned multi-class LDR.}
Since our LDR-Multi objective function maps input data of each class into a different (orthogonal) subspace in the feature space, we can generate images conditioned on each class by random sampling $\z$ in the subspace of each class and then decode them back to the input space as $\hat \x$.

To do random sampling in the learned subspace, we first calculate the mean feature $\Bar{\z}_j$ and the singular vectors $\bm v_{j}^{i}$ of the SVD (or principal components) of the learned features $\Z_j$ of the class $j$ of the training data, where index $i$ represents the $i$th principal components. We only use top $r =8$ principal components of each class on MNIST dataset. These statistics of the subspace can be used for guiding the random sampling. Then we sample $\z$ randomly along the principal components and around the mean feature as 
\begin{equation}\z_{random\_j} = \Bar{\z}_j + \alpha \sum_{i=1}^{r} n_{i} * \sigma_{j}^{i} * \bm v_{j}^{i},
\label{eqn:sample-components}
\end{equation}
where $\Bar{\z}_j$ is the mean feature of class $j$, $\sigma_{j}^{i}$ and $\bm v_{j}^{i}$ are the $i$-th singular value and principal component of class $j$, $n_i$ are  i.i.d. Gaussian $\mathcal{N}(0,1)$ random variables. That is, the feature in each subspace/class is modeled by a $r$-dimensional multivariate Gaussian, with variances $\sigma_{j}^{i}$ which characterize variances of the  training data in the feature space. Here, $\alpha$ is a hyper-parameter that controls the sampling range. As for visualization of random generated images  $g(\z_{random\_j})$ conditioned on the given class, we compare our method with some other conditional generation method such as ACGAN \citep{odena2017conditional} and InfoGAN \citep{infogan} (For ACGAN and InfoGAN, we generate images conditioned on class labels with randomly sampled latent $\z$ according the procedures mentioned in their respective works). Our model can give realistic and correct conditional generation results with high diversity in each class, while other methods may make mistakes in the generation between some similar classes such as classes 3 and 5 for InfoGAN.

\begin{figure}[htbp]
     \centering
     \subfigure[ACGAN]{
         \includegraphics[width=0.3\textwidth]{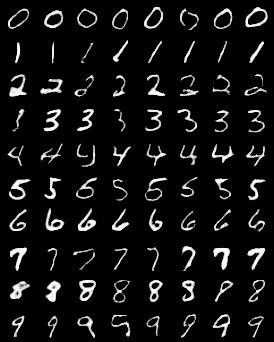}
         \label{fig:MNIST_ACGAN}
     }
    \subfigure[InfoGAN]{
         \includegraphics[trim=0.1cm 0.1cm 0.1cm 0.1cm,clip,width=0.3\textwidth]{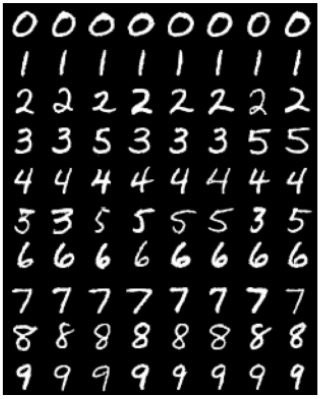}
         \label{fig:MNIST_InfoGAN}
     }
     \subfigure[LDR-Multi]{
         \includegraphics[width=0.30\textwidth]{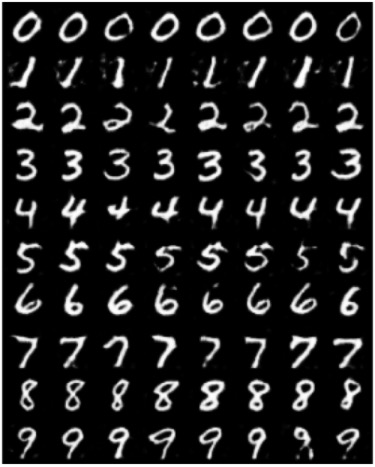}
         \label{fig:MNIST_LDR_multi_cond_gen}
     }
    \caption{Comparison of randomly generated images conditioned on each class.}
    \label{fig:MNIST_cond_gen}
\end{figure}

\begin{figure}[htbp]
\centerline{\includegraphics[trim=1cm 1cm 1cm 1cm,clip,width=0.5\textwidth]{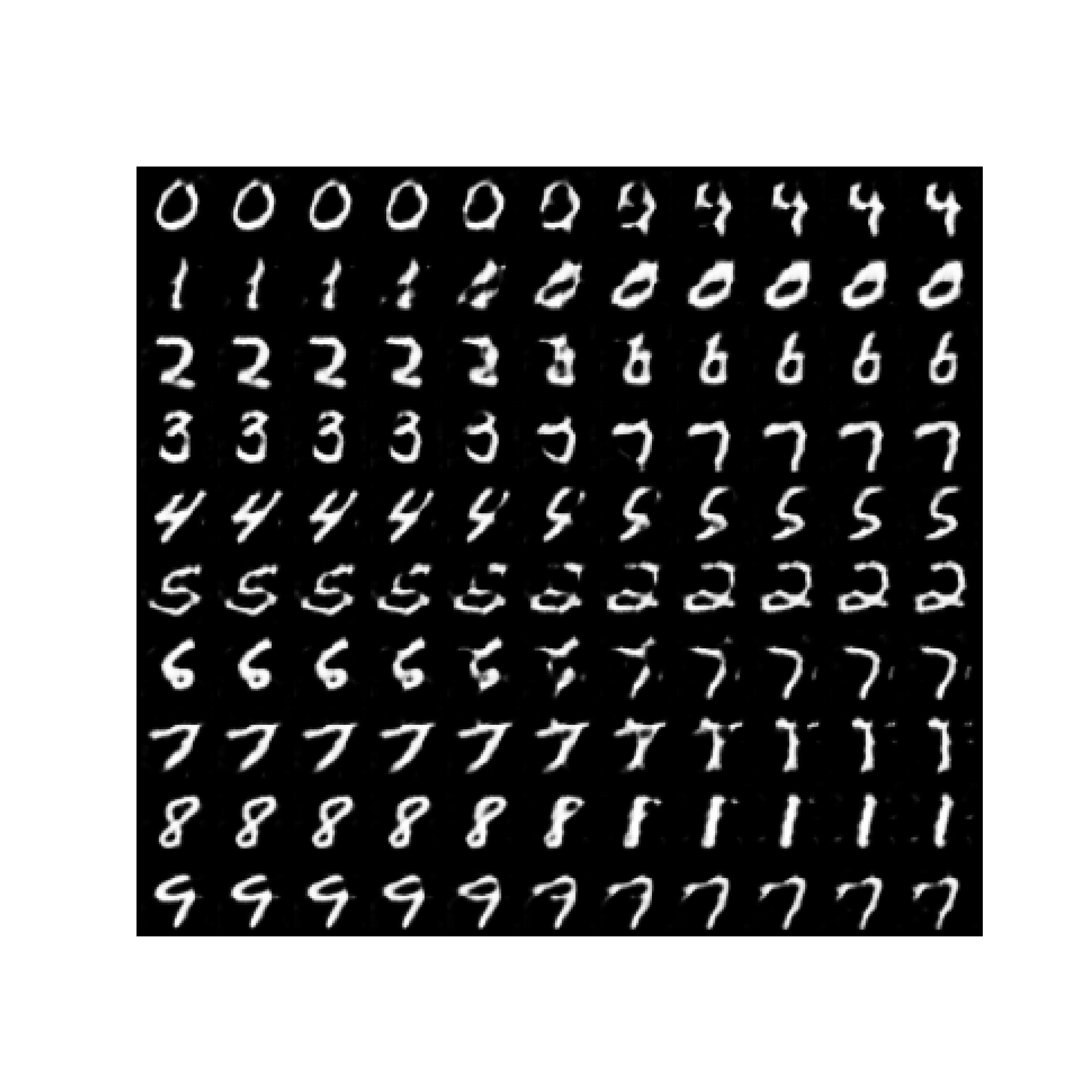}}
 \caption{Images generated from the interpolation between samples in different classes.}
 \label{fig:mnist_interpolation}
\end{figure}

\textbf{Interpolation between samples in different classes.}
We randomly sample some images from each class. For each image $\x_1$, we randomly sample another image $\x_2$ which is in a different class. For such a pair of images $\x_1$ and $\x_2$, we reconstruct based on their linearly interpolated feature representations by $\hat\x = g(\alpha f(\x_1)+(1-\alpha)f(\x_2)), \alpha \in[0, 1]$, the results of which are shown in the Figure \ref{fig:mnist_interpolation}. For each row in the figure from left to the right, the reconstructed images continuously morph from one digit to a different digit with a natural transition in shape rather than a simple superposition of the two images. This also confirms that space between subspaces for the digits does not represent valid digits but only shapes with digit-like strokes. Hence for generative purposes, knowing the supports of valid digits is extremely important. 

\subsection{{Transformed MNIST}}
\label{app:T-MNIST}
\begin{figure}[htbp]
     \centering
     \subfigure[Original $\X$]{
         \includegraphics[width=0.4\textwidth]{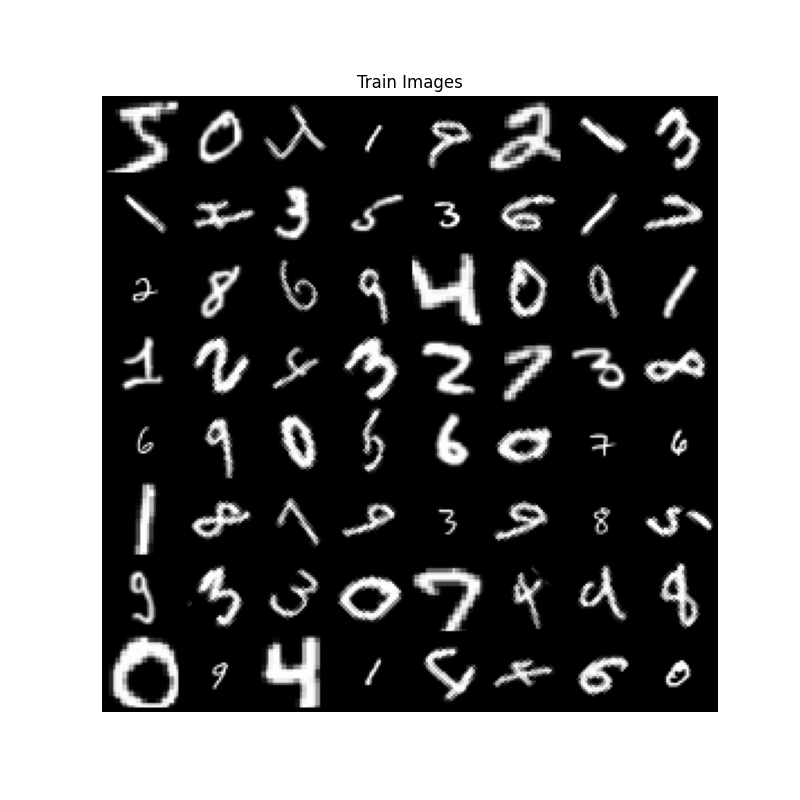}
         \label{fig:T-MNIST_train_imgs}
     }
     \subfigure[Decoded $\hat \X$]{
         \includegraphics[width=0.4\textwidth]{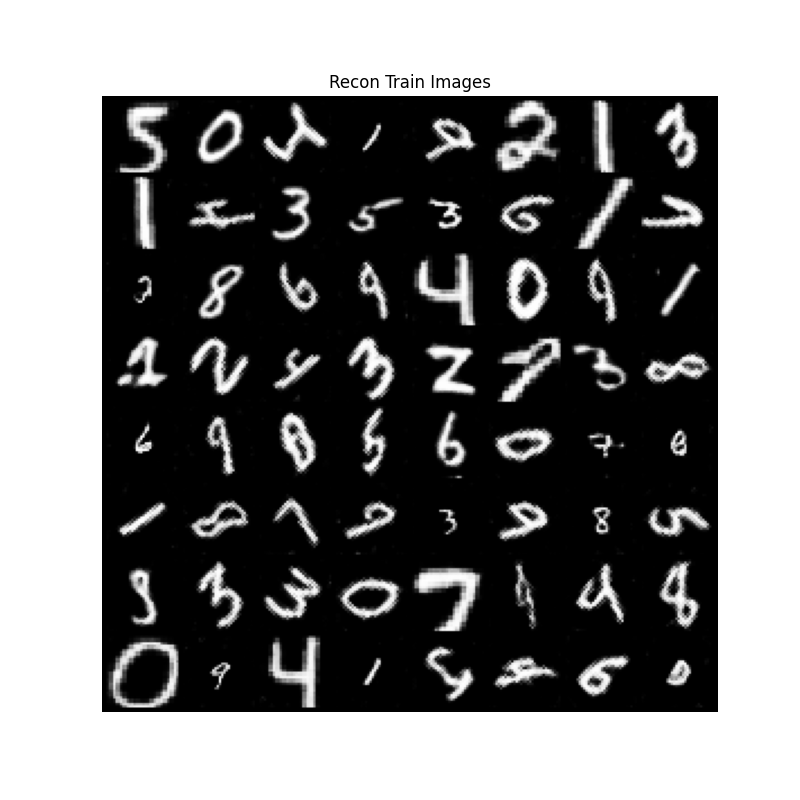}
         \label{fig:T-MNIST_train_recon}
     }
     \caption{Original (training) data $\X$ and their decoded version $\hat \X$ on transformed MNIST.}
    \label{fig:T-MNIST_train}
\end{figure}

\textbf{Settings.}
In this experiment, we verify that our LDR-Multi model can preserve diverse data modes in the learned feature embeddings.  We construct a transformed MNIST dataset with 5 modes: normal, large ($1.5\times$), small ($0.5\times$), rotate $45^{\circ}$ left, and rotate $45^{\circ}$ right. Each image data will be randomly transformed to one of the modes. Representative examples of such training data can be found in Figure \ref{fig:T-MNIST_train_imgs}. We train the model with learning rate 1e-4 and batch size 2048 for 15,000 iterations.

\textbf{Auto-encoding results.}
Figure \ref{fig:T-MNIST_train_recon} gives the decoded results of the training data with different modes. Even though the data are now much more diverse for each class, decoder learned from the LDR-Multi objective can still achieve high sample-wise similarity to the original images. 

\textbf{Identifying different modes.}
Similar to the earlier experiments of  Figure~\ref{fig:cifar_10_pca_sampling_main} for CIFAR-10 in the main paper, we find the top principal components of features of each class $\Z_j$ (via SVD) and generate new images using the learned decoder $g$ from features of the training images aligned the best with these components. 

\begin{figure}[htbp]
     \centering
     \subfigure[Components of class ``0'']{
         \includegraphics[width=0.3\textwidth]{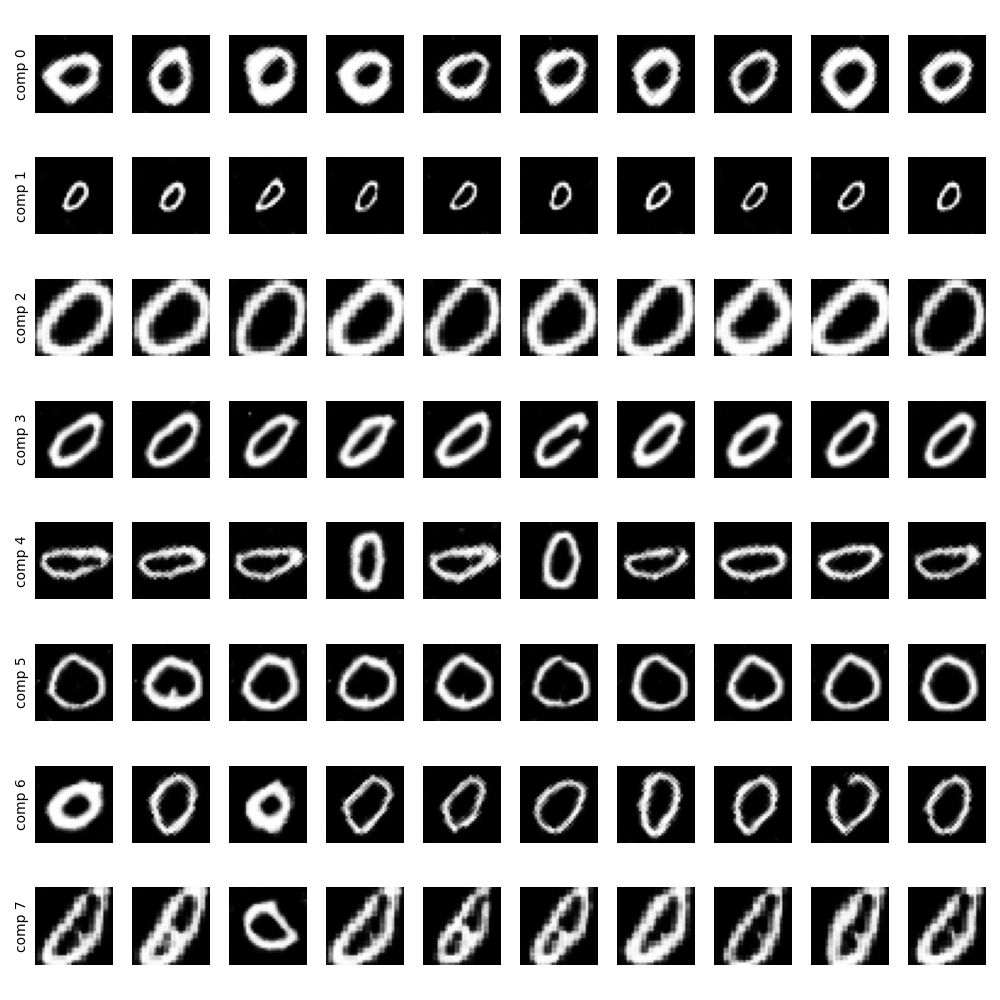}
         \label{fig:T-MNIST_class0}
     }
     \subfigure[Components of class ``1'']{
         \includegraphics[width=0.3\textwidth]{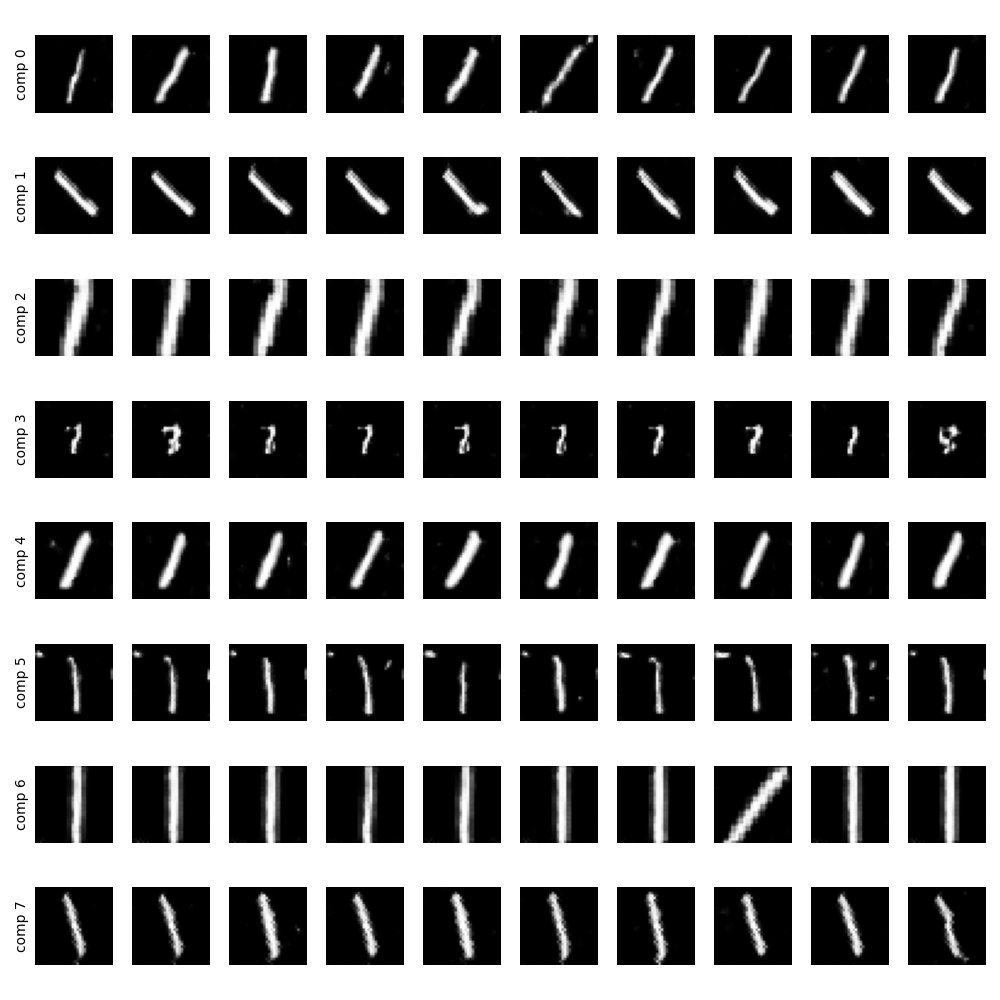}
         \label{fig:T-MNIST_class1}
     }
    \subfigure[Components of class ``2'']{
         \includegraphics[width=0.3\textwidth]{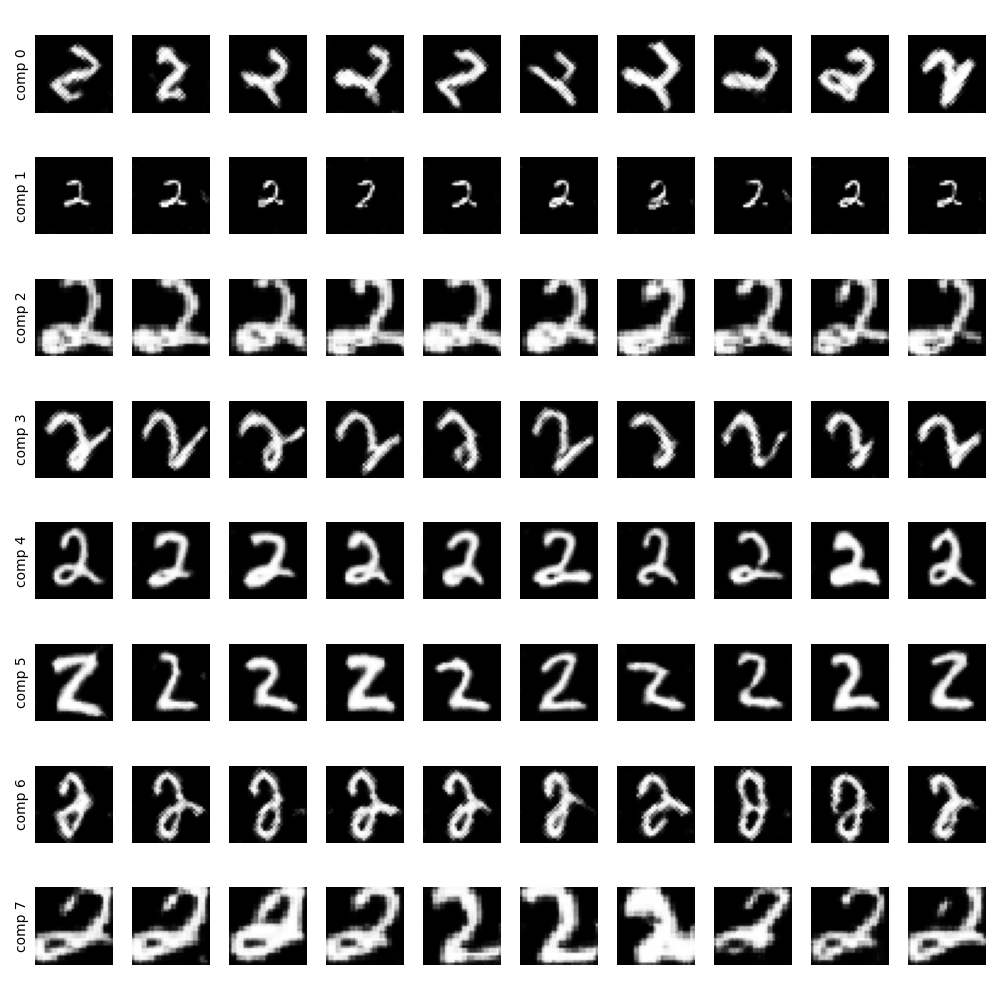}
         \label{fig:T-MNIST_class}
     }
     \caption{The reconstructed images $\hat \X$ from the features $\Z$ best aligned along top-8 principal components on the transformed MNIST dataset. Each row represents a different principal component.}
    \label{fig:T-MNIST_nearcomp_class}
\end{figure}

In Figure~\ref{fig:T-MNIST_nearcomp_class}, we select three classes 0, 1, 2 and visualize samples from top $r=8$ principal components for each class. Each row represents one principal component direction. As it can be seen in the figure, the decoded images along each principal component shows similar mode, and the modes along different component directions are rather incoherent. All major modes of the original data can be identified as one of these principal directions. This clearly shows that our LDR-Multi model can keep the different modes within each class of the data $\X_j$ as the principal component directions of $\Z_j$, and these modes can also be retained in the decoded images $\hat \X_j$.

\subsection{CIFAR-10}
\label{app:CIFAR}
\textbf{Settings.}
For all experiments on CIFAR-10, we follow the common training hyper-parameters in section~\ref{app:settings}. Beyond that, for each experiment, we run 450,000 iterations with mini-batch size 1600.

\begin{figure}[htbp]
     \centering
     \subfigure[ACGAN]{
         \includegraphics[width=0.31\textwidth]{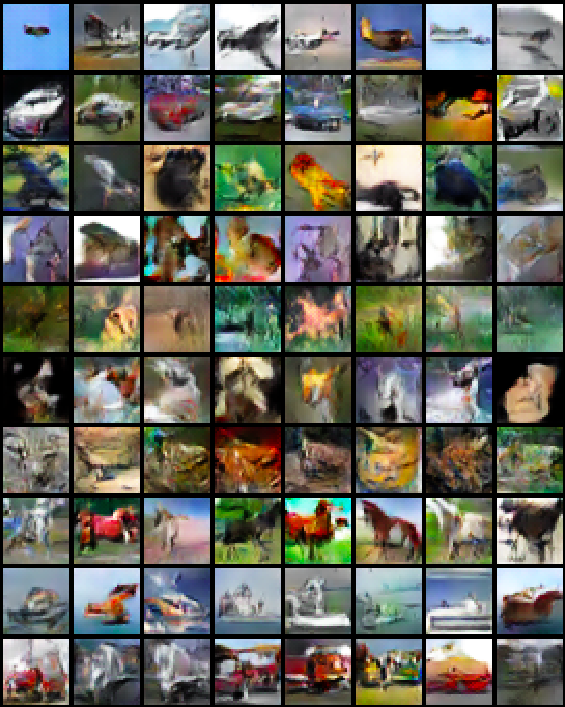}
         \label{fig:CIFAR_ACGAN}
     }
    \subfigure[InfoGAN]{
         \includegraphics[width=0.31\textwidth]{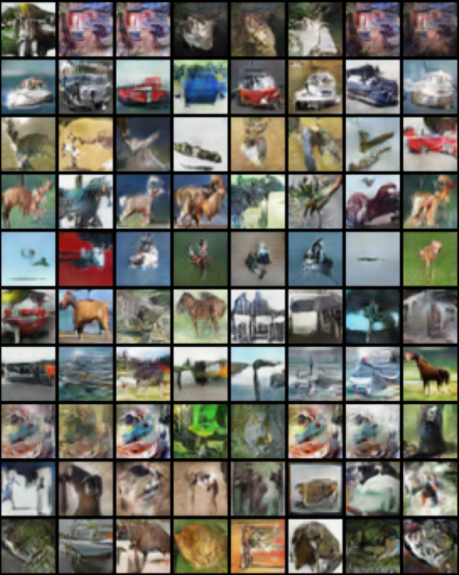}
         \label{fig:CIFAR_InfoGAN}
     }
     \subfigure[LDR-Multi]{
         \includegraphics[width=0.31\textwidth]{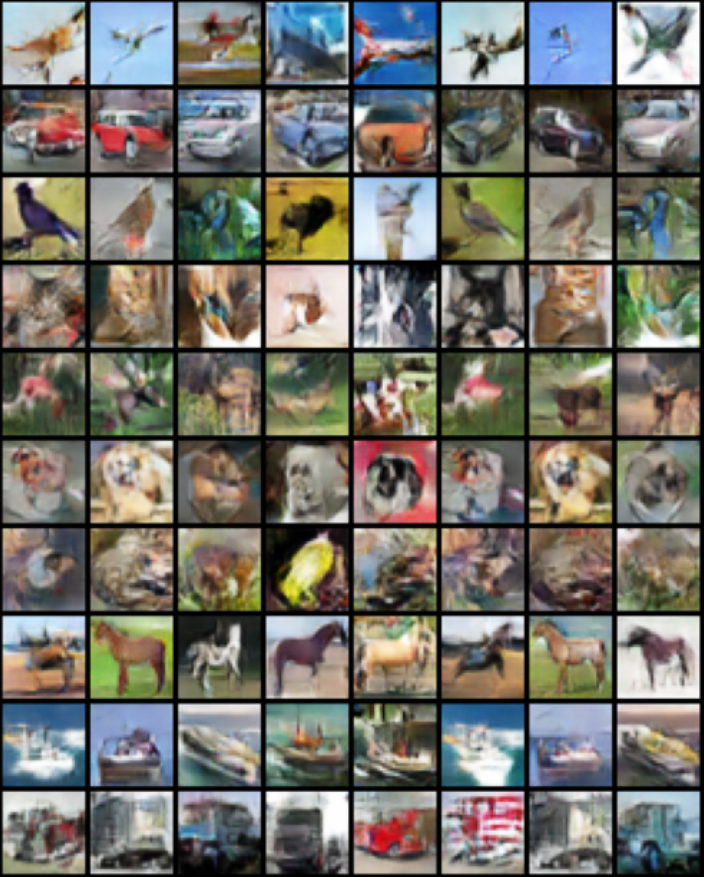}
         \label{fig:CIFAR_LDR_multi_cond_gen}
     }
    \caption{Comparison of randomly generated images conditioned on each class.}
    \label{fig:cifar_random_sample}
\end{figure}

\textbf{Images decoded from random samples on the learned multi-class LDR.} We sample $\z$ in the feature space randomly along the principal components and around the mean feature of each class $\Z_j$ as in the MNIST case, according to equation \eqref{eqn:sample-components}. The generated images from the sampled features are illustrated in Figure~\ref{fig:cifar_random_sample}, one row per class. As we see, the generator learned from the LDR-multi objective is capable of generating diverse images for each class. 

Further, for visualization of random generated images  $g(\z_{random\_j})$ conditioned on the given class, we also compare our method with some other conditional generation method such as ACGAN \citep{odena2017conditional} and InfoGAN \citep{infogan}. For all three experiments, we have randomly sampled 8 images per class in CIFAR-10. For more complex dataset like CIFAR-10, our model can give more realistic conditional generation results for different classes with high diversity within each class. 

\begin{figure}[htbp]
     \centering
     \subfigure[Airplane]{
         \includegraphics[width=0.315\textwidth]{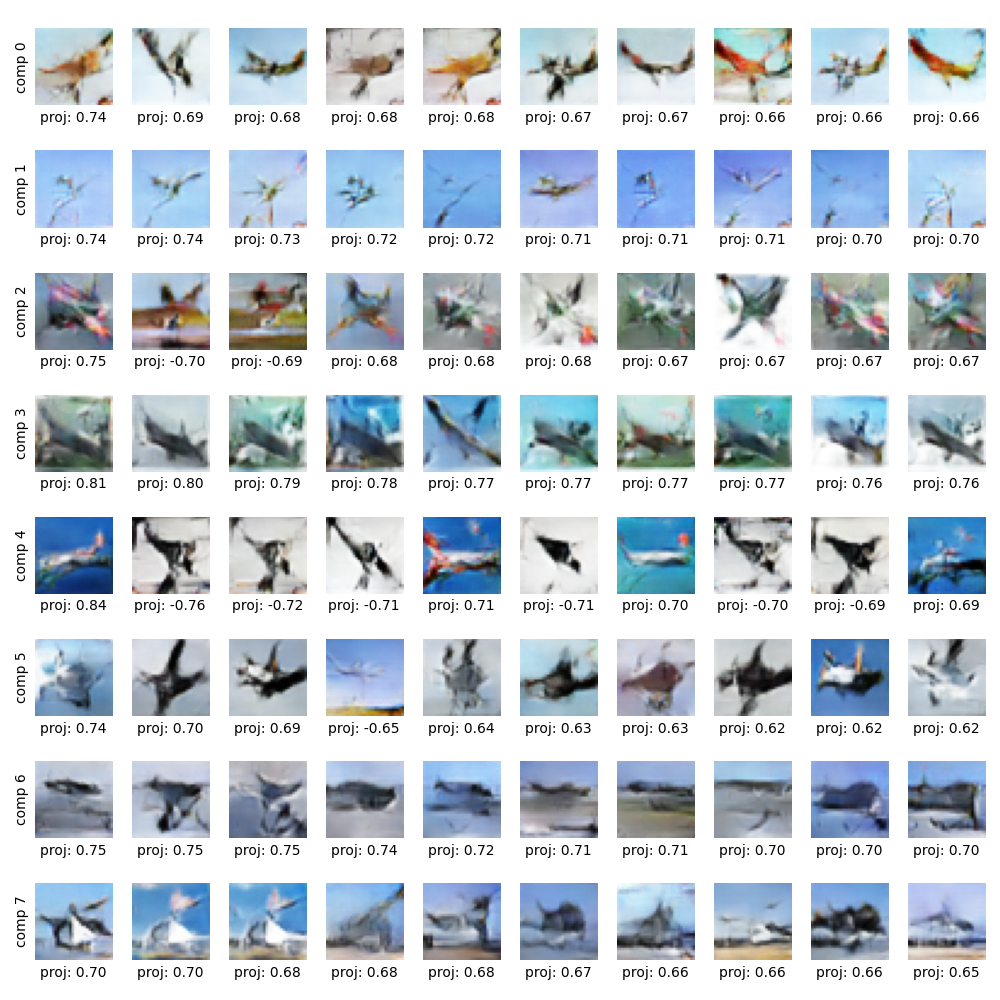}
         \label{fig:cifar_pca_images_a}
     }
     \subfigure[Automobile]{
         \includegraphics[width=0.315\textwidth]{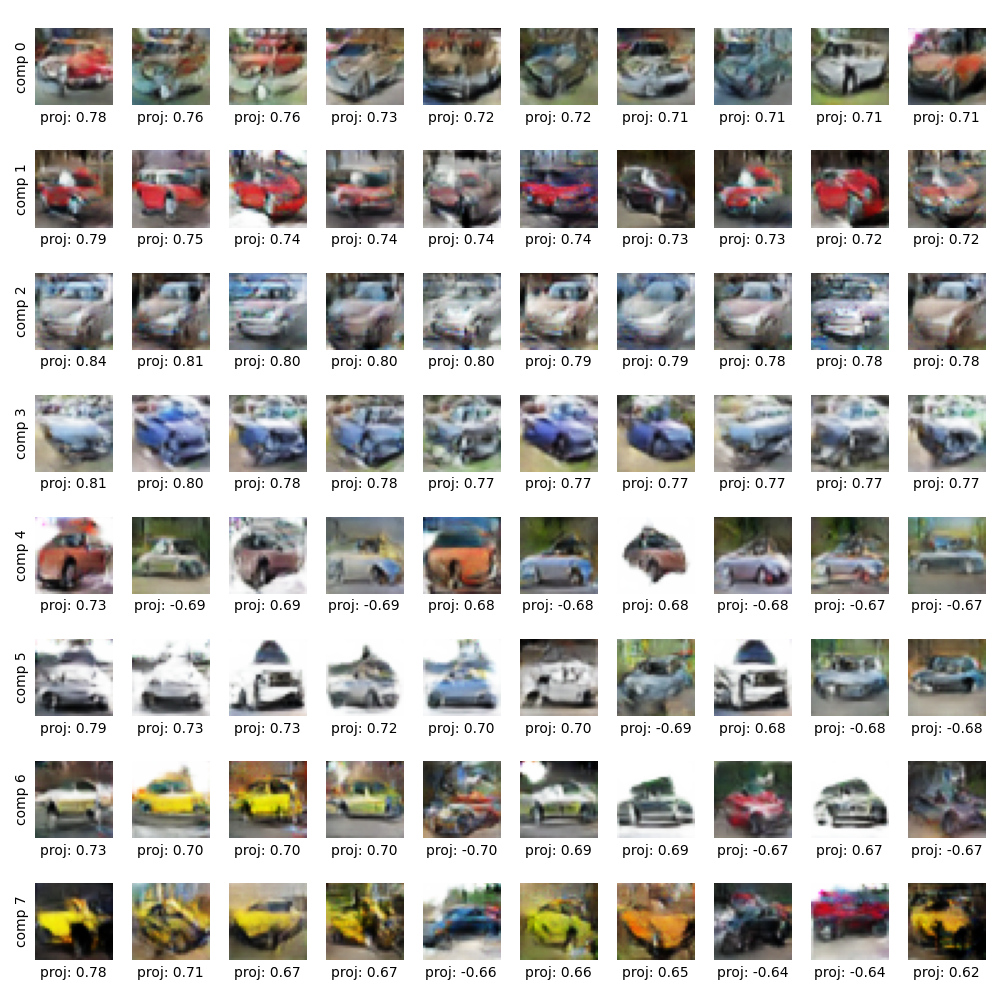}
         \label{fig:cifar_pca_images_b}
     }
    \subfigure[Bird]{
         \includegraphics[width=0.315\textwidth]{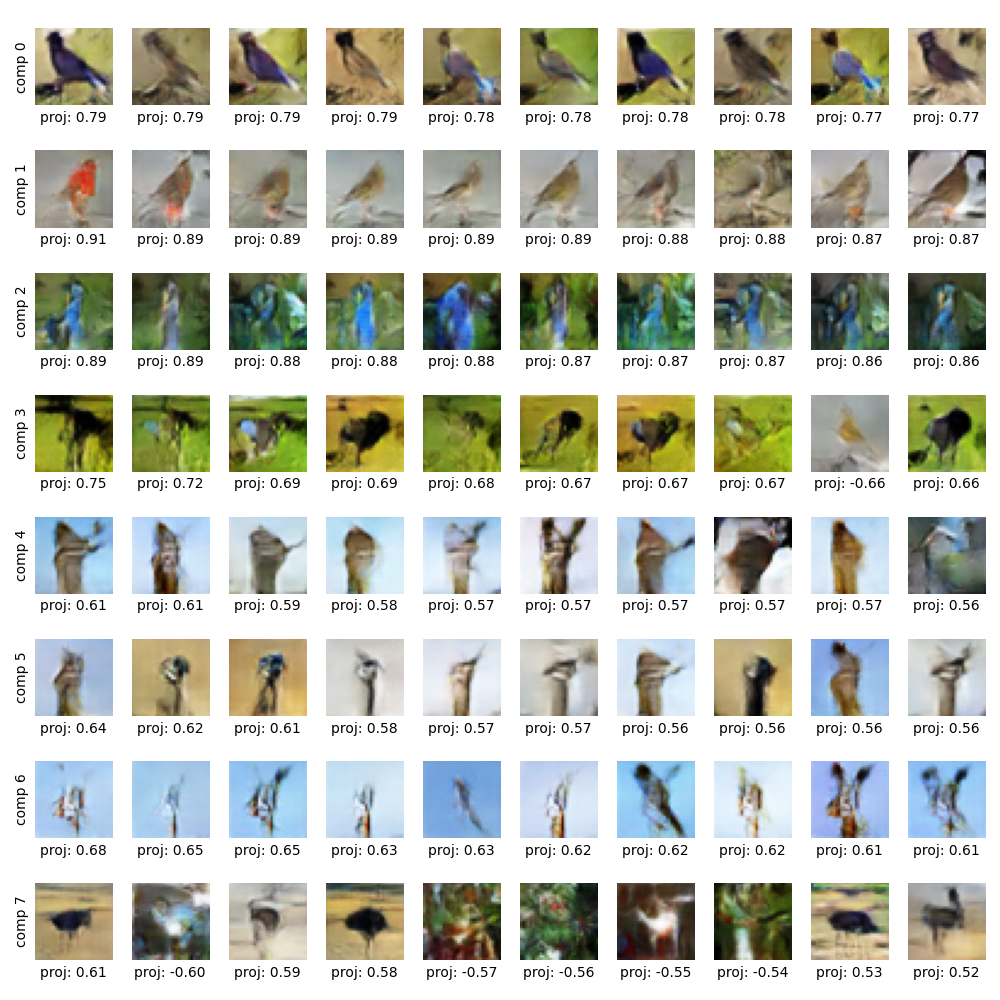}
         \label{fig:cifar_pca_images_c}
     }
     \subfigure[Cat]{
         \includegraphics[width=0.315\textwidth]{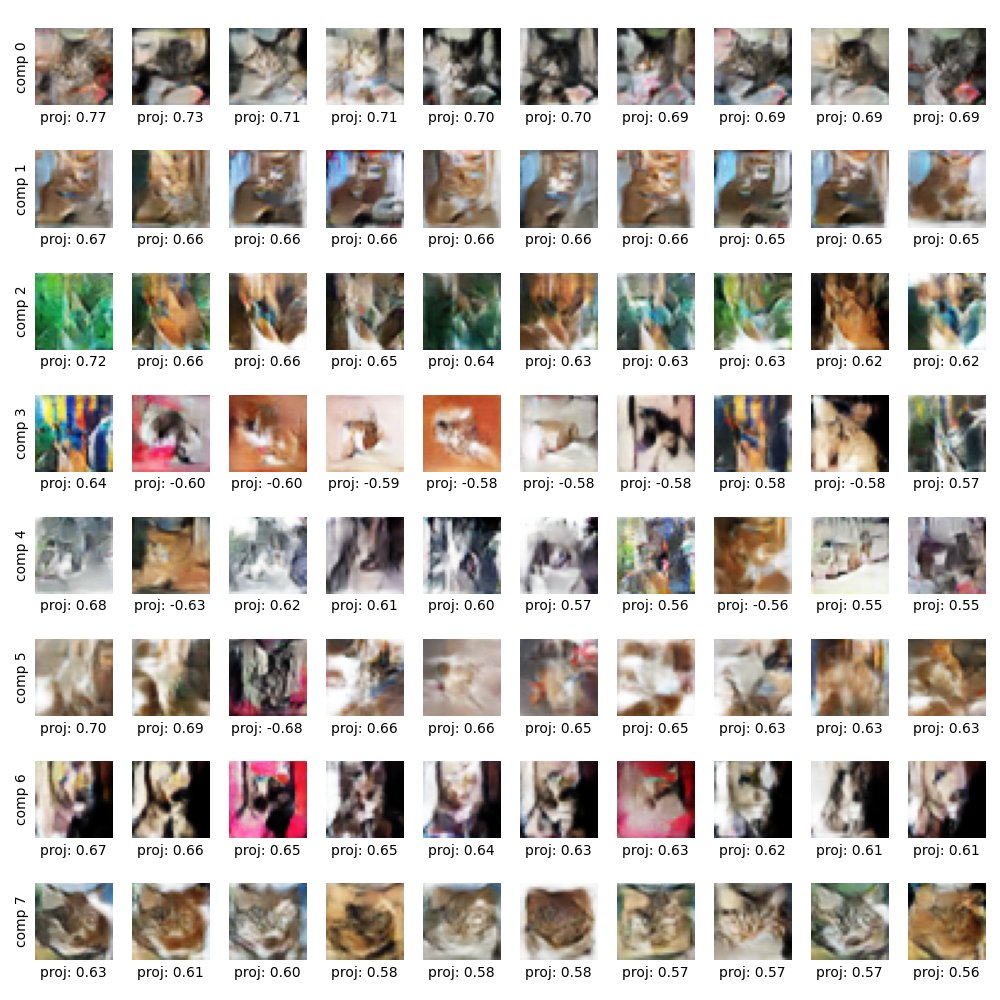}
         \label{fig:cifar_pca_images_d}
     }
     \subfigure[Deer]{
         \includegraphics[width=0.315\textwidth]{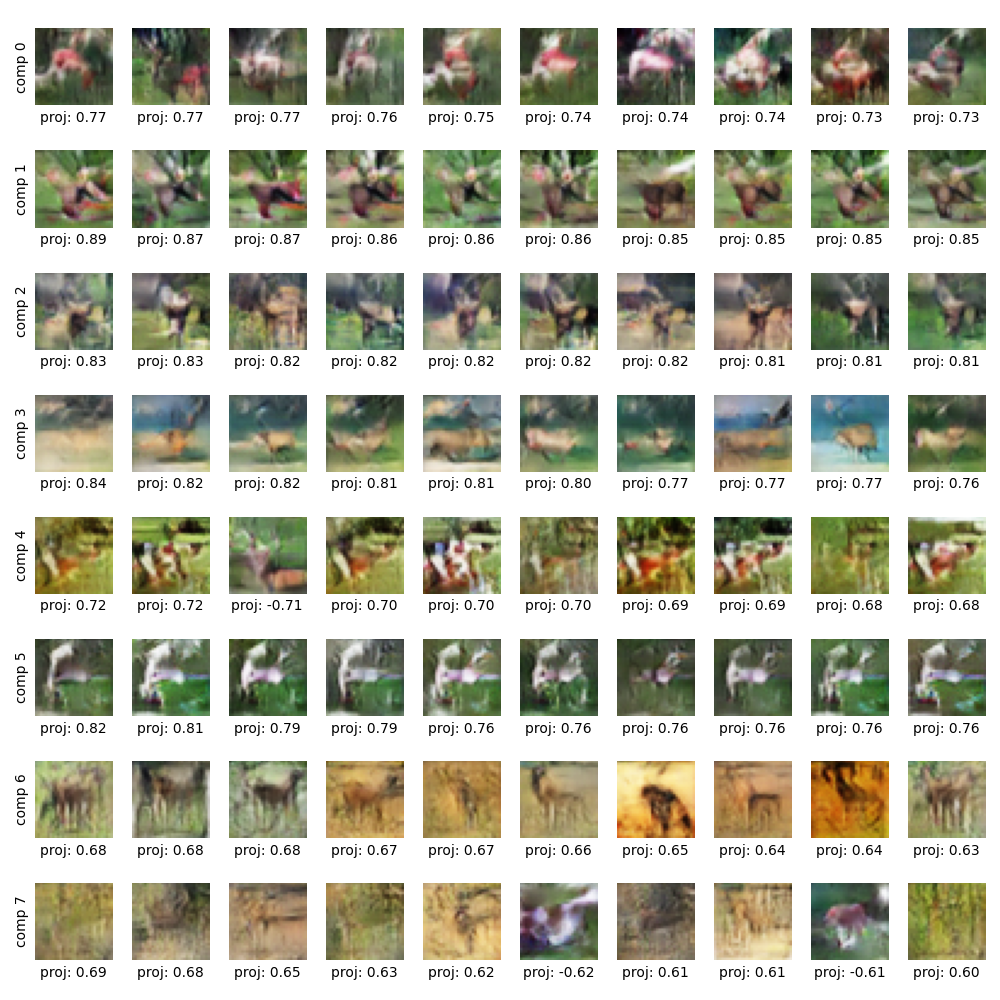}
         \label{fig:cifar_pca_images_e}
     }
     \subfigure[Dog]{
         \includegraphics[width=0.315\textwidth]{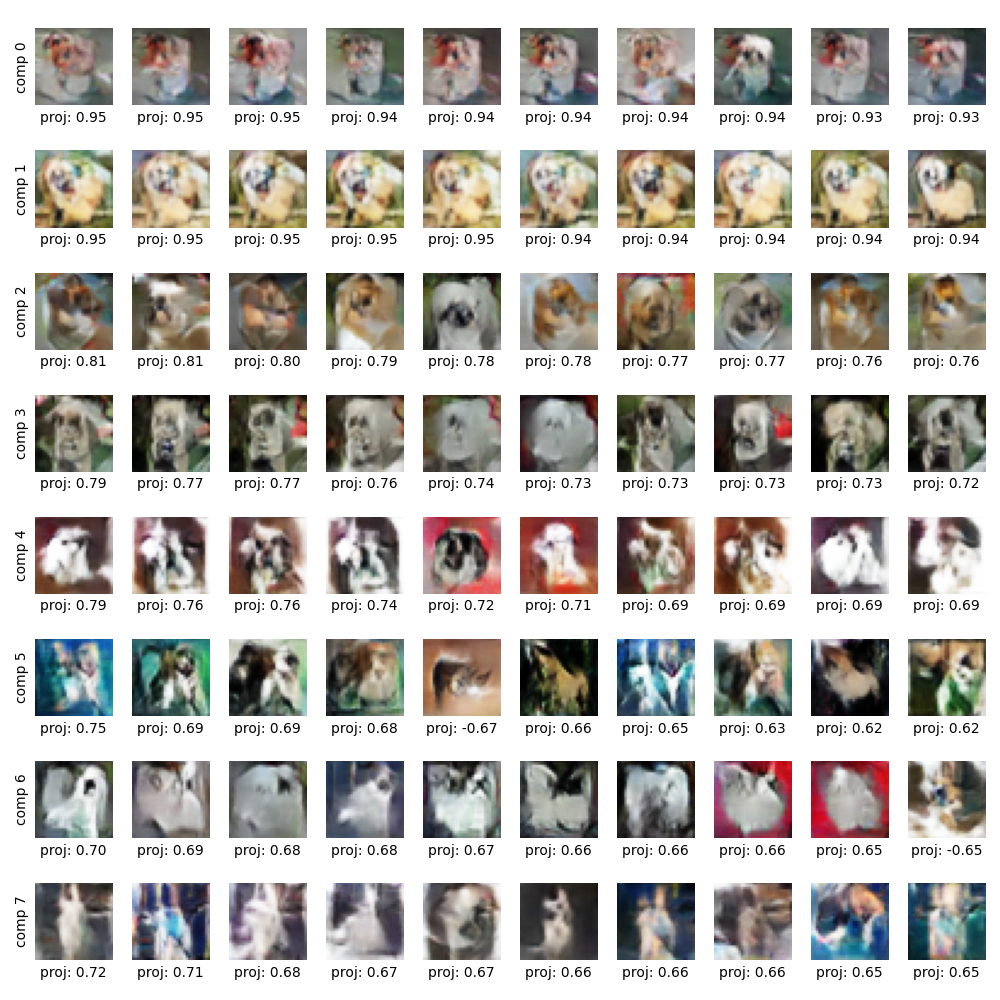}
         \label{fig:cifar_pca_images_f}
     }
     \subfigure[Frog]{
         \includegraphics[width=0.315\textwidth]{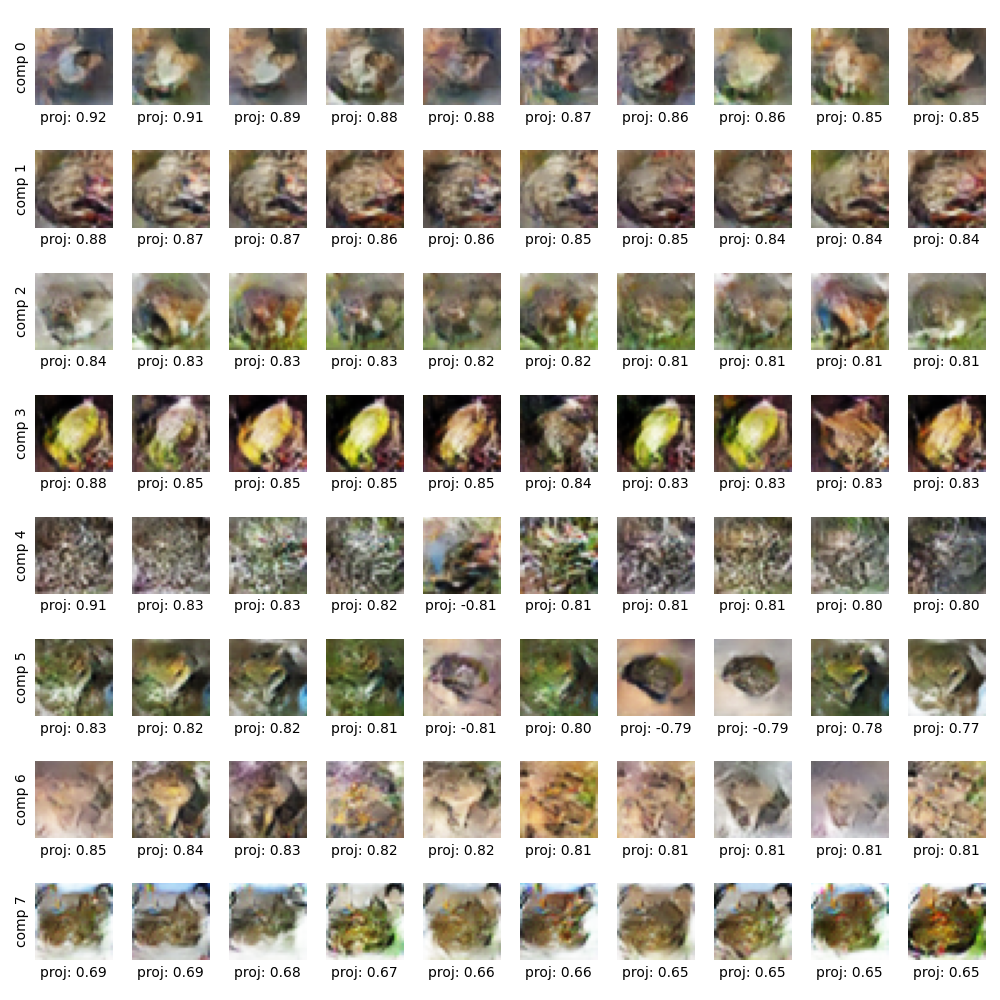}
         \label{fig:cifar_pca_images_g}
     }
     \subfigure[Horse]{
         \includegraphics[width=0.315\textwidth]{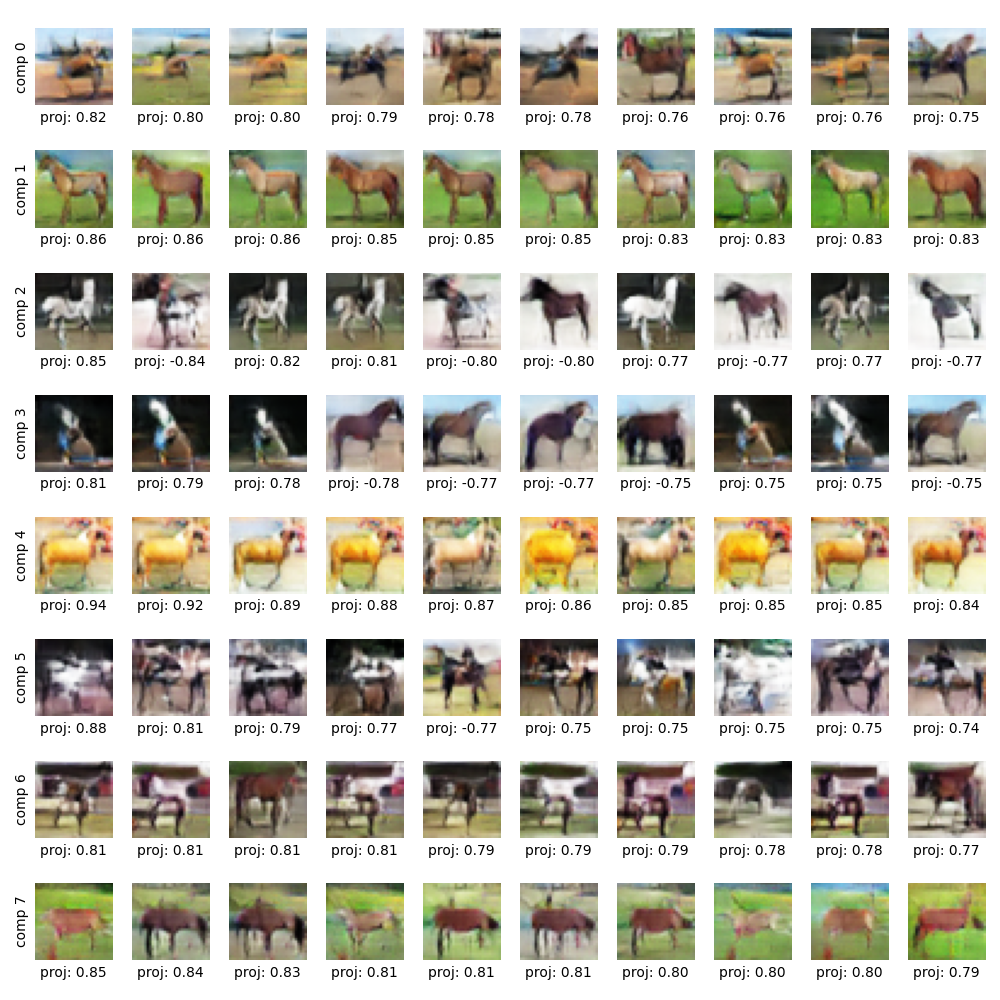}
         \label{fig:cifar_pca_images_h}
     }
     \subfigure[Ship]{
         \includegraphics[width=0.315\textwidth]{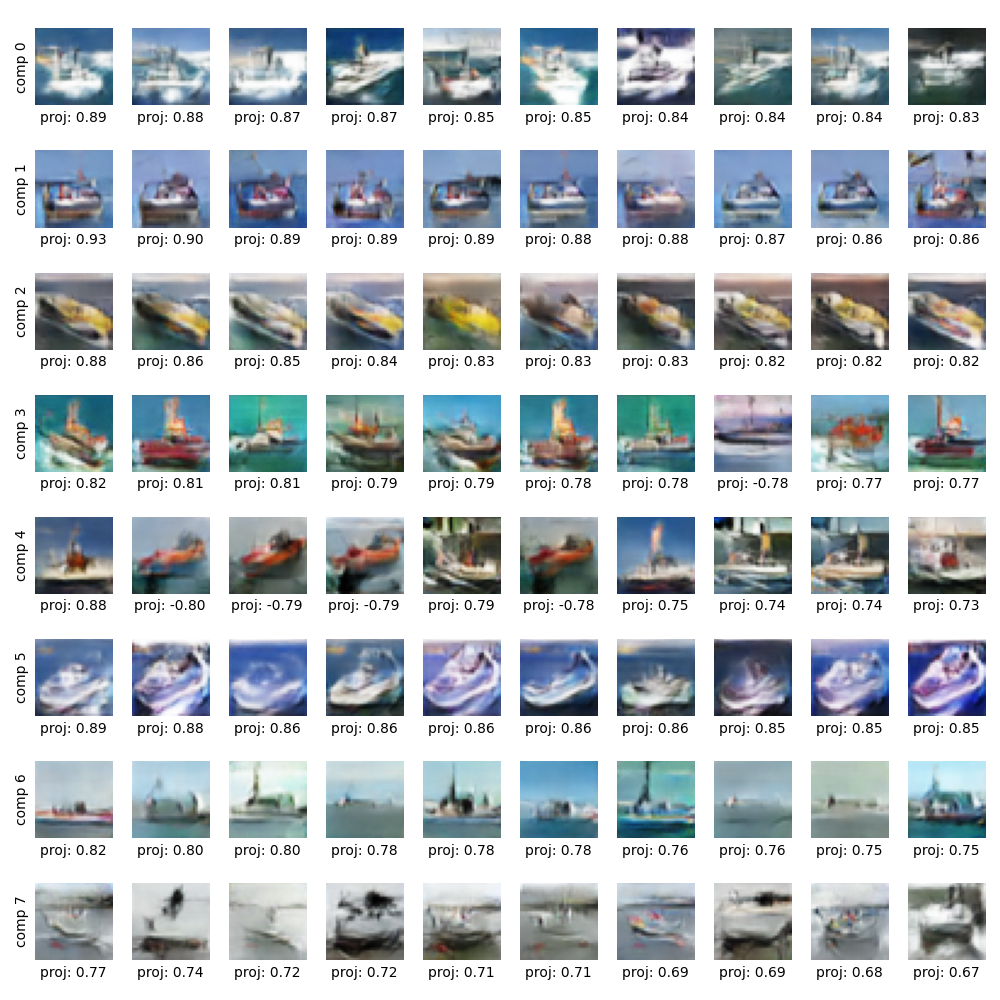}
         \label{fig:cifar_pca_images_i}
     }
     \subfigure[Truck]{
         \includegraphics[width=0.315\textwidth]{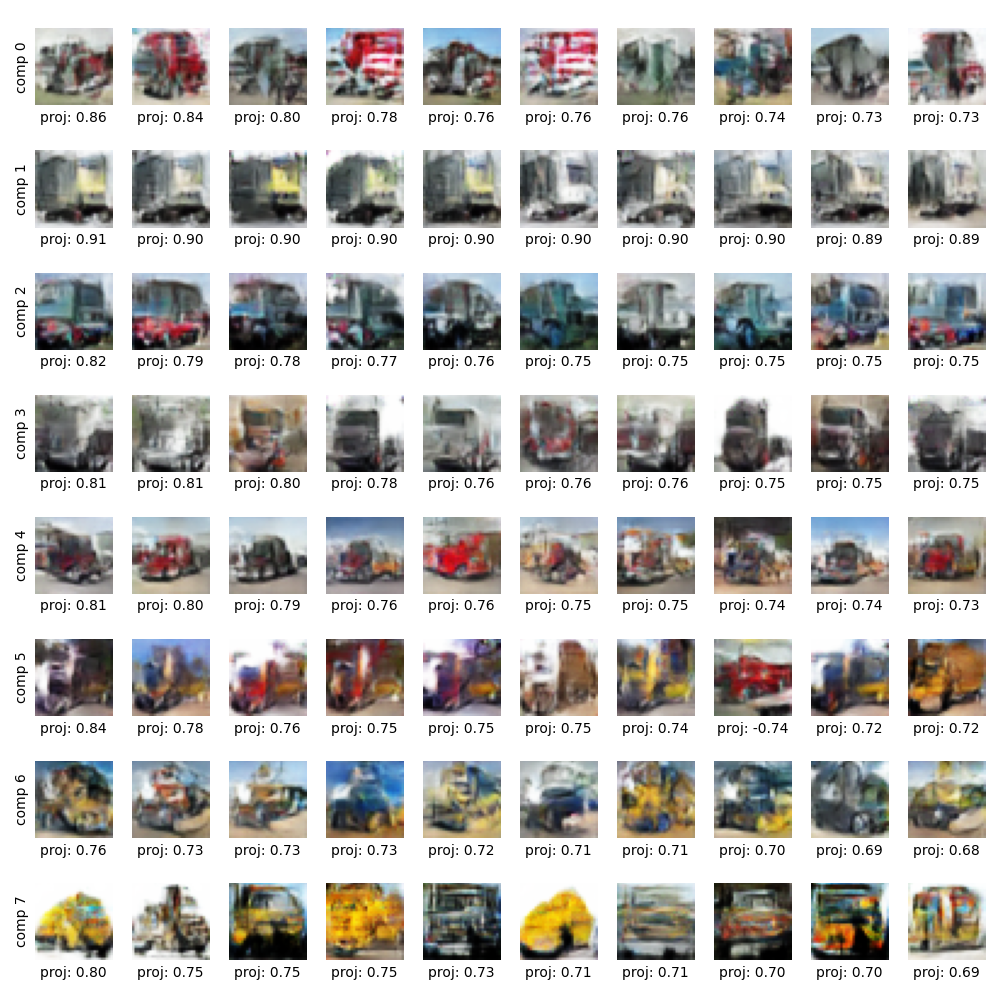}
         \label{fig:cifar_pca_images_j}
     }
     \caption{Reconstructed images $\hat \X$ from features $\Z$ close to the principal components learned for the 10 classes of CIFAR-10.}
    \label{fig:CIFAR10_PCA}
\end{figure}

\textbf{Generating image along different PCA components for each class.}
For each class, we first compute top-10 principal components (singular vectors of the SVD) of $\Z$ and then for each of the top singular vectors, we display in each row the top-10 reconstructed image $\hat \X$ whose $\Z$ are closest to the singular vector using method described in the main body of the paper, Section \ref{sec:use-LDR}. The results are given in Figure~\ref{fig:CIFAR10_PCA}. Notice that images in each row are very similar as they are sampled along the same principal component whereas images in different rows are very different as they are orthogonal in the feature space. These results indicate that the features learned by our method not only can disentangle different classes as orthogonal subspaces but also can disentangle different visual attributes within each class as (orthogonal) principal components within each subspace.

\subsection{STL-10}
\label{app:stl-10}
\textbf{Settings.} For all experiments on STL-10, we follow the common training hyper-parameters in section~\ref{app:settings}. For \ours{}-Binary setting, we train 150,000 iterations. For \ours{}-Multi setting, we initialize the weights from the 20,000-th iteration of \ours{}-Binary checkpoint and train for another 80,000 iterations (with the \ours{}-Multi objective). The IS and FID scores on the STL-10 dataset are reported in Table \ref{tab:comparsion_full}, on par or even better than existing methods such as SNGAN \citep{miyato2018spectral} or DC-VAE \citep{parmar2021dual}.  

\begin{figure}
\centering
 \subfigure[Original $\X$]{
     \includegraphics[width=0.4\textwidth]{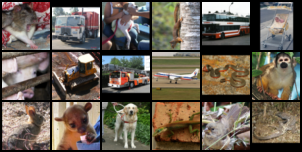}
 }
  \subfigure[Decoded $\hat \X$]{
     \includegraphics[width=0.4\textwidth]{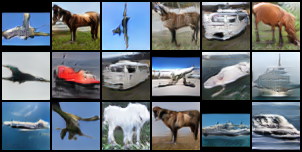}
 }
 \caption{Visualizing the original $\x$ and corresponding decoded $\hat \x$ results on STL-10 dataset. Note the model is trained from \ours{}-Binary \eqref{eq:MCR2-GAN-objective-binary} hence sample or class wise correspondence is relatively poor. But the decoded image quality is very good.}
 \label{app:stl-autoencode}
\end{figure}

\textbf{Visualizing auto-encoding property for \ours{}-Binary.} We visualize the original images $\x$ and their decoded $\hat{\x}$ using the LDR model learned from \ours{}-Binary objective. The results are shown in Figure~\ref{app:stl-autoencode} for STL-10.

\subsection{Celeb-A and LSUN}
\label{app:celeba_lsun}
To verify that our formulation works on images of higher-resolution, we conduct experiments on the Celeb-A and LSUN datasets, which have a resolution of $128\times 128$.

\textbf{Settings.} For all experiments on these datasets, we follow the common training hyper-parameters in Section~\ref{app:settings}. We choose a 300 mini-batch size for Celeb-A and LSUN. Both of them are trained with the  \ours{}-Binary formulation, and for 450,000 iterations. 

\begin{figure}
\centering
 \subfigure[Hat]{
     \includegraphics[width=0.3\textwidth]{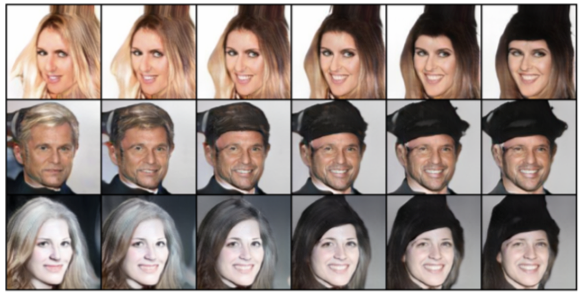}
 }
  \subfigure[Hair Color]{
     \includegraphics[width=0.3\textwidth]{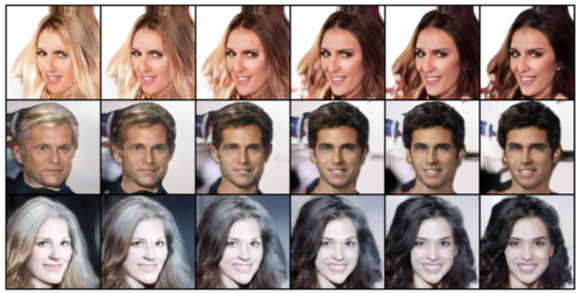}
 }
 \subfigure[Glasses]{
     \includegraphics[width=0.3\textwidth]{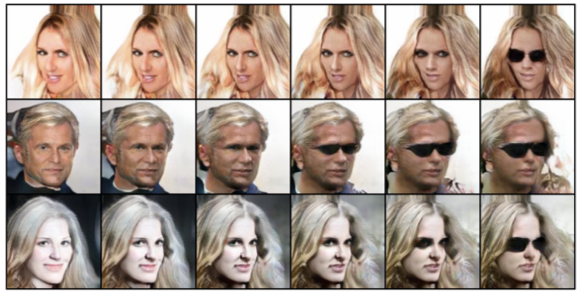}
 }
 \caption{Sampling along the 9-th, 19-th, and 23-th principal components of the learned features $\Z$ seems to manipulate the visual attributes for generated images, on the CelebA dataset.}
 \label{app::fig::sampling_along_pca_img-app}
\end{figure}

\textbf{Generating image along different PCA components.} We calculate the principal components of the learned features $Z$ in the latent subspace. We manually choose 3 principle components which are related to hat, hair color, and glasses (see Figure~\ref{app::fig::sampling_along_pca_img-app}). The three components are 9-th, 19-th, and 23-th respectively from the overall 128 principal components. These principal directions seem to clearly disentangle visual attributes/factors such as wearing hat, changing hair color, and wearing glasses. 

\begin{figure}[htbp]
\centerline{\includegraphics[width=0.7\textwidth]{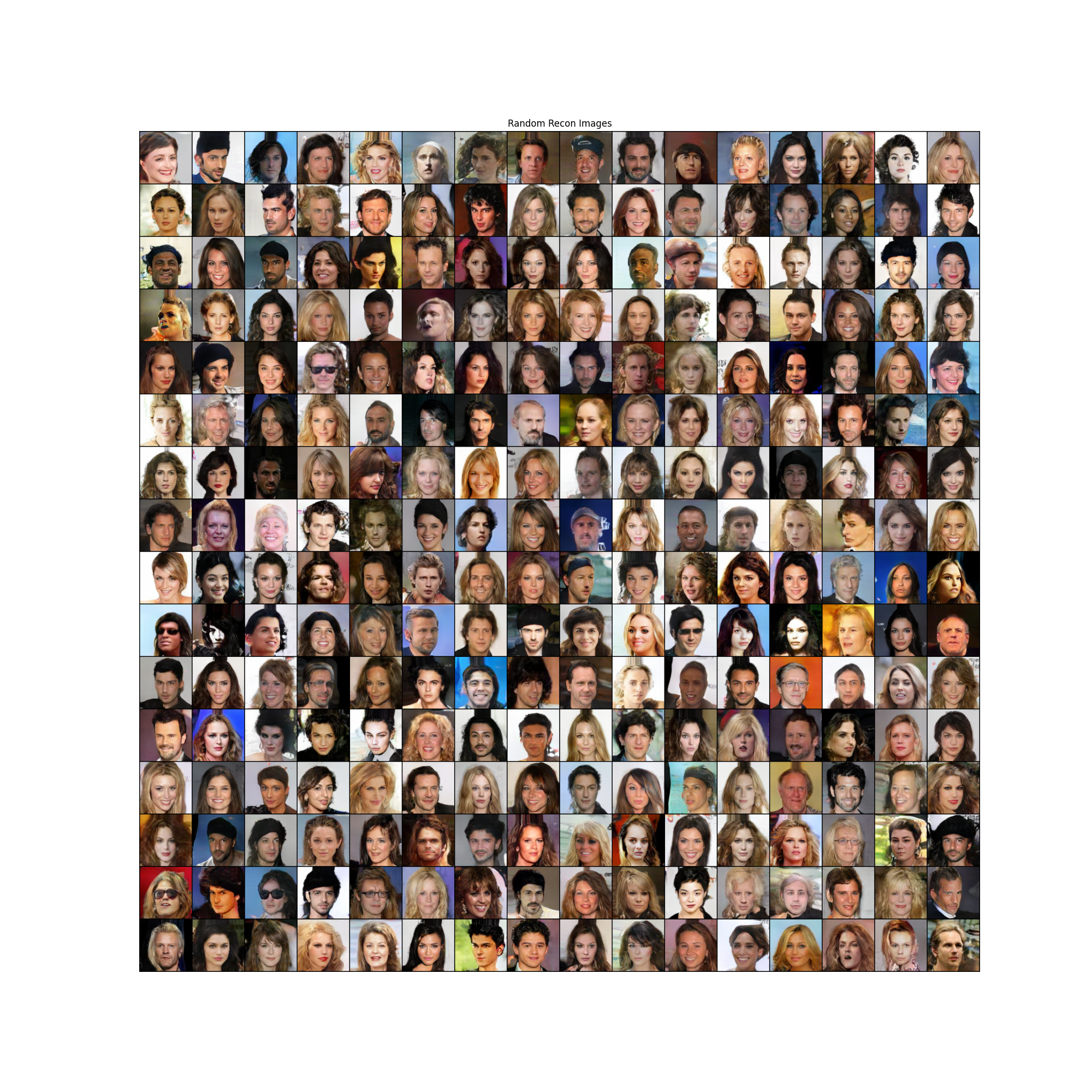}}
 \caption{Images decoded from randomly sampled features, as a learned Gaussian distribution \eqref{eqn:Gaussian-model}, for the CelebA dataset.}
 \label{app::fig::celeb_random_sample}
\end{figure}

\textbf{Images generated from random sampling of the feature space.} We sample $\z$ randomly according to the following Gaussian model: 
\begin{equation}
    \z_{random} = \Bar{\z} + \alpha \sum_{i=1}^{r} n_{i} * \sigma_i * \bm v_i,
\label{eqn:Gaussian-model}    
\end{equation}
where $\Bar{\z}$ is the mean feature, $\sigma_i$ and $\bm v_i$ are the $i$th singular value and singular vector, $n_i$ are i.i.d. Gaussian $\mathcal{N}(0,1)$ random variables. As before $\alpha$ is a hyper-parameter to control the sampling range. We use top r=100 principle components for random sampling. The random generated images are realistic and diverse. (see Figure~\ref{app::fig::celeb_random_sample})

\begin{figure}
\centering
 \subfigure[Original $\X$]{
     \includegraphics[width=0.4\textwidth]{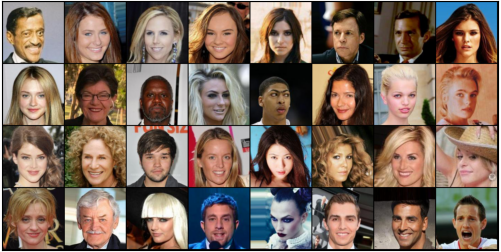}
 }
  \subfigure[Decoded $\hat \X$]{
     \includegraphics[width=0.4\textwidth]{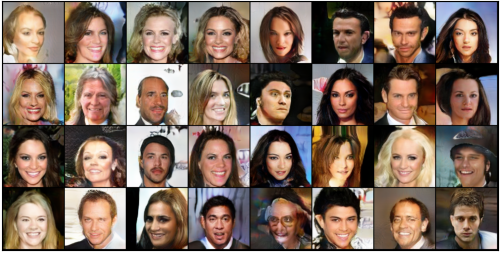}
 }
 \caption{Visualizing the original $\x$ and corresponding decoded $\hat \x$ results on Celeb-A dataset. The LDR model is trained from \ours{}-Binary \eqref{eq:MCR2-GAN-objective-binary}.}
 \label{app:face_auto_encode}
\end{figure}

\begin{figure}
\centering
 \subfigure[Original $\X$]{
     \includegraphics[width=0.4\textwidth]{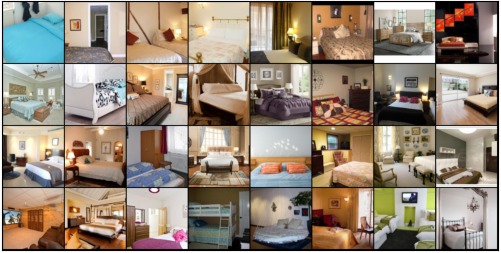}
 }
  \subfigure[Decoded $\hat \X$]{
     \includegraphics[width=0.4\textwidth]{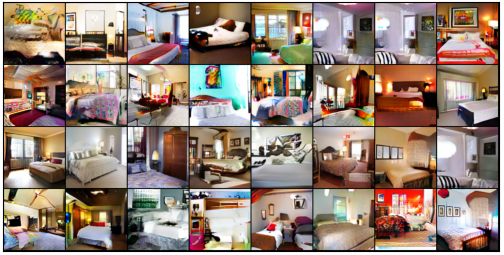}
 }
 \caption{Visualizing the original $\x$ and corresponding decoded $\hat \x$ results on LSUN-bedroom dataset. The LDR model is trained from \ours{}-Binary \eqref{eq:MCR2-GAN-objective-binary}.}
 \label{app:lsun_auto_encode}
\end{figure}

\textbf{Visualizing auto-encoding property for \ours{}-Binary.} We visualize the original image $\x$ and their decoded $\hat{\x}$ using the LDR model learned from  \ours{}-Binary formulation. The results are shown in Figure~\ref{app:face_auto_encode} and Figure~\ref{app:lsun_auto_encode} for the Celeb-A dataset and the LSUN dataset, respectively. The \ours{}-Binary formulation can give very good visual quality for $\hat{\x}$ but cannot ensure sample to sample alignment. Nevertheless, the decoded  $\hat{\x}$ seems to be very similar to the original $\x$ in some main visual attributes. We believe the binary formulation manages to align only the dominant principal component(s) associated with the most salient visual attributes, say pose of the face for Celeb-A or layout of the room for LSUN, between features of $\X$ and $\hat \X$.  

\subsection{ImageNet}
\label{app:imagenet}

\textbf{Settings.} To verify that our formulation works on large-scale datasets, we train a model on the entire ImageNet dataset. For all experiments on ImageNet, we follow the common training hyper-parameters in Section~\ref{app:settings}. 

We first train our model with the \ours{}-Binary objective \eqref{eq:MCR2-GAN-objective-binary} with a mini-batch size 1800 on the whole ImageNet ILSVRC 2012 dataset. The number of training iterations is 450,000.

After that, we fine-tune the Binary-pretrained model with \ours{}-Multi \eqref{eq:MCR2-GAN-objective}, on 10 selected classes. Information about the 10  classes can be found in Table~\ref{tab:imagenet10}. The fine-tune mini-batch size is 1024, and we train another 35,000 iterations. This experiment took 8 A100-SXM4 GPUs, each with 40GB of CUDA memory for 120 GPU hours. We also include results from another run that is trained for 200,000  iterations with an even larger batch of size 1800. Note that our choice of mini-batch size is substantially larger than those commonly adopted in other works while training on the ImageNet (e.g. 128 in \cite{miyato2018spectral}).
We empirically observe that training with a larger mini-batch generates images of better quality and clearer class alignment.
This is consistent with the proposed \ours{} framework as the \ours{}-Multi objective explicitly encourages alignment of class distributions, therefore benefiting from a larger batch that better captures overall data distributions.
We leave a more rigorous study of the effect of batch size for future work.

Due to the heavy computation of such large batch size, we present the following intermediate results obtained at these early iterations whereas most existing methods run with significantly larger number of iterations. Nevertheless, these intermediate results already verify the efficacy of our framework. In addition, we present the full version of the comparison with existing generative methods in Table~\ref{tab:comparsion_full}. We see the SI and FID scores for LDR-Multi degraded a little after the finetuning. This is expected as learning a more refined separation and alignment of 10 classes is a more challenging task than 2 classes. This is consistently observed from experiments on other datasets too. 

\begin{figure}[t]
\subfigure[$|\Z^\top\Z|$]{
    \includegraphics[width=0.48\textwidth]{Imagenet_heatmat_epoch200000.png}
}
\subfigure[$|\Z^\top\hat{\Z}|$]{
    \includegraphics[width=0.48\textwidth]{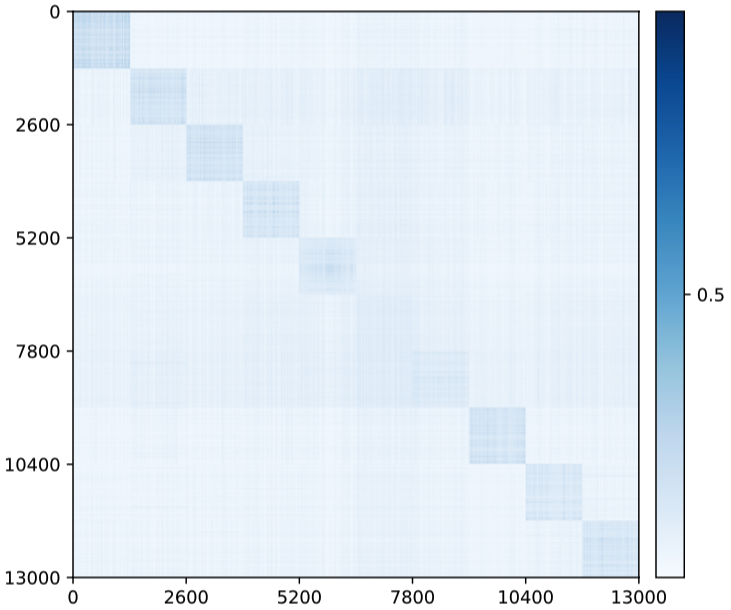}
}
\caption{Visualizing feature alignment: (a) among features $|\Z^\top\Z|$, (b) between features and decoded features $|\Z^\top\hat{\Z}|$. These results obtained after 200,000 iterations.}
\label{app:heatmaps_imagenet}
\end{figure}

\textbf{Visualizing feature similarity for \ours{}-Multi.} We visualize the cosine similarity among features $\Z$ of different classes learned from the \ours{}-Multi objective in Figure~\ref{app:heatmaps_imagenet}. In addition, we provide the visualization of alignment between features $\Z$ and decoded features features $\hat{\Z}$. These results demonstrate that not only the encoder has already learnt to discriminate between classes, $\Z$ and $\hat{\Z}$ also are aligned clearly within each class. 

\begin{table}[t]
    \centering
    \small
    \setlength{\tabcolsep}{6.5pt}
    \renewcommand{\arraystretch}{1.25}
    \begin{tabular}{l|cc|cc|cc}
    \multirow{2}{*}{Method} & \multicolumn{2}{c|}{CIFAR-10} & \multicolumn{2}{c|}{STL-10} & \multicolumn{2}{c}{ImageNet}\\
    \cline{2-7}
     ~  & IS$\uparrow$ & FID$\downarrow$ & IS$\uparrow$ & FID$\downarrow$ & IS$\uparrow$ & FID$\downarrow$ \\
    \hline
    \hline
    \textit{GAN based methods}          &  ~   & ~    & ~    & ~    & ~    & ~   \\
    DCGAN \citep{radford2015unsupervised}& 6.6  & -    & 7.8  & -    & -    & -   \\
    SNGAN \citep{miyato2018spectral}     & 7.4  & 29.3 & \textbf{9.1}  & 40.1 & -    & 48.73   \\
    CSGAN \citep{wu2019deep} & 8.1  & 19.6 & -    & -    & -    & -   \\
    LOGAN \citep{wu2019logan}            & \textbf{8.7} & \textbf{17.7} & -    & -    & -    & -   \\
    \hline
    \hline
    \textit{VAE/GAN based methods}    & ~       & ~    & ~    & ~    & ~    & ~   \\
    VAE    \citep{kingma2013auto}      & 3.8     & 115.8& -    & -    & -    & -   \\
    VAE/GAN \citep{larsen2016autoencoding}  & 7.4& 39.8 & -    & -    & -    & -   \\
    NVAE   \citep{vahdat2020nvae}      & -       & 50.8 & -    & -    & -    & -   \\
    DC-VAE \citep{parmar2021dual}      & \textbf{8.2}  & \textbf{17.9} & 8.1  & 41.9 & -    & -   \\
    LDR-Binary (ours)                 & \textbf{8.1}  & \textbf{19.6} & 8.4  & \textbf{38.6} &  7.74  & \textbf{46.95}  \\
    LDR-Multi (ours)                  & 7.1     & 23.9 & 7.7   & 45.7   & 6.44   & 55.51  \\
    \end{tabular}      
    \caption{\small Comparison on CIFAR-10, STL-10, and ImageNet.}
    \label{tab:comparsion_full}
\end{table}

\begin{figure}
\centering
 \subfigure[Original $\X$]{
     \includegraphics[width=0.48\textwidth]{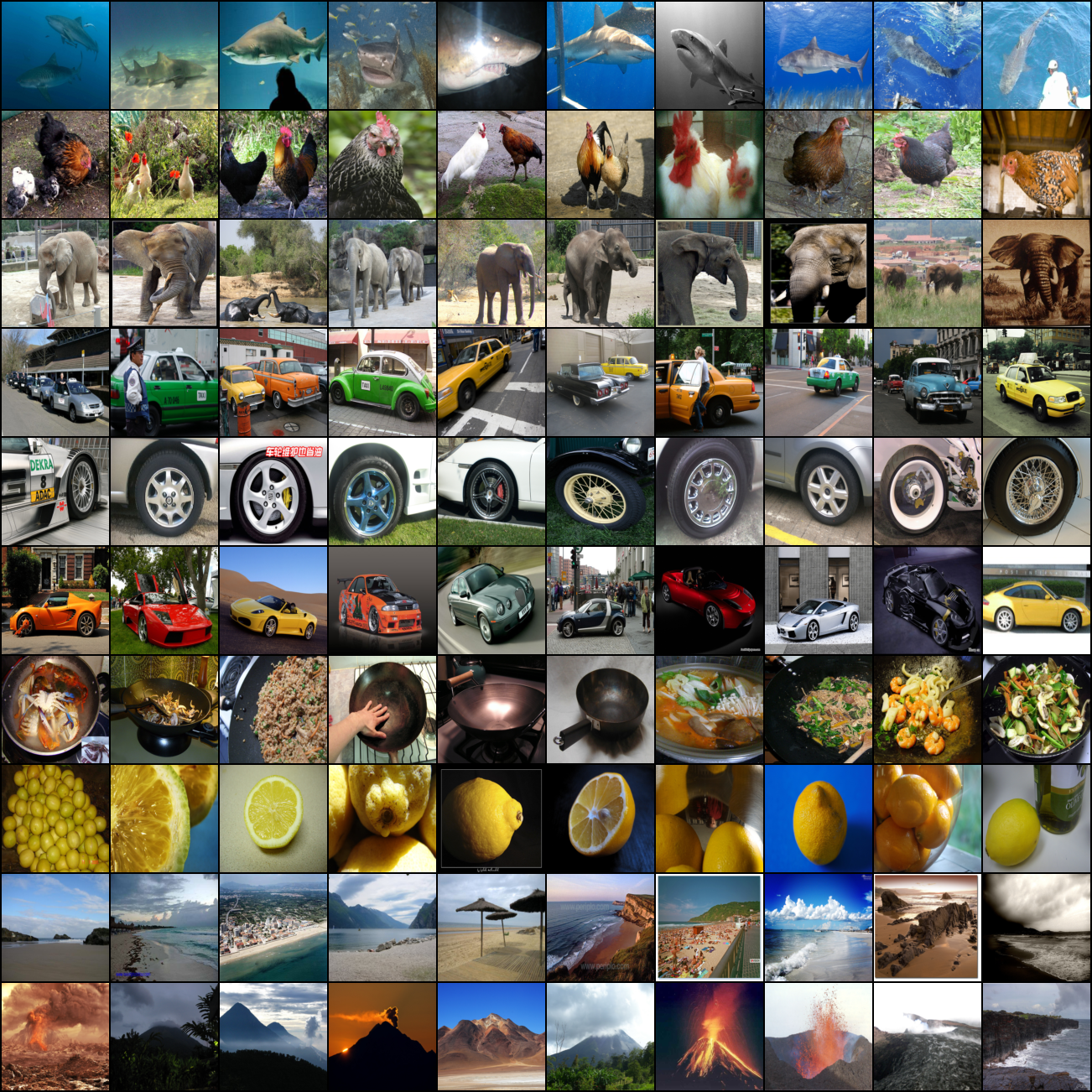}
 }
  \subfigure[Decoded $\hat \X$]{
     \includegraphics[width=0.48\textwidth]{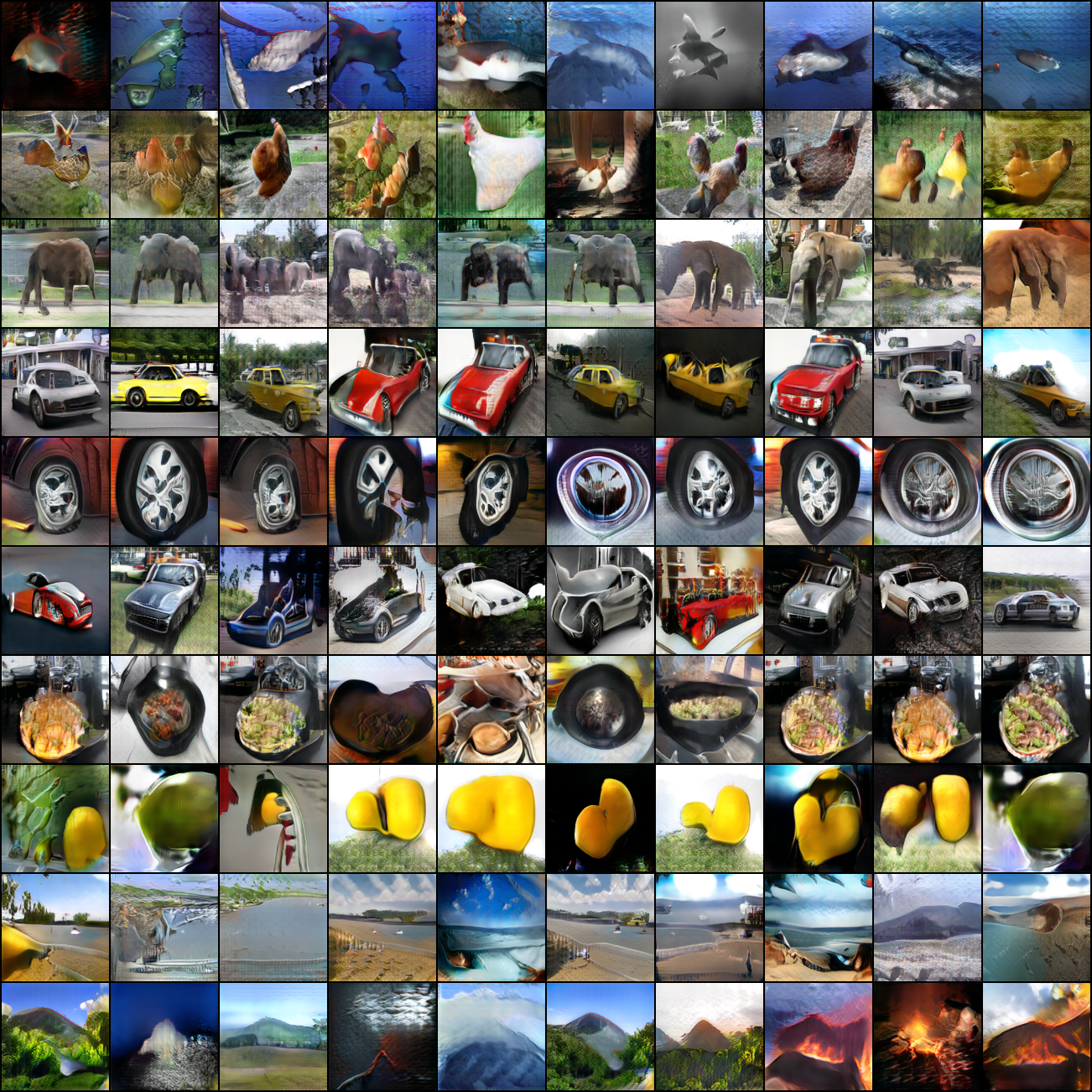}
 }
 \caption{Visualizing the original $\x$ and corresponding decoded $\hat \x$ results on ImageNet (10 classes). The LDR model fine-tuned using \ours{}-Multi \eqref{eq:MCR2-GAN-objective}. These visualizations are obtained after 35,000 iterations.}
 \label{app:imagenet_auto_encode}
\end{figure}

\textbf{Visualizing auto-encoding property for \ours{}-Multi.} We visualize the original image $\x$ and their decoded $\hat{\x}$ using the LDR model fine-tuned using \ours{}-Multi formulation. The results are shown in Figure~\ref{app:imagenet_auto_encode} for the selected 10 classes in ImageNet. The \ours{}-Multi formulation can give good visual quality for $\hat{\x}$ as well as decent sample-to-sample alignment.

\subsection{Ablation Study on Closed-Loop Framework and Objective Functions}
\label{app:ablation-objective}

To empirically validate the necessity and respective roles of the closed-loop framework and the rate reduction ($\Delta R$) loss, we conduct two sets of experiments. For the first set of experiments, we modify our closed-loop framework by instantiating more than two networks while keeping the objective function \eqref{eq:MCR2-GAN-objective} unchanged.
For the second set of experiments, we keep the closed-loop formulation but replace all rate reduction ($\Delta R$) loss terms in \eqref{eq:MCR2-GAN-objective} with corresponding cross-entropy loss, or remove some of the terms. Experiments here shed insight on how the closed-loop framework and the rate reduction  affect separately on the performance, including sample-wise reconstruction, the alignment of $\Z$ and $\hat{\Z}$ space, and the diversity of intra-class features. 

\subsubsection{The Importance of the Closed-Loop}
\label{app:closed-loop}

To evaluate the importance of a true closed-loop architecture, we experiment on modified versions of our closed-loop system \eqref{app:eq:LDR}. Notice that many architectures have been proposed and experimented before to promote the encoder $f$ and decoder $g$ to be mutually inverse or cycle consistent (at least for mappings between the data and feature distributions), such as  BiGAN \citep{donahue2016adversarial}, VAE-GAN \citep{VAE-GAN}, and CycleGAN \citep{zhu2017unpaired}. However, the cycle consistency is typically enforced through a third discriminator network.\footnote{In the case of CycleGAN \citep{zhu2017unpaired}, one needs two additional discriminator networks, one for each domain.}  

Here, we experiment to test if similar ideas work for the rate reduction framework. First, we break the closed-loop and use a different encoder network $f^2: \hat{\X} \rightarrow \hat{\Z}$ to replace the original encoder $f$. The workflow is summarized in the diagram \eqref{app:eq:LDR_not_share_f}. Second, to emulate the architecture of VAE-GAN \citep{VAE-GAN}, we also instantiate an extra encoder network $f^2$ and compute $\mathcal{T}$ using $\tilde{\Z}$ and $\hat{\Z}$. Overall workflow is also summarized in the diagram \eqref{app:eq:vae_LDR_objective}.

\begin{equation}
    \X \xrightarrow{\hspace{2mm} f(\x, \theta)\hspace{2mm}} \Z \xrightarrow{\hspace{2mm} g(\z,\eta) \hspace{2mm}} \hat \X  \xrightarrow{\hspace{2mm} f(\x, \theta)\hspace{2mm}} \ \hat \Z;
    \label{app:eq:LDR}
\end{equation}
\begin{equation}
    \X \xrightarrow{\hspace{2mm} f^{1}(\x, \theta^1)\hspace{2mm}} \Z \xrightarrow{\hspace{2mm} g(\z,\eta) \hspace{2mm}} \hat{\X}  \xrightarrow{\hspace{2mm} f^{2}(\x, \theta^2)\hspace{2mm}} \ \hat{\Z};
    \label{app:eq:LDR_not_share_f}
\end{equation}
\begin{equation}
    \X \xrightarrow{\hspace{2mm} f^{1}(\x, \theta^1)\hspace{2mm}} {\Z} \xrightarrow{\hspace{2mm} g(\z,\eta) \hspace{2mm}} \hat \X, \bm X   \xrightarrow{\hspace{2mm} f^{2}(\x, \theta^2)\hspace{2mm}} \ \hat{\Z}, \tilde{\bm Z}.
    \label{app:eq:vae_LDR_objective}
\end{equation}

\begin{figure}
\centering
 \subfigure[Input]{
     \includegraphics[width=0.41\textwidth]{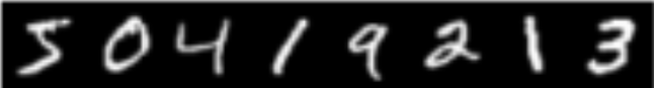}
 }
 \subfigure[$\hat{\X}$ of \ours{}-Multi \eqref{app:eq:LDR}]{
     \includegraphics[width=0.41\textwidth]{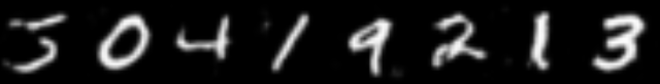}
 }
 \subfigure[$\hat{\X}$ of framework \eqref{app:eq:LDR_not_share_f}]{
     \includegraphics[width=0.41\textwidth]{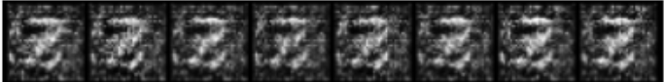}
 }
 \subfigure[$\hat{\X}$ of framework \eqref{app:eq:vae_LDR_objective}]{
     \includegraphics[width=0.41\textwidth]{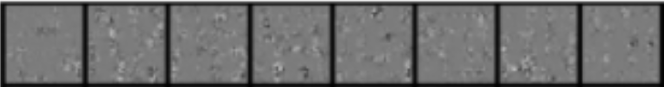}
 }
 \caption{Qualitative ablation study on the influence of different frameworks to \ours{}.}
 \label{app:viz_ablation_of_framework}
\end{figure}

We run the three experiments on MNIST, and choose architecture from Table~\ref{arch:mnist_g} for the encoder and Table~\ref{arch:mnist_d} for the decoder, and the training hyper-parameters follow Section~\ref{app:settings}. The qualitative results are shown in Figure~\ref{app:viz_ablation_of_framework}. Both frameworks \eqref{app:eq:LDR_not_share_f} and \eqref{app:eq:vae_LDR_objective} failed to generate meaningful images. These experiments show that directly applying rate reduction objectives without the closed-loop or architectures that loosely enforcing cycle-consistency fails to work. Instead, the closed-loop formulation allow us to use only two networks, without the need of any extra network.

\subsubsection{The Importance of Rate Reduction}
\label{app:objective-ablation}

To replace the rate reduction ($\Delta R$) terms in the objective function \eqref{eq:MCR2-GAN-objective} with cross-entropy, we introduce a linear mapping $ \bm{W} \in \Re^{d \times k}$ to map $\Z \in \Re^{d \times n}$ from feature space to logits $\gamma = \Z^{\top} \bm{W}$. We then calculate the softmax cross-entropy function on logits $\gamma$ and one hot label matrix $\bm Y$. Here $\mathcal{H}(\gamma, \bm Y) = \sum_{i=1}^{n}\sum_{j=1}^{k} Y_{ij} \log{\frac{e^{\gamma_{ij}}}{\sum_{j=1}^{k} e^{\gamma_{ij}}}}$ is the formulation of softmax cross-entropy function and $\bm Y \in \Re^{n \times k}$ is one-hot label matrix. Then, we can replace the first two terms of \eqref{eq:MCR2-GAN-objective} ($\Delta R\big(\Z \big)$ and $\Delta R\big(\hat{\Z}\big)$) with $\mathcal{H}(\Z^{\top} \bm{W}, \bm Y)$ and $\mathcal{H}(\hat{\Z}^{\top} \bm{W}, \bm Y)$. For the third term of \eqref{eq:MCR2-GAN-objective}, we extract $j$-th class one-hot feature $\gamma_j = \Z_j^{\top} \bm{W}$, $\hat{\gamma}_j = \hat{\Z}_j^{\top} \bm{W}$ from $\Z$ and $\hat{\Z}$, and define the distance $\mathcal{D}(\gamma_j, \hat{\gamma}_j)=\frac{e^{\gamma_j}}{e^{\gamma_j}+e^{\hat{\gamma}_j}}$ of them. 

For the third term of \eqref{eq:MCR2-GAN-objective}, we further introduce $k$ linear layers as discriminators $\{\mathcal{D}_j\}_{j=1}^{k}$ for each class. Then, we replace the third term with the GAN's objective function as  $\sum_{j=1}^{k}\mathbb{E}[\log{\mathcal{D}_j(\Z_j)}]+\mathbb{E}[\log(1-\mathcal{D}_j(\hat{\Z_j}))]$ \footnote{$\mathbb{E}[\X]$ denote the expectation of $\X$.}. Now, we have the cross-entropy version objective function \eqref{app::eq:ce-GAN-objective} for closed loop framework. We denote the closed loop framework with cross-entropy as Closed-loop-CE.

\begin{align}
\min_{\eta} \max_{\theta, \bm{W}, \mathcal{D}}  \mathcal{T}_{\X}(\theta, \eta, \bm{W}, \mathcal{D}) \; \doteq \; & \mathcal{H}\big(\Z^{\top}\bm{W}, \bm Y \big) + \mathcal{H}\big(\hat{\Z}^{\top}\bm{W},\bm Y\big) + \nonumber \\  &\sum_{j=1}^{k}\mathbb{E}[\log{\mathcal{D}_j(\Z_j)}]+\mathbb{E}[\log(1-\mathcal{D}_j(\hat{\Z_j}))].
\label{app::eq:ce-GAN-objective}
\end{align}

We run the experiments on MNIST and CIFAR10. The architectures of MNIST and CIFAR10 are given in  Table~\ref{arch:mnist_g} to Table~\ref{arch:cifar_d}\footnote{In the context of this section, we use the term Decoder and Generator interchangeably; similarly for Encoder and Discriminator.}.

\begin{figure}[t]
\centering
 \subfigure[Original $\X$]{
     \includegraphics[width=0.31\textwidth]{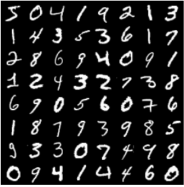}
     \label{ablation::mnist:reconstructionComparison_a}
 }
  \subfigure[$\hat{\X}$ by Closed-loop-CE]{
     \includegraphics[width=0.31\textwidth]{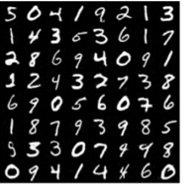}
     \label{ablation::mnist:reconstructionComparison_b}
 }
 \subfigure[$\hat{\X}$ by \ours{}-Multi]{
     \includegraphics[width=0.31\textwidth]{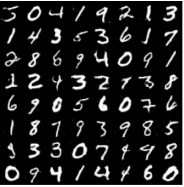}
     \label{ablation::mnist:reconstructionComparison_c}
 }
 \caption{The comparison of sample-wise reconstruction between Close-loop-CE and \ours{}-Multi. }
 \label{ablation::mnist:reconstructionComparison}
\end{figure}

\begin{figure}[t]
\centering
 \subfigure[Closed-loop-CE]{
     \includegraphics[width=0.42\textwidth]{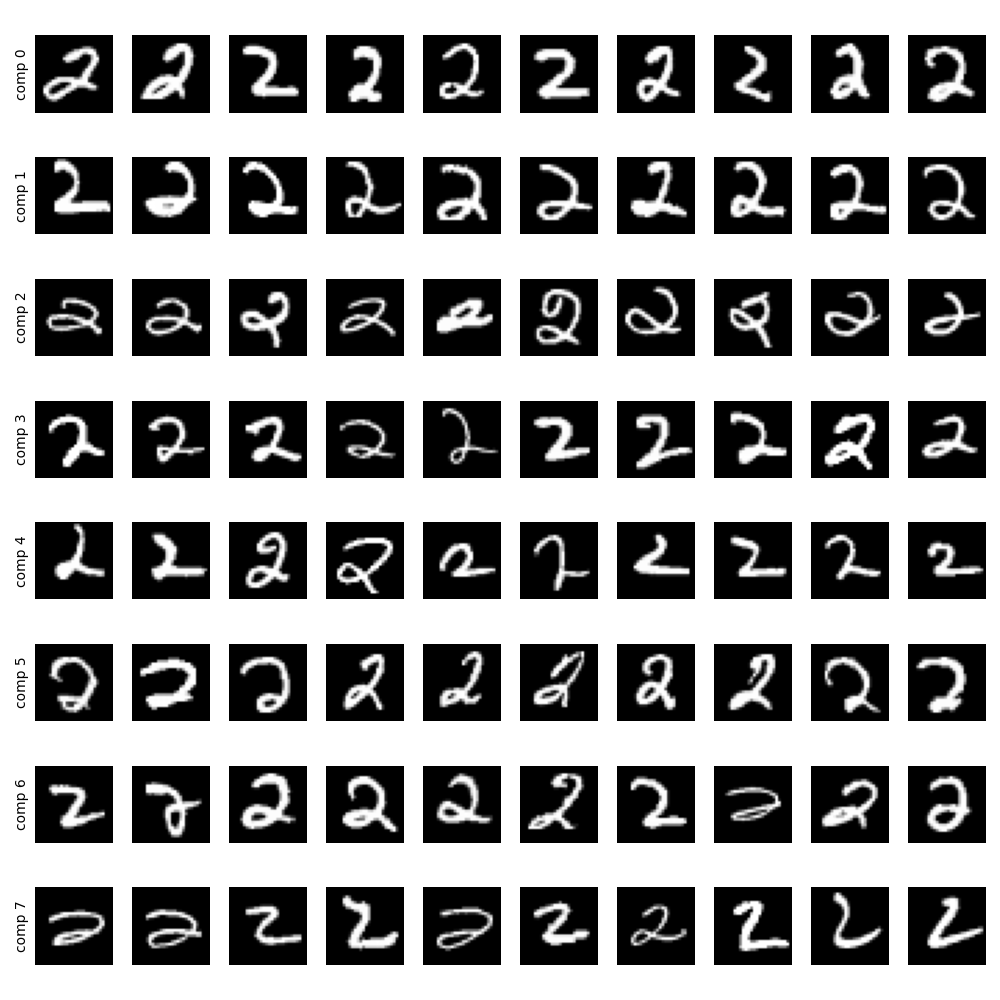}
     \label{ablation::mnist:pca_components_a}
 }
 \subfigure[\ours{}-Multi]{
     \includegraphics[width=0.42\textwidth]{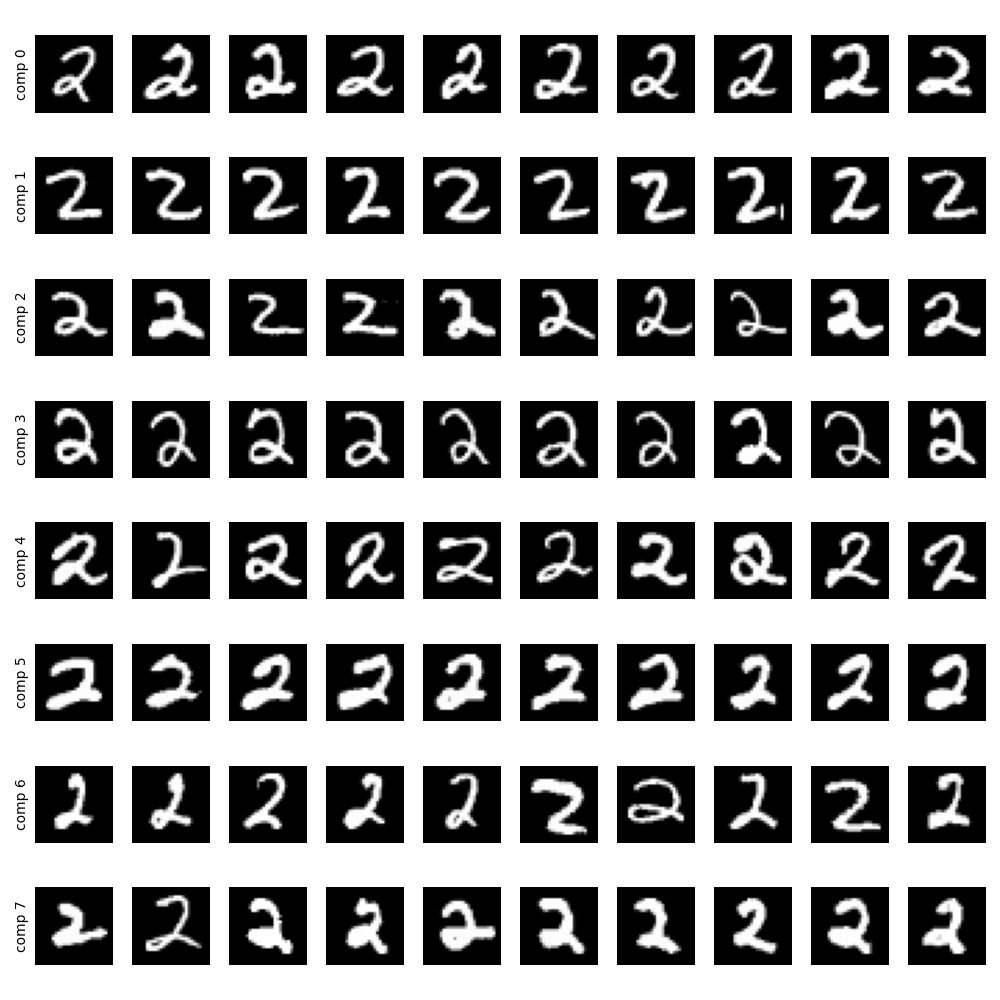}
     \label{ablation::mnist:pca_components_b}
 }
 \caption{Training samples along different principal components of the learned features of digit `2'.}
 \label{ablation::mnist:pca_components}
\end{figure}

\begin{figure}[t]
\centering
 \subfigure[$|\Z^\top \hat{\Z}|$ from Closed-loop-CE]{
     \includegraphics[width=0.42\textwidth]{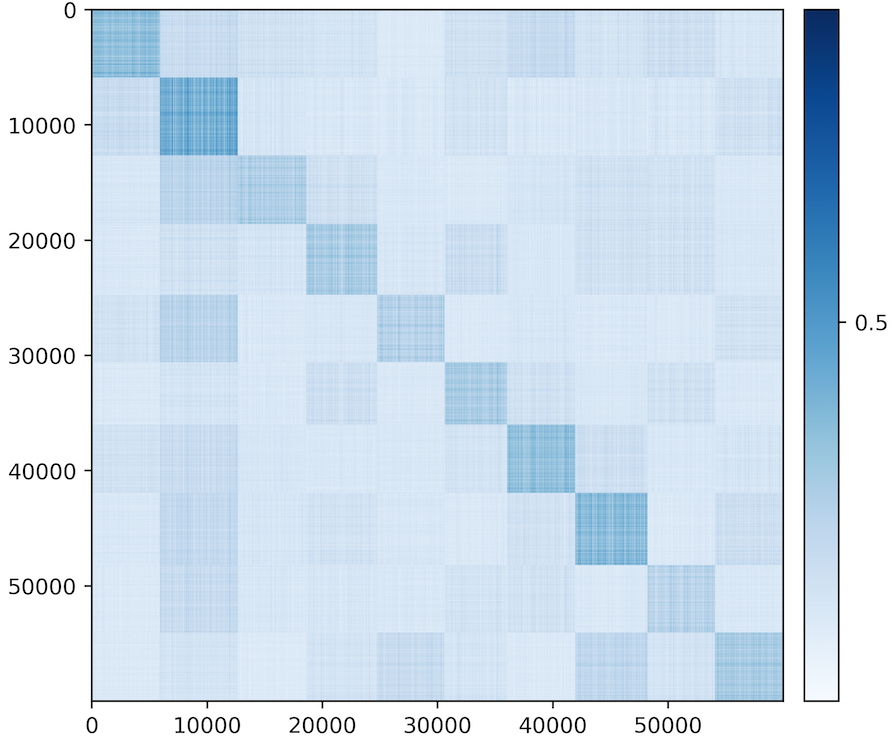}
     \label{ablation::mnist:zzhat_pca_singularv_a}
 }
 \subfigure[$|\Z^\top \hat{\Z}|$ from \ours{}-Multi]{
     \includegraphics[width=0.42\textwidth]{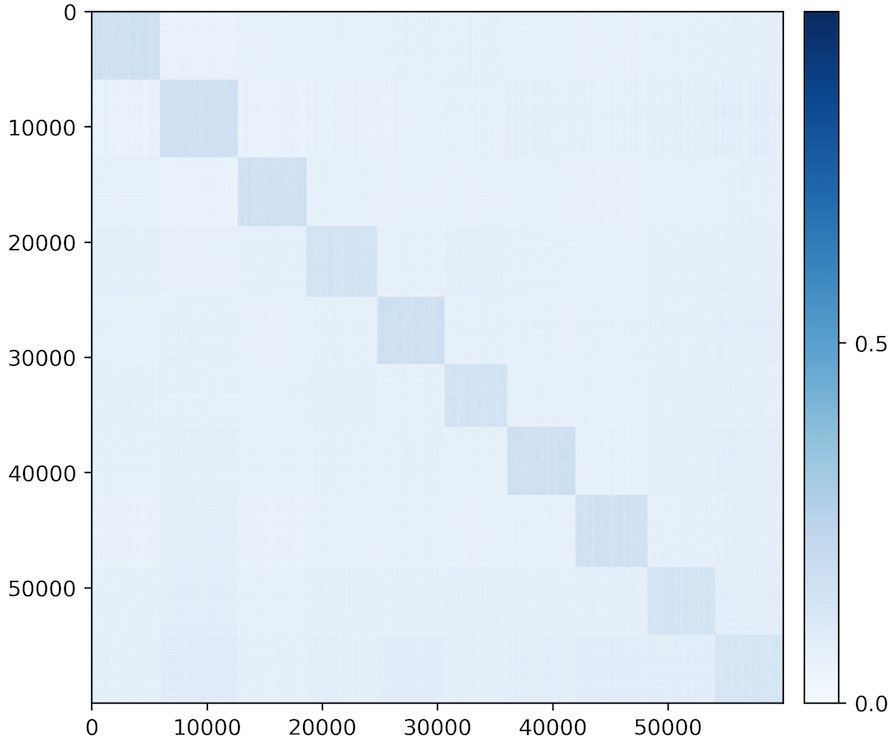}
     \label{ablation::mnist:zzhat_pca_singularv_b}
 }
 \subfigure[PCA: Closed-loop-CE learned feature for every class]{
     \includegraphics[width=0.42\textwidth]{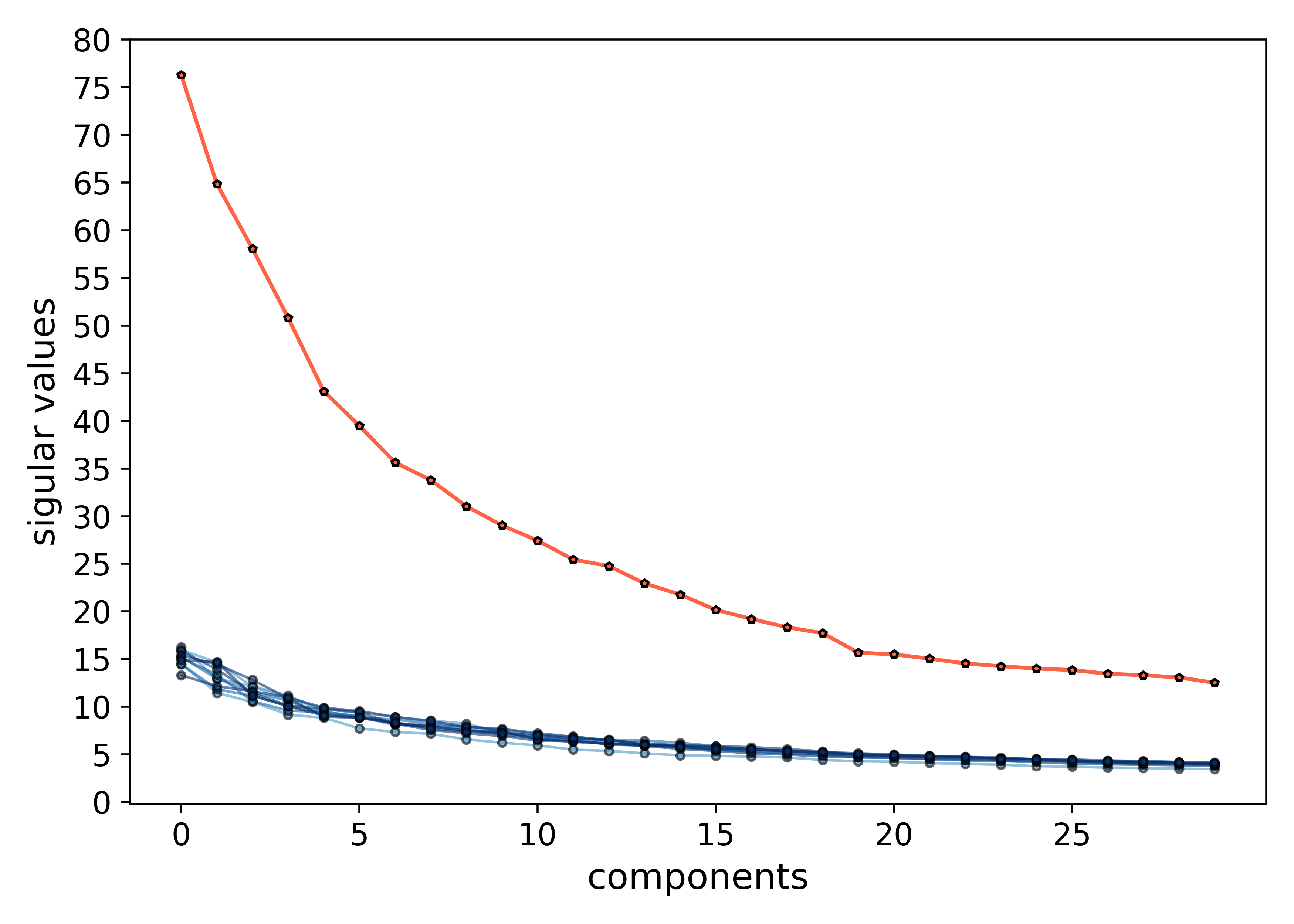}
     \label{ablation::mnist:zzhat_pca_singularv_c}
 }
 \subfigure[PCA: \ours{}-Multi learned feature for every class]{
     \includegraphics[width=0.42\textwidth]{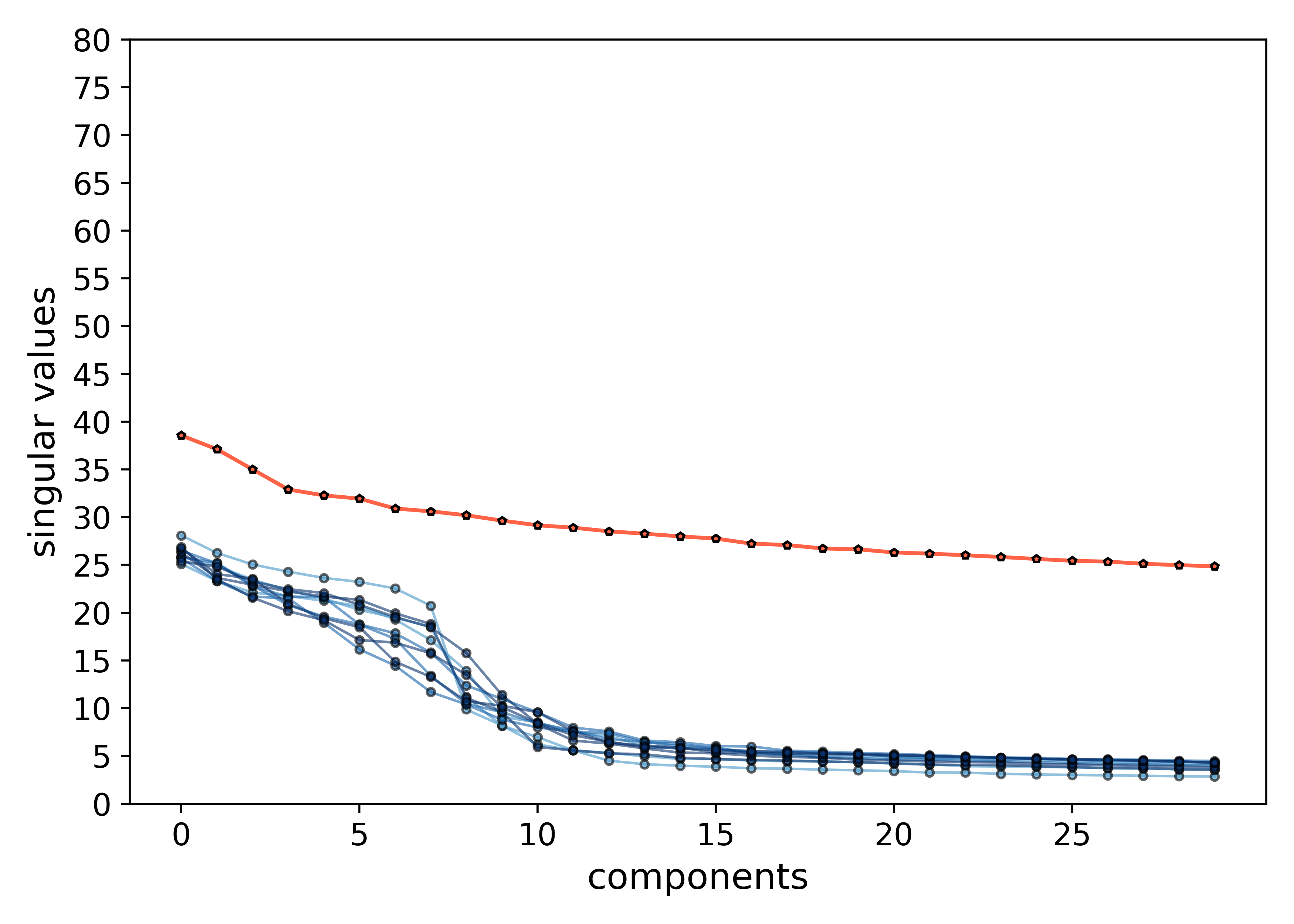}
     \label{ablation::mnist:zzhat_pca_singularv_d}
 }
 \caption{Comparison Closed-loop-CE and \ours{}-Multi on $|\Z^\top \hat{\Z}|$ and PCA singular values.}
 \label{ablation::mnist:zzhat_pca_singularv}
\end{figure}

\textbf{Results on MNIST.} The training hyper-parameters of \ours{}-Multi and Closed-loop-CE on MNIST are following Section~\ref{app:settings}. Comparisons between \ours{}-Multi and Closed-loop-CE are listed in Figure~\ref{ablation::mnist:reconstructionComparison}, \ref{ablation::mnist:pca_components}, and \ref{ablation::mnist:zzhat_pca_singularv}.

Figure~\ref{ablation::mnist:reconstructionComparison_b} and \ref{ablation::mnist:reconstructionComparison_c} show the reconstructed images $\hat{\X}$ from Closed-loop-CE and \ours{}-Multi. {Both methods can give sample-wise reconstruction results due to the transcription framework. However, comparing training images whose features are best aligned with the principal components of class `2' in Figure~\ref{ablation::mnist:pca_components}, we see that the principal components of CE features do not correspond to consistent visual attributes of the images, whereas ours do.}

From the heatmaps in Figure~\ref{ablation::mnist:zzhat_pca_singularv_a} and \ref{ablation::mnist:zzhat_pca_singularv_b}, we see the features learned by rate reduction possess clear orthogonal subspace structures, whereas those learned by Closed-loop-CE do not. {Moreover, Figure~\ref{ablation::mnist:zzhat_pca_singularv_c} and \ref{ablation::mnist:zzhat_pca_singularv_d} show that the learned features of \ours{}-Multi have higher singular values for the top principal components of each class, corresponding to a more linearized and diverse feature distribution, whereas those by Closed-loop-CE do not.}

\textbf{Failed Attempts on CIFAR-10 with Cross Entropy.}
The training hyper-parameters of Closed-loop-CE on CIFAR10 follow Section~\ref{app:settings}. We do the grid search on three hyper-parameters: learning rate $\{1.5\times 10^{-2}, 1.5\times 10^{-3}, 1.5\times 10^{-4}\}$, mini batch size (800 or 1600), and inner loop (1,2,3,4), conduct 24 experiments in total. All cases of the Closed-loop-CE fail to converge or experience model collapse on the CIFAR-10 dataset.

\subsubsection{Ablation Study of the Rate Reduction Terms}
\label{app:rate-reduction-terms}
\begin{figure}[t]
\centering
 \subfigure[Input]{
     \includegraphics[width=0.21\textwidth]{ablation_1_input.png}
     \label{ablation::mnist:objec_xhat_a}
 }
 \subfigure[$\hat{\X}$ from objective I]{
     \includegraphics[width=0.21\textwidth]{ablation_1_LDR.png}
     \label{ablation::mnist:objec_xhat_b}
 }
 \subfigure[$\hat{\X}$ from objective II]{
     \includegraphics[width=0.213\textwidth]{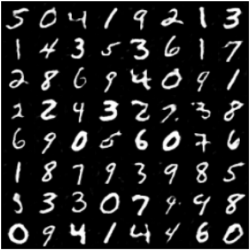}
     \label{ablation::mnist:objec_xhat_c}
 }
 \subfigure[$\hat{\X}$ from objective III]{
     \includegraphics[width=0.216\textwidth]{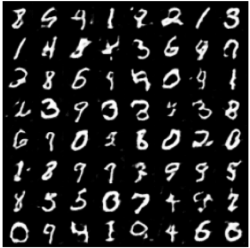}
     \label{ablation::mnist:objec_xhat_d}
 }
 \caption{Decoded images $\hat{\X}$ from objective I, II, or III. The influence of objective function on the reconstruction of \ours{}.}
 \label{ablation::mnist:objec_xhat}
\end{figure}

\begin{figure}[t]
\centering
 \subfigure[Objective I]{
     \includegraphics[width=0.31\textwidth]{ablation_1_zzhat_LDR.png}
 }
 \subfigure[Objective II]{
     \includegraphics[width=0.31\textwidth]{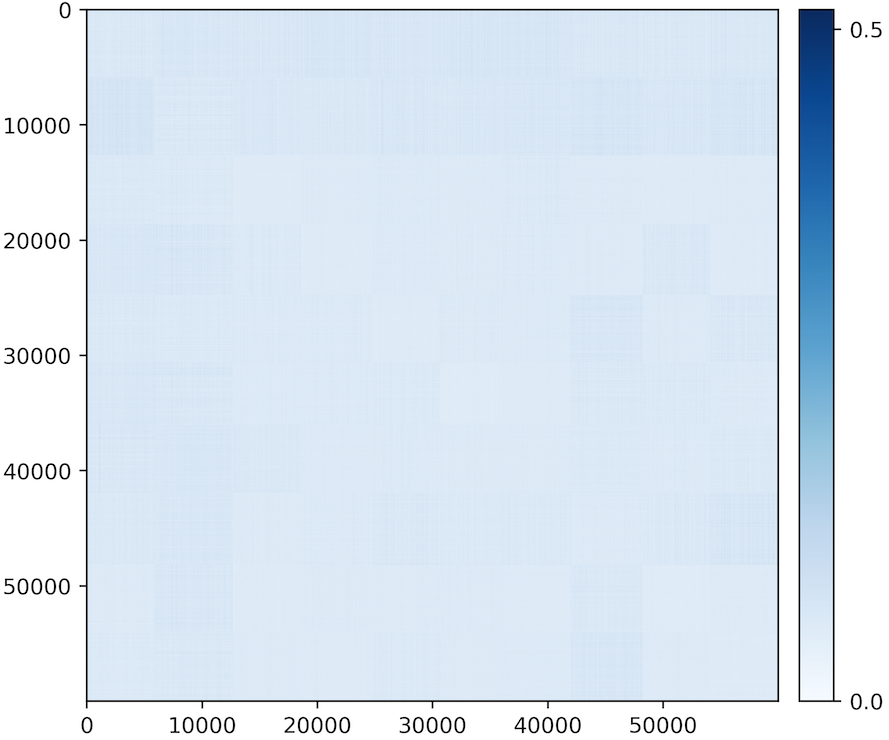}
 }
 \subfigure[Objective III]{
     \includegraphics[width=0.31\textwidth]{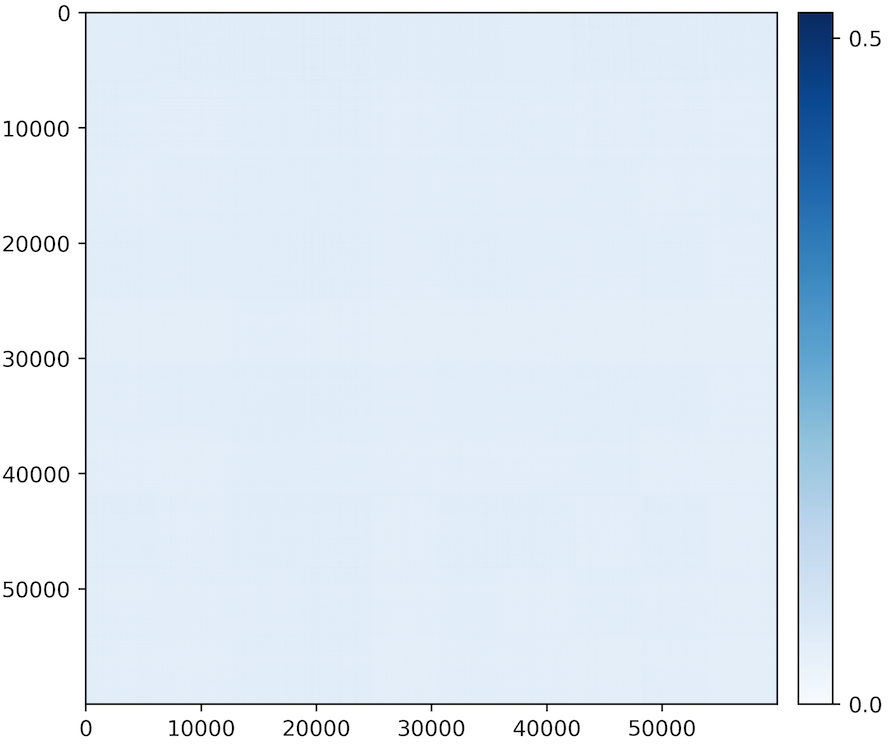}
 }
 \caption{Correlation $|\Z^\top \hat{\Z}|$ between features $\Z$ and $\hat{\Z}$ learned with Objective I, II, or III.}
 \label{ablation::mnist:objec_zzhat}
\end{figure}

\begin{figure}[t]
\centering
 \subfigure[Objective I]{
     \includegraphics[width=0.31\textwidth]{ablation_1_singularV_LDR.png}
 }
 \subfigure[Objective II]{
     \includegraphics[width=0.31\textwidth]{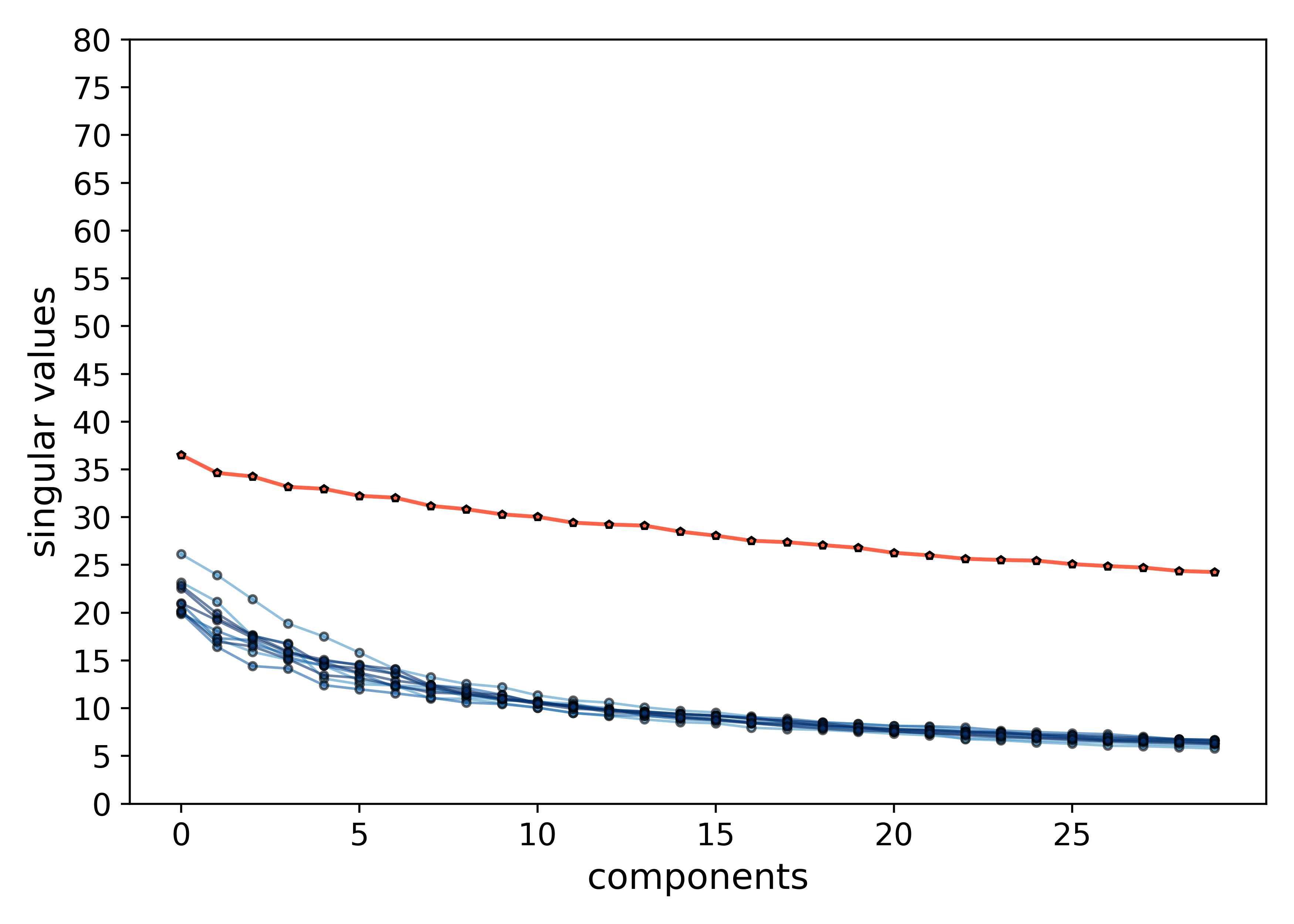}
 }
 \subfigure[Objective III]{
     \includegraphics[width=0.31\textwidth]{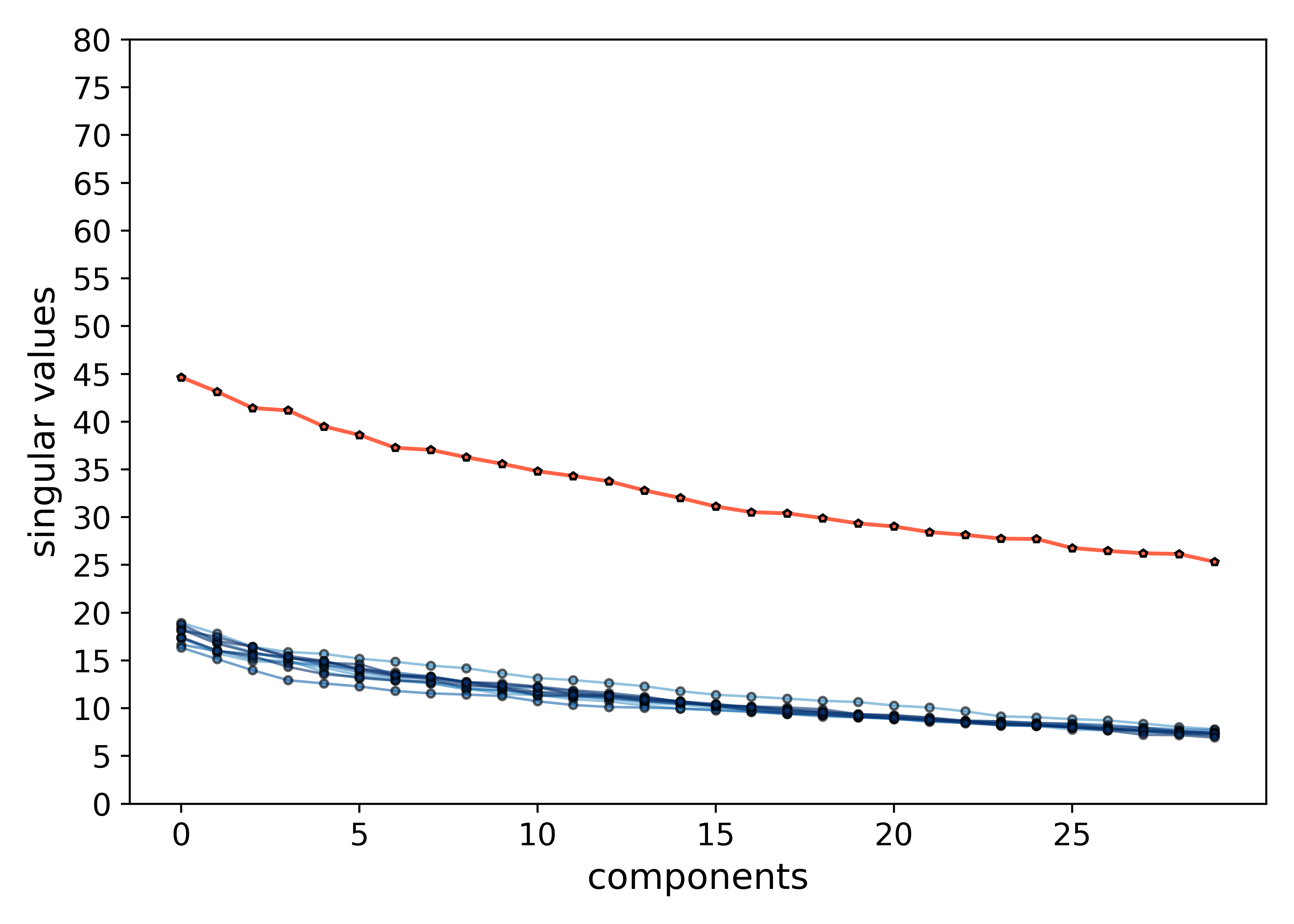}
 }
 \caption{PCAs of the features learned with Objective I, II, or III.}
 \label{ablation::mnist:objec_pca}
\end{figure}

In this section, we investigate the influence of each term of the objective function \eqref{eq:MCR2-GAN-objective} and see how they affect the learned features $\Z$, $\hat{\Z}$  and sample wise reconstruction. We follow the same experiment setting with \ours{}-Multi on MNIST (\ref{app:settings}), and conduct three experiments, each with a modified version of the original objective. Objective I is the original objective with all three terms; Objective II removes the second term $\Delta R (\hat{\Z})$, and; Objective III keeps only the third term $\Delta R (\Z, \hat{\Z})$. The results in  Figure~\ref{ablation::mnist:objec_xhat} show that using Objective II we can still maintain the sample-wise reconstruction property, but the image quality is lower when compared those constructed by Objective I (Figure~\ref{ablation::mnist:objec_xhat_b} vs. \ref{ablation::mnist:objec_xhat_c}). Objective III loses the sample-wise reconstruction property (Figure~\ref{ablation::mnist:objec_xhat_a} vs. \ref{ablation::mnist:objec_xhat_d}). Finally, the results from Figure~\ref{ablation::mnist:objec_zzhat} and Figure~\ref{ablation::mnist:objec_pca} show that without the first two terms, the learned features  $\Z$ and $\hat{\Z}$ have poor class-to-class alignment and their principal components do not show clear subspace structure {with higher singular values within each class}.

\begin{table}[t]
    \centering
    \small
    \setlength{\tabcolsep}{6.5pt}
    \renewcommand{\arraystretch}{1.25}
    \begin{tabular}{l|l}
    ~ & ~ \\
    \hline
    Objective I:         &   $\min_\eta \max_\theta  \mathcal{T}_{\X}(\theta, \eta) = \Delta R\big(\Z(\theta) \big) + \Delta R\big(\hat \Z(\theta, \eta)\big) + \sum_{j=1}^k \Delta R\big(\Z_j(\theta), \hat \Z_j(\theta, \eta) \big).$ \\ 
    Objective II:         &   $\min_\eta \max_\theta  \mathcal{T}_{\X}(\theta, \eta) = \Delta R\big(\Z(\theta) \big) + \sum_{j=1}^k \Delta R\big(\Z_j(\theta), \hat \Z_j(\theta, \eta) \big).$        \\ 
    Objective III:         &   $\min_\eta \max_\theta  \mathcal{T}_{\X}(\theta, \eta) = \sum_{j=1}^k \Delta R\big(\Z_j(\theta), \hat \Z_j(\theta, \eta) \big). $       \\
    \end{tabular}
    \caption{Three different objective function on \ours{}.}
    \label{tab::ablation_obj}
\end{table}

\subsection{Ablation Study on Sensitivity to Spectral Normalization}
\label{app:spectral-normalization}
It is known that spectral normalization is important to improve the stability of training GANs. Here, we test our formulation with and without the spectral normalization. We follow the setting from Section~\ref{app:settings} and test on CIFAR10, use the network architecture from Table~\ref{arch:cifar_d} and \ref{arch:cifar_g}. All settings of the two experiments are exactly same except with or without spectral normalization. We see that our formulation is stable in both settings and generate similar images. The only difference is that the quantitative scores in terms of IS and FID is higher with the spectral normalization.

\begin{table}[th]
    \centering
    \small
    \setlength{\tabcolsep}{9.5pt}
    \renewcommand{\arraystretch}{1.25}
    \begin{tabular}{cc|cc|cc}
    ~ & ~ & \multicolumn{2}{c|}{\ours{}-Binary} & \multicolumn{2}{c}{\ours{}-Multi} \\
    \multicolumn{2}{c|}{backbone=SNGAN} & SN=True & SN=False & SN=True & SN=False   \\
    \hline
    \hline
    \multirow{2}{*}{CIFAR-10} & IS $\uparrow$ & 8.1  & 6.6  & 7.1  & 5.8    \\
    ~                      &FID $\downarrow$  & 19.6 & 27.8 & 23.9 & 41.5    \\
    \end{tabular}
    \caption{Ablation study the influence of spectral normalization.}
    \label{tab:ablation_w_wo_sn}
\end{table}

\subsection{Ablation Study on the Trade-off between Network Width and Batch Size}\label{app:ablation-batch-size}
Empirically, we observed that for our formulation, the larger the batch size, the better the results. To justify our use of mini-batch size that is larger than those adopted in previous works such as \cite{miyato2018spectral}, we conduct the following experiment which studies the training behavior of our proposed \ours{}-Multi framework. 
Specifically, we train on the selected 10 classes of ImageNet with varying number of widest channels in our chosen architecture (specified in Section \ref{app:settings}) and mini-batch size. 
We train both the encoder and decoder from scratch without fine-tuning.
Other hyper-parameter settings detailed in Section \eqref{app:imagenet} are fixed.
We present the results in Table \ref{app::ablation_BS_width}.
In the table, we denote training sessions that do not produce meaningful images as ``failure'' and those that do as ``success''.
In the ``failure'' scenario, we noticed that the second term in the \ours{}-Multi objective \eqref{eq:MCR2-GAN-objective} would collapse to near $0$ and could not be recovered, implying the decoder has essentially lost in the minimax game.
In the ``success'' scenario, both the first terms of \eqref{eq:MCR2-GAN-objective} stay close to each other and neither would collapse to near $0$. 
The results present an interesting diagonal pattern that captures the relationship between mini-batch size and network width.
With a wider network and more channels, the network contains a greater capacity but would require a larger batch to stabilize training.
This experiment justifies our use of a larger batch in our experiment in Section \ref{app:imagenet} and also presents an interesting trade-off between network capacity and batch size for training. 

\begin{table}[th]
    \centering
    \small
    \setlength{\tabcolsep}{3.5pt}
    \renewcommand{\arraystretch}{1.25}
    \begin{tabular}{c|ccc}
    ~               & Channel\#=1024  & Channel\#=512    & Channel\#=256    \\
    \hline
    BS=1800         &   success     &  success       &    success     \\ 
    BS=1600         &   success     &  success       &    success     \\ 
    BS=1024         &   failure     &  success       &    success     \\
    BS=800          &   failure     &  failure       &    success     \\ 
    BS=400          &   failure     &  failure       &    failure     \\ 
    \end{tabular}
    \caption{Ablation study on ImageNet about trade-off between batch size (BS) and network width (Channel \#).}
    \label{app::ablation_BS_width}
\end{table}

\begin{figure}[t]
\centering
 \includegraphics[width=0.81\textwidth]{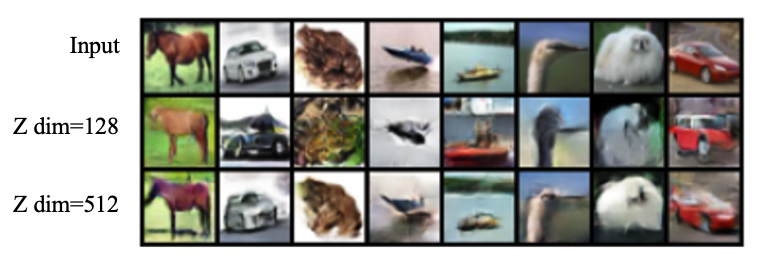}
 \caption{Reconstruction results by LDR models learned with different feature dimensions.}
 \label{ablation::zdim_viz}
\end{figure}

\begin{table}[t!]
    \centering
    \small
    \setlength{\tabcolsep}{9.5pt}
    \renewcommand{\arraystretch}{1.25}
    \begin{tabular}{cc|cc|cc}
    ~ & ~ & \multicolumn{2}{c|}{dim=128} & \multicolumn{2}{c}{dim=512} \\
    ~ & ~ & \ours{}-Binary & \ours{}-Multi & \ours{}-Binary & \ours{}-Multi \\
    \hline
    \hline
    \multirow{2}{*}{CIFAR-10} & IS $\uparrow$ & 8.1  &  7.1    & 8.4 &  8.2   \\
    ~                      &FID $\downarrow$  & 19.6 &  23.6   & 18.7 &  20.5  \\
    \end{tabular}
    \caption{IS and FID scores of images reconstructed by LDR models learned with different feature dimensions.}
    \label{tab:ablation_zdim}
\end{table}

\subsection{Ablation Study on Feature Dimension}\label{app:feature-dim}
In this paper so far, for simplicity and uniformity, we have chosen the feature dimension $d=nz$ to be 128 for all experiments. In practice, however, the choice of feature dimension may affect the performance of the learned features: common practices suggest the large the model, the better the performance could be. Hence, in this last section, we conduct experiments to show how the feature dimension affects the performance. It is not our intention to find the best feature dimension (nor the best network) with this work. We only want to show there is obviously plenty of room to improve the results presented in this paper.

The baseline experiment is conducted on CIFAR-10 with architectures from Table~\ref{arch:mnist_d} and Table~\ref{arch:mnist_g}, training hyper parameters are following the setting in Appendix~\ref{app:settings}. 
Here, we change the feature dimension $nz$, batch size, and learning rate to 512, 8196, and $0.5 \times 10^{-4}$ respectively. Figure \ref{ablation::zdim_viz} shows the comparison of (randomly selected, not cherry-picked) reconstructed images with the original ones. We observe a significantly improvement in visual quality over the results with a lower feature dimension. The IS and FID scores reported in Table \ref{tab:ablation_zdim} also confirm the improvement.

\end{document}